\documentclass[10pt,journal,twocolumn,twoside]{IEEEtran}
\pdfoutput=1
\usepackage[latin1]{inputenc}
\usepackage[english]{babel}
\usepackage{graphics}
\usepackage{graphicx}
\usepackage{algorithmic}
\usepackage[pdftex]{hyperref}
\usepackage{color}
\usepackage{fancybox}
\usepackage[config,font=footnotesize]{caption,subfig}
\usepackage{amsmath}

\hypersetup{hypertexnames, plainpages=false,colorlinks=true,pdfauthor={Juan Cardelino},pdftitle={A Contrario Selection of Optimal Partitions for Image Segmentation}}

\title{\emph{A Contrario} Selection of Optimal Partitions for Image Segmentation}
\author{Juan Cardelino\IEEEauthorrefmark{1}\IEEEauthorrefmark{2}, Vicent Caselles\IEEEauthorrefmark{2}, Marcelo Bertalm\'{i}o\IEEEauthorrefmark{2} and  Gregory Randall\IEEEauthorrefmark{1} 
\thanks{\IEEEauthorrefmark{1}Inst. Ingenier\'{i}a El\'ectrica, Universidad de la Rep{\'u}blica. Montevideo, Uruguay.}
\thanks{\IEEEauthorrefmark{2}Dept. Tecnologies de la Informaci\'o.  Universitat Pompeu Fabra. Barcelona, Spain.}
}

\renewcommand{\leftmark}{{SIAM} Journal on Imaging Sciences (to appear, 2013).}
\renewcommand{\rightmark}{CARDELINO et al.: \emph{A Contrario} selection of optimal partitions for image segmentation. Revision 7.0.}

\date{\today}

\if@twoside                 
\else                       
\fi
 \usepackage{jfc}

\newcommand{\h}{\mathcal{H}}
\newcommand{\p}{\mathcal{P}}
\newcommand{\R}{\mathcal{R}}

\newcommand{\s}{\mathcal{S}}
\newcommand{\ml}{\mathcal{L}}

\newcommand{\om}{\Omega}
\newcommand{\ob}{\mathcal{O}}
\newcommand{\kr}{k\!\mathrm{-}\!\R}
\newcommand{\kp}{\p_k}
\newcommand{\ac}{\emph{a contrario} }
\newcommand{\req}[1]{(\ref{#1})}
\newcommand{\lnfa}{LNFA}
\newcommand{\nfar}{NFA_r}

\renewcommand{\anchotres}{2.55cm}
\newcommand{\anchocuatro}{4cm}
\renewcommand{\anchocinco}{3.8cm}

\usepackage{marginnote}

\newtheorem{theorem}{Theorem}

\begin{document}
\maketitle
\begin{abstract}
We present a novel segmentation algorithm based on a hierarchical representation of images.
The main contribution of this work is to explore the capabilities of the \ac reasoning when applied to the segmentation problem, and to overcome the limitations of current algorithms within that framework. This exploratory approach has three main goals.

Our first goal is to extend the search space of greedy merging algorithms to the set of all partitions spanned by a certain hierarchy, and to cast the segmentation as a selection problem within this space. In this way we increase the number of tested partitions and thus we potentially improve the segmentation results. In addition, this space is considerably smaller than the space of all possible partitions, thus we still keep the complexity controlled.

Our second goal aims to improve the locality of region merging algorithms, which usually merge pairs of neighboring regions. In this work, we overcome this limitation by introducing a validation procedure for complete partitions, rather than for pairs of regions.

The third goal is to perform an exhaustive experimental evaluation methodology in order to provide reproducible results.

Finally, we embed the selection process on a statistical \ac framework which allows us to have only one free parameter related to the desired scale.
\end{abstract}
\begin{IEEEkeywords}
A contrario methods, Segmentation, Hierarchy, Quantitative evaluation
\end{IEEEkeywords}

\section{Introduction}
\PARstart{I}{mage} segmentation is one of the oldest and most challenging problems in image processing.  In this work we will focus on the problem of low-level segmentation, which means that our goal will be to find \emph{interesting} partitions of the image, but without a high level interpretation of the objects present in the scene. This interpretation is application dependent and should be performed by a different kind of algorithm, possibly using this low-level segmentation as input.

 Although it is a simpler problem,  even for a human observer, it is hard to determine a unique meaningful partition of an image, and it is even harder to find consensus between different observers. In general, the biggest problem is to determine the scale of the segmentation. 
 The choice of this scale is application dependent and thus the low level segmentation task should avoid it.
In \cite{segm:graphs:guigues:06:scale_sets_image_analysis} Guigues et al. remarked the idea that a low level segmentation tool should remain scale uncommitted
and the output of such an algorithm should be a multi-scale description of the image. Higher level information, as one
single partition of the image, could be obtained afterwards by using additional criteria or manually inspecting that multi-scale representation. 

An usual approach is to find a multiscale representation of the image rather than an unique partition. These representation usually takes the form of a hierarchy of nested partitions which are usually constructed by means of region merging/region growing or split and merge algorithms. In the case of region merging algorithms, the hierarchies are constructed in a a bottom-up fashion, starting with small regions and iteratively joining similar ones until a stopping criterion is met.

The literature on image segmentation is vast, and a complete review of all the families of available algorithms is beyond the scope of this work. For this reason, we will focus on the approaches based on merging algorithms, multiscale representation of images and \ac models.%
\subsection*{Region Merging algorithms}
According to Morel et al. \cite{book:morel:95:variational_pde}, the first merging algorithms presented in the literature aimed to consecutively merge adjacent regions until a stopping criterion was met, thus yielding a single partition. All these algorithms share three basic ingredients: a region model, which tells us how to describe regions; a merging criterion, which tells us if two regions must be merged or not; and a merging order, which tells us at each step which couple of regions should be merged first.
Many of the existing approaches have at least one of the following problems: the use of region information alone (no boundaries are taken into account), a very simple region model, a very simple merging order, or a considerable number of manually tuned parameters. Most of the recent work in this area has been directed to improve the merging criterion and the region model, but little effort was carried out towards the merging order and the reduction of the number of parameters.

Regarding the merging order, Calderero et al. \cite{segmentation:calderero:2010:region_merging_techniques} presented an improvement over classic merging algorithms which uses a non-parametric region model and a merging order depending on the scale. However, this work still has some free parameters and only one feature (color) is taken into account in the merging criterion.

Regarding the parameter selection, Burrus et al. \cite{burrus:2009:Image_segmentation_by_a_contrario_simulation} introduced an \ac model to remove the free parameters and a criterion combining different region descriptors. However, they require a training process which is offline but very slow, and which has to be done for each image size and for each possible parameter combination of the initialization algorithm. In addition, segmentation results are heavily dependent on those initialization parameters.%
\subsection*{Multiscale image segmentation and representation}
Instead of working at a single scale, a more robust approach is to use the intermediate results of the merging process to build up a hierarchy of partitions, which is often represented as a tree structure. Once the hierarchy is built, the tree is pruned according to some scale parameter \cite{Koepfler:1994:a_multiscale_algorithm_for_image_segmentation}.
However, most algorithms ignore the fact that this hierarchy spans a subset of all the possible partitions of the image, and that the segmentation problem could be cast as the selection of an optimal partition on this subset. 
For example, in the approach of Koepfler et al. \cite{Koepfler:1994:a_multiscale_algorithm_for_image_segmentation} the hierarchy is thresholded at a fixed scale, an operation we will call an \emph{horizontal cut}. These kind of cuts only explore a smaller subset of the possible partitions spanned by the hierarchy.\\

On the other hand, when filtering the hierarchy to find the optimum partition, the selected regions could exist at different scales on different parts of the image. 
This optimal filtering has been extensively explored in the field of mathematical morphology under many forms.
In one of the first works using these ideas, Najman et al. \cite{morphology:watershed:najman:1996:geodesic_saliency} extended the notion of watershed to produce a saliency map, in which each contour of the watershed was labeled with the \emph{importance} or \emph{saliency} of the contour. This structure encoded all possible output partitions, and a particular segmentation could be obtained by thresholding this structure.
In \cite{segmentation:morphology:soille:2008:constrained_connectivity} an approach based on constrained connective segmentation was used to produce a hierarchy of partitions of the image by scanning all possible values of the range parameters.
In Guigues et al. \cite{segm:graphs:guigues:06:scale_sets_image_analysis}, this selection was called a \emph{non-horizontal cut} of the hierarchy and computed by minimizing a Mumford-Shah energy over the hierarchy.
In a recent work \cite{meyer_najman:2010:segmentation_minimum_spanning_tree_and_hierarchies} the authors presented a detailed review of the theory and applications of hierarchical segmentation in the framework of mathematical morphology. In this context the optimal filtering is carried out by using the flooding operator, and the approaches of Najman and Guigues are explained as applications of this framework.
In \cite{segmentation:morphology:najman:2011:equivalence_between_hierarchical_segmentation_and_ultrametric} a unified theory is proposed, in which every hierarchical segmentation can be represented as an ultrametric watershed and viceversa.
In particular, the saliency maps introduced in \cite{morphology:watershed:najman:1996:geodesic_saliency} are shown to be a particular case of an ultrametric watershed. 
Finally, in a previous work \cite{segm:a_contrario:cardelino:09:a_contrario_hierarchical_segmentation}, we used a similar approach to overcome some of the usual problems of merging algorithms. In addition, our algorithm used both boundary and region information within a statistical framework which allowed us to select the optimal partition with only one free parameter, which acted as a scale parameter.\\

The work of Guigues et al. is very similar in spirit to the present work.
For this reason we will review some of its main features here and we will compare each feature of our algorithm with that work both from the theoretical and experimental points of view. The first contribution of Guigues is to define and characterize the so-called scale sets representation. 
This representation is the region based analogous of the classical Scale Space used in the context of boundary detection, which was first introduced by Marr \cite{marr:1982:computational_investigation_into_the_human} and later formalized by Witkin \cite{scale_space:witkin:1983:scale_space_filtering} and Koenderick \cite{scale_space:koenderink:1983:structure_of_images}. This representation holds two important properties: causality and geometrical accuracy.
These two conditions are equivalent to the \emph{Strong Causality Principle} introduced by Morel and Solimini \cite{book:morel:95:variational_pde}.
The second contribution is to characterize how to build such a representation by means of an energy minimization procedure.
Finally, they prove that under certain conditions, the global minimum of these energies can be efficiently computed by means of a dynamic programming algorithm.
In particular, it is important to remark that Guigues improves the locality of region merging algorithms by providing a scale climbing algorithms that allows them to compute the global optimum of a certain type of energies.

\subsection*{Objective evaluation}
Although segmentation is one of the most studied areas in image processing, there are not too many works that rely on a precise evaluation methodology, or show quantitative results. Another point to be mentioned is the reproducibility of published algorithms, since there are very few binaries or source codes available, making it difficult to determine their real capabilities and limitations.

In this sense, the work of Arbelaez et al. \cite{segmentation:Arbelaez:2010:contour_detection_and_hierarchical_segmentation} is in the same spirit as the present work. They propose a hierarchical approach constructed on top of a contour detector, which uses spectral clustering techniques to integrate different features for the segmentation task. In addition, they present a comprehensive quantitative evaluation of their experimental results.
\subsection*{Contributions}
The main contribution of this work is to explore the capabilities of the \ac approach for image segmentation. In this framework, only ideas taken from classical region merging algorithms had been proposed, which inherit the main problems of such approaches.
In this sense, we further improve the model from \cite{segm:a_contrario:cardelino:09:a_contrario_hierarchical_segmentation} by addressing two of its main problems. First by removing the locality introduced by validating only local merging decisions, computing instead the meaningfulness of an entire partition of the image. Second, we improve the behaviour of the free parameter by introducing a theoretical estimation of the number of tests.
Some of these ideas were already presented in the morphology field and in the work of Guigues. One of the original contributions of this work is to apply these ideas in the \ac framework.
Finally, we present an exhaustive quantitative evaluation of our results, by testing our algorithm with publicly available databases, and comparing the results against human segmented images using several performance metrics.
We have also created a web page with detailed evaluation results and an online demo to validate our algorithm. Let us mention that the source code can be downloaded, allowing the community to verify our claims\footnote{\url{http://iie.fing.edu.uy/rs/wiki/ImageSegmentationAlgorithms}}.%
\subsection*{Outline of the paper}
The rest of the paper is organized as follows. In Section \ref{sec:background} we review some background concepts related with our approach. In section \ref{sec:meaningful_regions} we review our previous definitions of meaningful regions. An improved \ac model, called Meaningful Partitions, is introduced in Section \ref{sec:meaningful_partitions}.
In Section \ref{sec:meaningful_partitions:algorithm} we build the proposed algorithm based on that model. Section \ref{sec:results} shows experimental results of the proposed approach. In Section \ref{sec:implementation} we discuss the implementation of the algorithm. And finally, in Section  \ref{sec:conclusions} we summarize our conclusions and discuss future work.%
\section{Background}\label{sec:background}
\subsection{Hierarchies of partitions}\label{sec:background:hierarchy}
\begin{figure}[ht]
	\centering
	\subfloat[]{\label{fig:approach:algorithm:algo:im}
		\includegraphics[width=2cm]{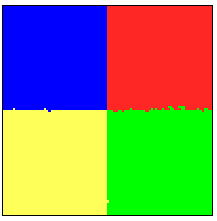}
	}%
	\subfloat[]{\label{fig:approach:algorithm:algo:tree}
		\includegraphics[width=3cm]{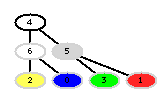}
	}\\
	\subfloat[]{\label{fig:approach:algorithm:algo:partitions}
		\includegraphics[width=6cm]{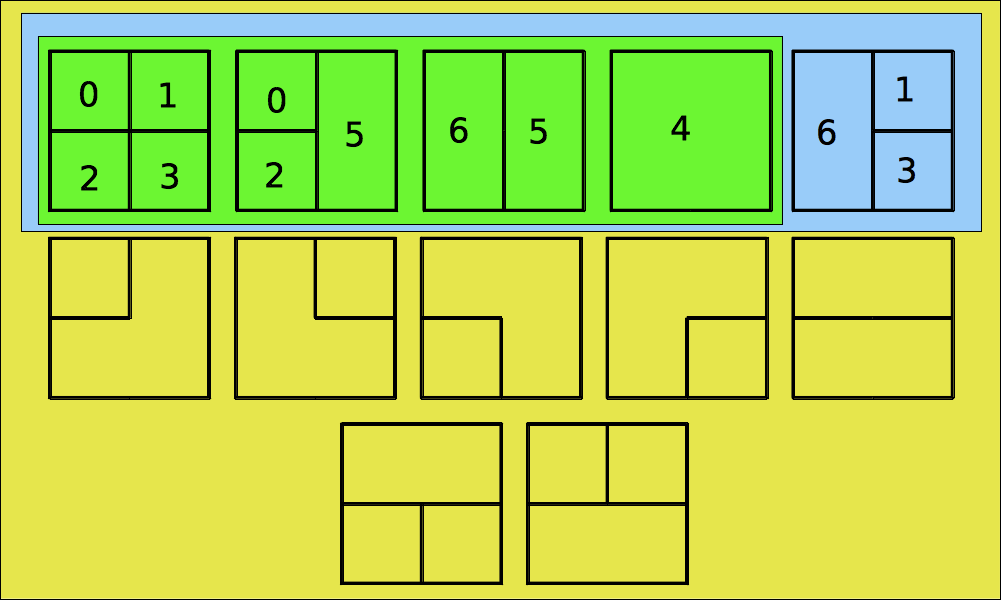}
	}%
	  \caption{\protect\subref{fig:approach:algorithm:algo:im} Sample image composed by four regions of random noise with different means. \protect\subref{fig:approach:algorithm:algo:tree} Corresponding Mumford-Shah (M-S) hierarchy, constructed from an initial partition composed of the 4 original regions (labeled 0,1,2,3). \protect\subref{fig:approach:algorithm:algo:partitions} Different choices for the partition search space. The yellow set represents the set of all possible partitions that could be generated with the initial four regions. The blue set represents all possible partitions spanned by the tree. The green set represents the partititions considered by the region merging algorithm used to construct the hierarchy. Each region is labeled with the corresponding node number in the hierarchy. Note that there are possible partitions that are not spanned by the tree.}\label{fig:approach:algorithm:algo}
\end{figure}

A hierarchy of partitions $\h$ is a multi-scale representation of many possible segmentations of an image. For our purpose, it is a tree structure where each node represents a region, and the edges represent inclusion between regions (see Fig. \ref{fig:approach:algorithm:algo}). This tree can be constructed in two ways: top-down or bottom-up. 
In the first case, the algorithm starts with the coarsest possible partition, which is iteratively split until a convergence criterion is met.
 In the second case, in which this work is contained, region merging algorithms are used to construct the hierarchy. They usually start with an initial set of regions, often called seeds, which conform the finest possible partition. Then two or more adjacent regions are iteratively merged and  the resulting regions on the graph are represented as a new node which is the parent of the merged regions. 
One of the most popular ways to merge regions is to minimize the well-known Mumford-Shah (M-S) functional \cite{segm:variational:mumford:88:optimal_approx}.
The binary tree resulting of this merging procedure is often called binary partition tree \cite{segm:trees:garrido:00:binary_partition_tree} or dendrogram in the clustering literature.

In this work we will use an initial hierarchy $\h$ constructed by means of a greedy iterative region merging algorithm, which minimizes the piecewise constant Mumford-Shah functional. However our results are easily extensible to other kind of hierarchies, like trees of shapes (also called component trees) \cite{segm:tmaps:caselles:99:topographic_maps}.
\subsection{\emph{A contrario} framework}\label{sec:background:a_contrario}
\emph{A contrario} models were introduced by Desolneaux et al. in \cite{book:desolneaux:08:from_gestalt_theory}, within the framework of the Computational Gestalt Theory (CGT). Given a set of events the aim is to find the \emph{meaningful} ones, according to the so-called Helmholtz principle, which states that: ``we naturally perceive whatever could not happen by chance''. To apply this principle, CGT algorithms rely on a \emph{background} or noise model, and then define the interesting events as those extremely unprobable under this model. To measure how exceptional an event is, they define the \emph{Number of False Alarms (NFA)}, which is an upper bound on the expected number of occurrences of a certain event under the background model. If this expectation is low, it means that the considered event is meaningful and could not arise by chance. Given an event $\ob$, its \emph{NFA} is computed as $NFA(\ob)=N_{tests}.P(\ob)$, where $N_{tests}$ is the number of possible events in the set and $P(\ob)$ is the 
probability of occurrence of event $\ob$. Defined in this way, it can be proven that if $NFA(\ob)<\epsilon$, then the expected number of detections of event $\ob$ in an image of noise generated by the background model is lower than $\epsilon$.
This reasoning is analogous to classical hypothesis testing, where one assumes a null hypothesis (images are generated by the background model) and given an observation one computes the probability of ocurrence of the event under these assumptions. The null hypothesis is then rejected if the computed probability is low enough (given a threshold). In this context, the NFA plays the role of the type I errors, i.e. rejecting the null hypothesis when the event was actually generated by it.
In the \ac jargon, this means to detect meaningful objects in a noise image, which is a False Alarm (or a false positive). For a more detailed explanation of the A contrario approach please refer to \cite{book:desolneaux:08:from_gestalt_theory}.
\section{Meaningful regions and mergings}\label{sec:meaningful_regions}
The algorithm presented in this work is the natural evolution of the algorithm presented in \cite{segm:a_contrario:cardelino:09:a_contrario_hierarchical_segmentation}, which we will denote GREEDY.
 For this reason, we will start by reviewing the basic concepts introduced in that work, and we will discuss its main problems.
 The understanding of these concepts is crucial to understand the present work and thus is important to review them here.
 Please note that the material shown in this section contains an excerpt of the aforementioned work, but with a more in-depth review of some concepts and newly added sections.\\

The main idea of the GREEDY algorithm is to combine region and boundary information to select the optimal partition from the set of all partitions spanned by the given hierarchy. This selection is embedded into an \emph{a contrario} model which states that a region is meaningful if its gray level is homogeneous \emph{enough} and its boundary is contrasted \emph{enough}.
\subsection{Region term}\label{sec:algorithm:approach:region_term}
The region term of the model will define a \emph{meaningful region} as a region which is well adjusted to a certain model. In this work we use the simplified Mumford-Shah model (M-S model from now on) and we will say that a region is meaningful when its \emph{error} is small, in a similar way to \cite{stereo:nfa:almansa:06:low_baseline}. To explain this, let us review the data term in the M-S model for a single region:
\begin{equation}\label{eq:meaningful_regions:region_term:data_term}
E_R=\sum_{x \in R}(I(x)-\mu_R)^2,
\end{equation}
where $\mu_R$ is the mean gray value of region $R$. This can be seen as the $L_2$ error when we approximate each pixel of the region by $\mu_R$. We can also define the pixel-wise error as
$e_R(x)=(I(x)-\mu_R)^2$, so with this notation the error becomes $E_R=\sum_{x \in R}e_R(x)$.

If we consider that $E_R$ is a random variable generated by the background model and we define $\hat{E}_R$ as the observed region error we can define the number of false alarms of a region as
\begin{equation}\label{eq:approach:NFA_R}
\nfar(R)=N_r . P(E_R<\hat{E}_R),
\end{equation}
where $N_r$ is the number of possible regions to consider. In practice, the number $N_r$ can be estimated only for simple problems, so a common practice is to replace the number of possible regions, by the number of \emph{tested} regions.
Defined in this way, this NFA measures the goodness of fit of the region pixels to the model given by the M-S mean $\mu_R$.
If the probability $P$ is very low, it means that the error is extraordinarily small and could not arise by chance. Thus, the NFA is small and the region will be marked as \emph{meaningful}.

In order to complete the definition of eq. (\ref{eq:approach:NFA_R}) we need to compute the probability $P(.)$ of the error in each region. We do this in two steps, first we estimate the probability density function $p(e)$ of the pixel-wise error $e(x)$. After that, we compute the probability that region $R$ has error less or equal than $\hat{E}_R$.

Note that the error $\hat{e}_R(x)$ depends on $\mu_R$ which in turn depends on the region the pixel belongs to. Before the segmentation starts, we do not know which region the pixel will be assigned to, so we can't compute this quantity. To overcome this, we can compute the error with respect to all possible regions the pixel $x$ could possibly be assigned to. In this way, we are not computing the error with respect to a single partition, but to all the possible partitions spanned by $\h$.
In Figure \ref{fig:algo:probabilty_computation} we show the  pseudo-code of the pixel-wise error computation, in order to illustrate the estimation of the probability density function of the error in detail.

\begin{figure}[h]
\centering
\shadowbox{
\begin{minipage}{7cm}
\begin{enumerate}
\item Let $\Omega \in R^2$ be the image domain. Let $R(H)$ be the set of all possible regions belonging to the hierarchy $H$
\item For each pixel $x \in \Omega $ find the set of regions $X \subset R(H)$ such that $R \in X iif x \in R$
\item The maximum possible error e(x) will occur in the case that $I(x)=255$ and $\mu_R=0$ or viceversa. The minimum possible error will occur when $I(x)=\mu_R$. Thus $x(x) \in [0, e_{max}]=[0, 255^2]$
\item Create an empty histogram $h_e$of the error e(x) in the range $[0, e_{max}]$ with $n$ bins.
\item For each $R \in X$
\begin{itemize}
\item Compute the error of pixel x with respect to X. That is $e_R(x)=(I(x)-\mu_R)^2$
\item Add $e_R(x)$ to the corresponding bin of $h_e$
\end{itemize}
 \item Normalize $h_e$ such that $\Sigma h_e(t) dt=1$
\end{enumerate}
\end{minipage}
\cornersize{.1}
}
\caption{Estimation of the probability density function of the pixel-wise error over a hierarchy.}
\label{fig:algo:probabilty_computation}
\end{figure}
Computed in this way $h_e$ could be regarded as an estimation of probability density function of an error of magnitude $\epsilon$ given the hierarchy $H$, 
$h_e(\epsilon)=p(e_R(x)=\epsilon | H)$.
We could argue that this model is a contrario in the following sense. As we don't choose any segmentation, the histogram of the error is computed from every possible region each pixel could belong to.
Note that, computed in this way, the PDF of the error is independent of any particular choice of a partition, but depends on the initial hierarchy $\h$ chosen.\\

Finally, to compute $P(E_R<\hat{E}_R)$ we make the assumption of independence between image intensities at the different locations of the $n$ pixels of the region. Thus, looking at equation (\ref{eq:approach:NFA_R}), the random variable $E_R$ is a sum of $n$ independent and identically distributed random variables $e_R$. Then, the probability can be approximated (for large values of $n$) by a normally distributed random variable, using the Central Limit Theorem (CLT). In practice, with $n>20$ the Gaussian approximation of the CLT is very accurate.
\subsection{Meaningful regions vs meaningful mergings}\label{approach:meaningful_mergings}
From our definition of meaningful region, it can be seen that the NFA associated with each region depends on the given initial partition. Given a certain tree, it is a common practice to remove the leaves with small area, to reduce the computational burden of the algorithm. Usually, leaves have small errors, so removing one of them will decrease the probability of observing a small error. If we remove a big number of small leaves we will make the small errors less probable, thus making all nodes more meaningful. This is not a desirable behavior, because we want the result of our algorithm to be independent of the pre-processing performed.

This problem arises because our definition of meaningful region is absolute, which makes it strongly dependent on the histogram of the error. One way to overcome this problem is to compare pairs of regions instead of validating each of them independently. For this reason, we introduce a similar definition of meaningfulness but applied to mergings. We will say that a certain merging is meaningful if it \emph{improves} the previous representation. That is, if the meaningfulness of the merged region is greater than the meaningfulness of considering the regions as separate entities.

To compute the meaningfulness of a merging we must compute two quantities: the NFA of the union of two regions, and the NFA of their separate existence. To compute the first quantity, we can apply the definition of the previous section. However, we need to adjust  the number of tests, because now we are not testing all possible regions, but only those created as a result of a union.
Let $R_1$ and $R_2$ be the regions to be merged and $R_u = R_1 \cup R_2$. Thus, the NFA of the union is:
\begin{equation}\label{eq:approach:mean_merge:NFA_U}
\nfar(R_1 \cup R_2)=N_u . P(E_{R_u}<\hat{E}_{R_u}),
\end{equation}
where $N_u$ is the number of possible unions in the tree, which is exactly $\frac{N}{2}$, where $N$ is the number of nodes of the tree. Here, as we modeled the union as a single region, the error $\hat{E}_{R_u}$ is computed using the mean $\mu_u$ of the union.

Now we need to consider the existence of two separate and independent regions $R_1$ and $R_2$. In this case, we have a different model for each region, given by the means $\mu_1$ and $\mu_2$, and the corresponding errors $\hat{E}_{R_1}$ and $\hat{E}_{R_2}$. As the M-S error is additive, we can consider that the total error of approximating both regions by their means is $\hat{E}_{R_1;R_2}=\hat{E}_{R_1} + \hat{E}_{R_2}$. Thus we can define the NFA of two separate regions as
\begin{equation}\label{eq:approach:mean_merge:NFA_sep}
\nfar(R_1;R_2)=N_{c} . P(E_{R_1;R_2}<\hat{E}_{R_1;R_2}),
\end{equation}
where $N_{c}$ is the number of possible couples $(R_1,R_2)$ which can be also approximated by $\frac{N}{2}$.
As the involved quantities are probabilities which could take very small values, it is usual to take the logarithm of the $NFAs$, which we will call $LNFA$. Thus, we will say that a merging is meaningful if
\begin{equation}\label{eq:approach:mean_merge:NFA_merge}
\s_r=L\nfar(R_1 \cup R_2) - L\nfar(R_1;R_2) < 0.
\end{equation}
This condition is very similar to the Patrick-Fisher distance used for region merging in classic approaches.
For instance in section 2.2 of \cite{segm:act_reg:zhu:96:reg_comp}, a parametric form of this test for the Gaussian case is used to test if two regions should be merged or not.
This approach relies on deciding if both samples come from the same underlying distribution, and if true (up to a confidence parameter) they should be merged.
The main difference resides here in the relative nature of our term, which not compares both regions but decides if the union is a better explanation than the separate regions.
\subsection{Boundary term}\label{sec:meaningful_regions:boundary_term}
Regarding region boundaries, we propose to merge two regions when the boundary between them is not contrasted enough. In addition, we want to obtain the regularizing effect of the boundary term in the M-S functional, which favors short curves \cite{segm:variational:mumford:88:optimal_approx}. 
For this reason, we use a definition similar to the one introduced by Desolneaux et al. \cite{segm:morel:01:edge_detection_helmholtz} and Cao et al. \cite{segm:cao:03:extracting_meaningful}, in the sense that we say a curve is meaningful if it has an \emph{extraordinarily} high contrast. However, we also penalize long curves by using the accumulated contrast along the curve instead of the minimum as in Cao's model.
Thus, we define \emph{meaningful regions} as those having a short and contrasted boundary. To measure the length of the curves we use the geodesic curve length
\begin{equation}\label{eq:meaningful_regions:boundary_term:length}
\ml(\Gamma)=\int_{\Gamma}l(x(s))ds,
\end{equation}%
where $l(x)=g\left(|\nabla I(x)|\right)$ is a pixel-wise contrast detection function. Here $g(x)$ yields small (near $0$) values in well contrasted pixels and values near 1 in the low contrasted ones.
In this case we learn the background model from the image to be segmented, so we use the histogram of the gradient as proposed in \cite{segm:cao:03:extracting_meaningful}.
 From the image, we can obtain the empirical distribution of the gradient norm, or the distribution of $l(x)$ which is a function of $|\nabla I|$. However, we need to compute the distribution of the sum over all the pixels of the curve, so our new random variable will be $L=\sum_{x\in\Gamma}l(x)$.
As we did in section \ref{sec:algorithm:approach:region_term} we can compute the probability by means of the CLT from the distribution of $l$, and define the NFA of a curve $\Gamma$ as
\begin{equation}
 NFA_b(\Gamma)=N_{curves}.P(L<\hat{L}),
\end{equation}
where $N_{curves}$ is the number of possible curves in the image, which is again approximated by the number of curves tested. In our case, it is the number of curve segments separating two neighboring regions in $\h$.
\begin{figure}[h]
\centering
\shadowbox{
\begin{minipage}{7cm}
\begin{enumerate}
\item construct the initial partition $\p$
\item build the hierarchy $\h$
\item for each couple $C=(R_1,R_2)$ compute its \lnfa
\item for each possible height $h$
\begin{enumerate}
\item while $N(h)\neq \emptyset$
\item select the most meaningful couple $C=(R_1,R_2)$
\item merge the couple $C$.
\item remove $C$ from $N(h)$.
\end{enumerate}
\end{enumerate}
\end{minipage}
}
\caption{Outline of the GREEDY algorithm.}\label{fig:algo:ms_greedy}
\end{figure}
\subsection{Combination of terms}
Using the two previous definitions of meaningful regions, we developed an unified approach which is able to take into account both definitions at the same time. For that purpose, we propose to compute the following quantity:
\begin{equation}\label{eq:approach:nfa_joint}
NFA_j(R)=N_r.P\left(E_R<\hat{E}_R ;L_{\partial R}<\hat{L}_{\partial R}\right),
\end{equation}
where $N_r$ is the number of tested regions and $P$ is the joint probability of the region having a small error and a contrasted boundary at the same time.
A way to make this model computable is to make the (strong) assumption of independence between boundary and region terms. This allows us to factorize $P$ and write the new NFA as
\begin{equation}\label{eq:approach:nfa_joint_indep}
NFA_j(R)=N_r . P(E_R<\hat{E}_R) . P(L_{\partial R}<\hat{L}_{\partial R}).
\end{equation}
Taking this into account, we say that a merging is meaningful using region and boundary information, if
\begin{equation}\label{eq:approach:mean_merge:sign_comb}
\s_j=\lnfa_j(R_1 \cup R_2) - \lnfa_j(R_1;R_2) < 0.
\end{equation}
Definition (\ref{eq:approach:mean_merge:sign_comb}) allows us to construct a parameterless algorithm; however, in practice we verified that our algorithm tends to oversegment images. The explanation of this phenomenon relies on our estimation of the number of tests. This estimation is very hard to compute analytically, so we used a very rough estimate ($N_{u} \approx N_{c}$).
To overcome this problem, we consider an alternate formulation based on the following observation. Remembering that

\begin{footnotesize}
\begin{eqnarray}\label{eq:approach:mean_merge:probabilities}
P(R_1 \cup R_2)= P(E_{R_u}<\hat{E}_{R_u}) . P(L_{\partial {R_u}}<\hat{L}_{\partial {R_u}}) \\
P(R_1;R_2)\!=\! P(E_{R_1;R_2} \! <\hat{E}_{R_1;R_2}) . P(L_{\partial {R_1;R_2}} \! <\hat{L}_{\partial {R_1;R_2}}) \nonumber
\end{eqnarray}
\end{footnotesize}

we can expand equation (\ref{eq:approach:mean_merge:sign_comb}) to explicitly show the number of tests, obtaining

\begin{footnotesize}
\begin{equation}\label{eq:approach:mean_merge:sign_comb_expanded}
\s_j= \log N_{u} + \log P(R_1 \! \cup \! R_2) -\! \log N_{c} -\! \log P(R_1;R_2) 
\end{equation}
\end{footnotesize}

Thus, if we do not want to estimate both numbers of tests, we can merge them into a single variable called $\alpha$. So, our definition of meaningful merging becomes:
\begin{equation}\label{eq:approach:mean_merge:sign_comb_alpha}
\s_j=\log P(R_1 \cup R_2) - \log P(R_1;R_2) < \alpha(R_1;R_2),
\end{equation}
where $\alpha(R_1;R_2)=\log (\frac{N_{c}}{N_{u}})$. For the sake of simplicity, we assume that $\alpha$ is a constant value for every couple of regions $(R_1,R_2)$. In \cite{segm:a_contrario:cardelino:09:a_contrario_hierarchical_segmentation} this parameter was set manually, but it could be estimated as in \cite{burrus:2009:Image_segmentation_by_a_contrario_simulation}.
\subsection{The algorithm}
To explain the GREEDY algorithm, we need to introduce some notation. We define the height of a node $n$ as the length of the longest branch starting on a leaf and ending in node $n$. We also define $N(h)$ as the set of nodes of height $h$. Starting from an initial partition, composed either by all pixels or by a set of given regions, we iteratively merge pairs of regions as shown in Figure \ref{fig:algo:ms_greedy}.
\subsection{Initial Hierarchy}\label{sec:meaningful_regions:initial_hierarchy}
The presented algorithm does not need a particular choice of the hierarchy $\h$ to be applied, as long as it can be encoded as a tree structure. However, the results are dependent of the initial $\h$ chosen.
As we mentioned in the introduction, in this work we will use a greedy region merging algorithm which obtains a local minimum of the piecewise constant M-S functional. The implementation used is inspired in \cite{Koepfler:1994:a_multiscale_algorithm_for_image_segmentation}.
The first step of the algorithm is to compute a region adjacency graph (RAG) of the pixels of the image, let $\mathcal{G}$ be this graph. At the same time, we initialize the tree structure $\mathcal{T}$, where each node of the RAG is a leaf. To speed up the computation, we create a heap with the initial edges of the RAG, where the cost associated to each edge is the minimum value of $\lambda$ in which the node will be removed. Then we will iteratively merge adjacent regions, picking in each step the one that will decrease the M-S energy the most, taken from the top of the heap. Each time a pair of regions is merged, the corresponding edge is removed from the graph, and a new node is created in the tree.
This node will be linked as the parent of the two merged ones, and have as an attribute the $\lambda$ value at which it has been created. This value is called \emph{scale of appearance} in Guigues paper.
The algorithm stops when only one region is left and the RAG is reduced to a trivial form, this node will be the root of the tree.

\subsection{Complexity}\label{sec:meaningful_regions:complexity}
The computational cost of the algorithm can be roughly divided in two parts: computing the tree and selecting the meaningful regions on it. A usual way to construct the tree is to use all pixels as the initial partition. So, for a $n$ pixel image, the tree will have $2n-1$ nodes. The computational cost of the second part is proportional to the number of nodes of the tree, thus we can reduce the computational cost by pruning the tree.

In practice we rarely segment regions of small area, thus we remove the lower nodes of the tree corresponding to the smaller regions. To explain this, let us recall the simplified M-S functional:
given a partition $\p$ of the image, the energy $F(\p)=E_R+\lambda \mathcal{L}_{\partial R}$ balances the data adjustment term $E_R$ with the regularity term $\mathcal{L}_{\partial R}$. The parameter $\lambda$ controls the relative importance of each term.
The pruning of the tree is performed with a fixed (and small enough) value of the scale parameter $\lambda$, which avoids that important regions are lost.

We start by estimating the complexity of the construction of the initial tree.
The first step is to construct the RAG, which has a complexity of $\Theta (n)$. Then we build a priority queue, using a binary heap as the underlying data structure, which will cost also $\Theta (n)$.
If we merge only one region at a time, we will have $n-1$ steps, and at each step we have to remove the top of the heap and re-order it, which will take $\Theta (\log{n})$ operations. In addition, we have to update the $\lambda$ values of all the edges incident into the merged regions.
Assuming an average degree of $k<<n$ neighbors per node, and taking into account that changing the value of an edge implies re-ordering, the complexity of this part will be $\Theta (k\log{n})=\Theta (\log{n})$. So the total complexity of the construction of $\h$ is $\Theta (n\log{n})$.
Assume then a pruning that keeps only $N$ nodes of the tree. Usually, $N<<n$ and the reduction is more than linear, a rough but reasonable estimate being $N\approx n^{\gamma}$, with $\gamma \leq 0.5$.
Regarding the complexity of the region selection part, we need to visit all pixels to compute the M-S terms on the initial partition, which costs  $\Theta (N)$. Then we need to visit the nodes once to compute the error and the probabilities. Then, at each step of the algorithm, we need to find the minimum LNFA node, which costs $\Theta (\log{N})$.
Thus the complexity of this part is $\Theta (N)$. Given our assumption that $N<<n$, the overall complexity of the algorithm is dominated by $\Theta (n\log{n})$.
\subsection{Implementation}
For example, in the \emph{Church} image (Fig. \ref{greedy:im:merge_length:04:detalle}) the total number of pixels is 38400, but using $\lambda \! = \! 50$ we have to process only 2297 regions, which results on a $94\%$ reduction on the computational cost. In spite of this pruning, the resulting image still retains enough level of detail and no important objects are lost. Note that this preprocessing step is optional and that the final result of the algorithm is independent of it, it is just performed to speed-up the algorithm.
\renewcommand{\resDir}{im}
\renewcommand{\resSubDir}{merge_length}
\renewcommand{\figDir}{img/ms_nfa_merge_length/church}
\begin{figure}[ht]
  \centering
  \subfloat{ \label{\resDir:\resSubDir:04:detalle:init}%
  \includegraphics[width=\anchocinco]{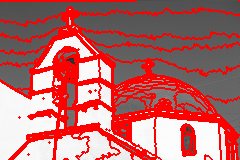}%
  }%
  \subfloat{ \label{\resDir:\resSubDir:04:detalle:res}%
  \includegraphics[width=\anchocinco]{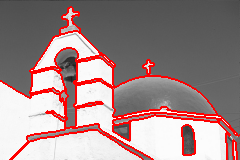}%
  }%
  \caption{Results of the GREEDY algorithm on the (240x160) \emph{Church} image. The original image has $38400$ regions. Left: Initial regions after pruning with $\lambda=50$, $2297$ regions.
  Right: Segmentation results with $\alpha=60$, $24$ regions. }
  \label{greedy:im:merge_length:04:detalle}
\end{figure}
\section{Meaningful partitions}\label{sec:meaningful_partitions}
One of the main problems of the previous approach is its locality: as we are validating only one merging at a time in a bottom-up fashion, all further decisions are conditioned by the previous merging performed.
For this reason, in this section we seek a more global definition of meaningfulness which would measure the goodness of fit of a whole partition instead of the goodness of fit of a single merging operation.

\subsection{Multi-regions \& meaningful partitions}\label{sec:meaningful_partitions:multi_regions}
We start by defining the basic concepts of \emph{multi-regions} and \emph{multi-partitions}. Given a set of $k$ disjoint meaningful regions $R_i$, one can ask if the union $R$ of them is a better explanation of the image than the separate regions alone. To deal with this we define a multi-region $\kr=\{R_i\}_{i=1}^k$ as the entity resulting of considering the $k$ regions as separate entities. A multi-partition on the other hand is a special case of a multi-region where the union of the composed regions gives the whole image domain $\Omega$, i.e. $\bigcup_i R_i = \Omega$.\\
We will define a $k$-multipartition \mbox{$\p_k$} as a partition composed by $k$ regions and the number $k$ will be called the order of the partition.\\
As explained in \ref{approach:meaningful_mergings} we can model the separate existence of $k$ regions, using $k$ means $\mu_i$ and $k$ corresponding errors $\hat{E}_i$ obtained by approximating those regions with their means. Thus the total error committed when approximating the image with $k$ regions will be $\hat{E}_{\p_k}=\sum_{i=1}^k\hat{E}_i$. \\
Then we can define the \emph{NFA} of a multi-partition as
\begin{equation}\label{eq:approach:nfa_multipart}
	NFA(\kp)=N(n,k) P(E_{\kp}<\hat{E}_{\kp}),
\end{equation}
where $N(n,k)$ is the number of possible k-partitions that could be formed from an image with $n$ pixels.
Again, the two key points here are the computation of $N(.)$ and $P(.)$.
In the following, we explain these computations in detail.
\vskip 11pt
\subsubsection{Number of multi-partitions}\label{sec:meaningful_partitions:number_of_mp}
On most existing \ac  models (see \cite{book:desolneaux:08:from_gestalt_theory}) the number of tests is computed analytically, because the events to be considered are simple enough to enable the computation. However, in many cases, these computations are too difficult to carry out analytically. This leaves us with two options: either estimate bounds on these quantities or approximate them empirically.
In our particular case, we need to count for each of the image dimensions $p,q$ and for each order of partition $k$, the number $N(p,q,k)$ of possible partitions of $k$ regions that could be constructed in this image.
This is a very hard combinatorial problem which could be found under various forms in many fields of mathematics, and is a problem still open with no suitable solutions available for our case. One of the closest problems studied comes from the clustering literature, where they count the number of separable clusters that can be constructed with $n$ unorganized points in $\R^d$. In 
\cite{math:alon:99:separable_partitions}, the authors propose bounds on the number $N_c(n,d,k)$ of ways in which $k$ clusters can be constructed by splitting those points with hyperplanes. The regions that can be built in this way are simple, in the sense that they cannot contain holes. Thus, the number of regions estimated in this way will be a rough lower bound on the total number of regions. The number of points to be organized into regions in our case is $n=pq$ and our space has dimension $d=2$, thus the number of possible partitions could be expressed as $N(n,k)$, and the lower bound  under these hypotheses is $\hat{N}(n,k)=O(n^{6(k-2)})$.
\vskip 11pt
\subsubsection{A true a contrario model}
The background model explained in section \ref{sec:algorithm:approach:region_term} computed the empirical PDF of the errors, considering all the regions a pixel could belong to. However, those were not all possible regions but only those spanned by the hierarchy $\h$. This has two main problems: first, the statistics are dependent on the hierarchy used, thus it is not a true \emph{a contrario} model; second, the computed statistics are dependent on the preprocessing performed.
The second problem was mostly alleviated by the validation of mergings introduced in section \ref{approach:meaningful_mergings}, however the first problem is still present.
To address this, we propose a new \emph{a contrario} model, which states that under the hypothesis of noise, we should not detect any region. A detection in our case is to find a meaningful partition of the image, so if the image is generated by an i.i.d. noise, we should not find any partitions and we should say that the image is composed of a single region. This is equivalent, using our multi-partition model, to say that our random image is a $\p_1$. Thus, to compute the probability, we will consider the error of approximation of the whole image by only one global mean $\mu$.
\begin{table}[t]
\centering

\begin{tabular}{|c|c|c|c|c|}
\hline
	\hspace{-5pt} order (k) \hspace{-5pt} & \hspace{-5pt} $\log \hat{N}(n,k)$ \hspace{-5pt} & error & $\log P(\mathcal{P}^*)$  & \hspace{-5pt} $\lnfa (\mathcal{P}^*)$ \hspace{-5pt} \\
\hline
	1 & 0		& 134.5	& -0.68		& -0.7 \\
	2 & 18.5	& 18.4	& -2909.8	& -2891.4 \\
	3 & 55.3	& 24.5	& -7687.1	& -7631.9 \\
	4 & 110.5	& 1.8	& -11173.9	& -11063.3 \\
\hline
\end{tabular}
\caption{Sample result of the MP algorithm for the image in Fig. \ref{fig:approach:algorithm:algo}. For each order $k$ we compute the lower bound of the number of partitions $\hat{N}(n,k)$, we select the \emph{best} partition $\mathcal{P}^*$ (the one with lowest probability), and we compute its corresponding NFA. Note that the lower the LNFA value is, the more meaninfgul the partition.}
\label{table:approach:multipartition}
\end{table}
\vskip 11pt
\begin{figure}[ht]
\centering
\shadowbox{
\begin{minipage}{7cm}
\begin{enumerate}
\item construct the initial partition $\p$.
\item build the hierarchy $\h$.
\item for each internal node $r_i$: compute the number $n_{r_i}$ of possible partitions spawned by its subtree, and its probability $P(.)$ according to Section \ref{sec:meaningful_partitions:probability_computation}.
\item start from the root node.
\item recursive step: 
\begin{enumerate}
\item for the current node $r$ (let $a,b$ be its siblings):
\item create a table of probabilities $T_r$ of length $n_r$; in each position $k$ of the table, we will store the probability of a k-partition.
\item for each possible partition of order $k>1$ between $1$ and $n_r$: compute the probability $p$ calling the recursive function $p=prob(r,k)$.
\begin{enumerate}
\item find all possible couples of subpartitions of order $i,j$ (of nodes $a$ and $b$) such that $i+j=k$.
\item compute the probability as $p(i,j)=prob(a,i)*prob(b,j)$
\item return $p$ as $\min_{i,j}p(i,j)$.
\end{enumerate}
\item if $k=1$, $p(r,1)$ is the probability computed in 3.
\item  store $p$ in $T_r$ at position $k$.
\end{enumerate}
\item for each order $k$ in the root node table ($T_{root}$), compute the $NFA$ from the probabilities and the number of tests. Call these computed quantities, $NFA(root,k)$.
\item select the final partition order as $o=argmin_k\left\{ NFA(root,k) \right\}$.
\end{enumerate}
\end{minipage}
}
\caption{Outline of the MULTIPARTITION (MP) algorithm.}\label{fig:algo:ms_multipart}
\end{figure}
\subsubsection{Probability computation}\label{sec:meaningful_partitions:probability_computation}
For the computation of the probability in (\ref{eq:approach:nfa_multipart}) we rely again on the fact that errors are additive and use the CLT to estimate the probability, as we did in previous sections.
Since a $\kp$ is composed by $k$ independent regions $R_i$, under the assumption of statistical independence we can compute its probability as
\begin{equation}\label{approach:prob_multipart}
	P(E_{\kp}<\hat{E}_{\kp})=\prod_i P(E_{R_i}<\hat{E}_{R_i})
\end{equation}
and to compute the probability of each region, $P(E_{R_i}<\hat{E}_{R_i})$, we can use the CLT as we did  in section \ref{sec:algorithm:approach:region_term}.
\section{The MULTIPARTITION(MP) algorithm}\label{sec:meaningful_partitions:algorithm}
We have seen in the previous section a way to compute the NFA for a given partition. After that, the straightforward procedure is to evaluate this NFA for all possible partitions. However, as we have shown, the number of partitions is enormous and thus we can not afford to explore all of them. For this reason, we introduce a fast algorithm to compute probabilities over the hierarchy $\h$.
In the first place we compute the mean value and the error for each node on the tree. Then, the NFA for each multi-partition is computed recursively over the tree starting from the root. The key point to understand our algorithm is to bear in mind that we are not only obtaining the best overall partition, but we also keep the best $k$-multipartition for every possible $k$.

To explain this algorithm (outlined in Fig. \ref{fig:algo:ms_multipart}), let us start from the results it yields. Given a hierarchy $\h$ with $n$ leaves, the order of all possible partitions is bounded between $1$ and $n$. For each order $k$ there is a subset of $N(n,k)$ possible multi-partitions to choose from, and our algorithm will pick the \emph{best} (lowest NFA) partition from that subset. The output of this algorithm can be summarized as a table with the best NFA for each partition order, as shown in Table \ref{table:approach:multipartition}.
So at the end,  to choose the best partition we will only need to choose the order $k$ with the lowest NFA, among all multi-partitions of each order, from $\p_1$ to  $\p_n$.
Note in Table \ref{table:approach:multipartition} the opposite effects of the probability (which in turn relates to data adjustment) and the number of tests (which represents a notion of regularity or complexity of the partition). Thus, the optimal partition comes from the balance between number of tests and probability; or as the previous interpretation suggests, a balance between complexity and data fidelity.
An interesting side effect of the way this algorithm works can be seen in Table \ref{table:approach:multipartition}. As we keep the best partition for each order $k$, we could do a variant of the algorithm were we specify the desired number of regions. This can be particularly useful in some applications were we know  beforehand how many objects we are looking for. Some examples of this approach over real biological images are shown in section \ref{sec:results:fixed}.
\subsection{Synthetic Images}\label{sec:results:synthetic}
\renewcommand{\resDir}{im/set_cielab_final/gris/ejemplo03}
\renewcommand{\figname}{fig:res_synt:blobs}
\renewcommand{\ancho}{\anchotres}
\begin{figure}[t]
	\centering
	\subfloat{ \label{\figname:orig}%
	\includegraphics[width=\ancho]{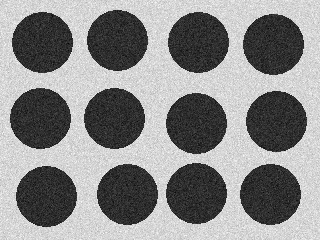}%
	}%
	\subfloat{ \label{\figname:init}%
	\includegraphics[width=\ancho]{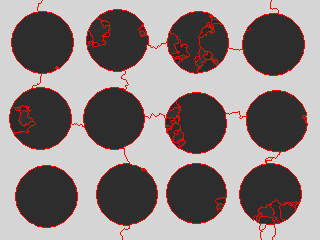}%
	}%
	\subfloat{ \label{\figname:res}%
	\includegraphics[width=\ancho]{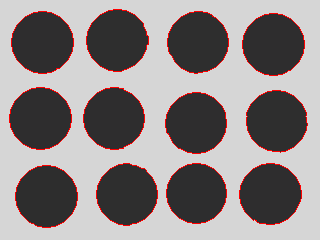}%
	}%
	\caption{\emph{Blobs}: Example of segmentation over a synthetic image. \protect\subref{\figname:orig} Original Image $N=76800$ pixels. \protect\subref{\figname:init} Initial regions after pruning with $\lambda=7$, $N_i=117$  \protect\subref{\figname:res} Segmentation results using only the data term $N_d=13$. $N$,$N_i$ and $N_d$ are the number of regions of each stage respectively.}
	\label{\figname}
\end{figure}
To explain the basic behavior of our algorithm, we show in Figure \ref{fig:res_synt:blobs} a simple synthetic
example where the image is a piecewise constant image with gaussian noise added. In this case, the image fully complies with the model, so it is the best case for the algorithm.
The objects to detect are dark blobs over the light background.
As explained in section \ref{sec:meaningful_partitions:algorithm}, the output of our algorithm is the NFA of each partition. 
In Figure \ref{fig:res_synt:plots} (left column) the NFA is plotted against the order of the partition.
As it can be seen, the minimum NFA corresponds to order 13, which is the number of objects in the image. Looking at the probability graph in Fig. \ref{fig:res_synt:plots} 
we can see that, starting from a 1-partition, the probability shows a great decrease every time we add a new region corresponding to one of the blobs. Once we add all the blobs, we
start adding subdivisions of the blobs (or the background) which only slightly improve the model, so the rate of decrease is dramatically slowed. This
phenomenon occurs in all images, but it is more salient when working with this kind of synthetic images.
The right side of Figure \ref{fig:res_synt:plots} illustrates the same behavior in a more realistic case, which is the \emph{Church} image of Figure \ref{greedy:im:merge_length:04:detalle}.
In this case, we can see that the decrease of the probability shows softer transitions around the desired number of objects, and it is not that clear how to choose the optimal partition. In addition, as the probability is always decreasing, the algorithm would select the finest partition available. But as the NFA graph shows, the number of tests acts as a regularizer, penalizing finer partitions and thus helping to choose the correct segmentation (around 36 regions in this case). 
\renewcommand{\resDir}{im/set_cielab_final/gris/ejemplo03}
\newcommand{\resDirB}{im/set_cielab_final/bsds_subset/ejemplo01}
\renewcommand{\ancho}{3.8cm}
	\begin{figure}[ht]
	\centering
	\includegraphics[width=\ancho]{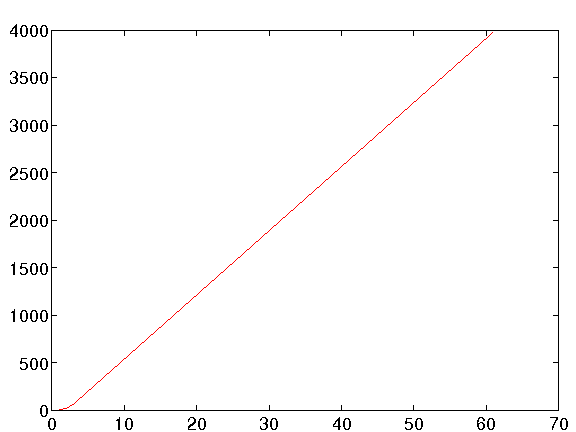}%
\includegraphics[width=\ancho]{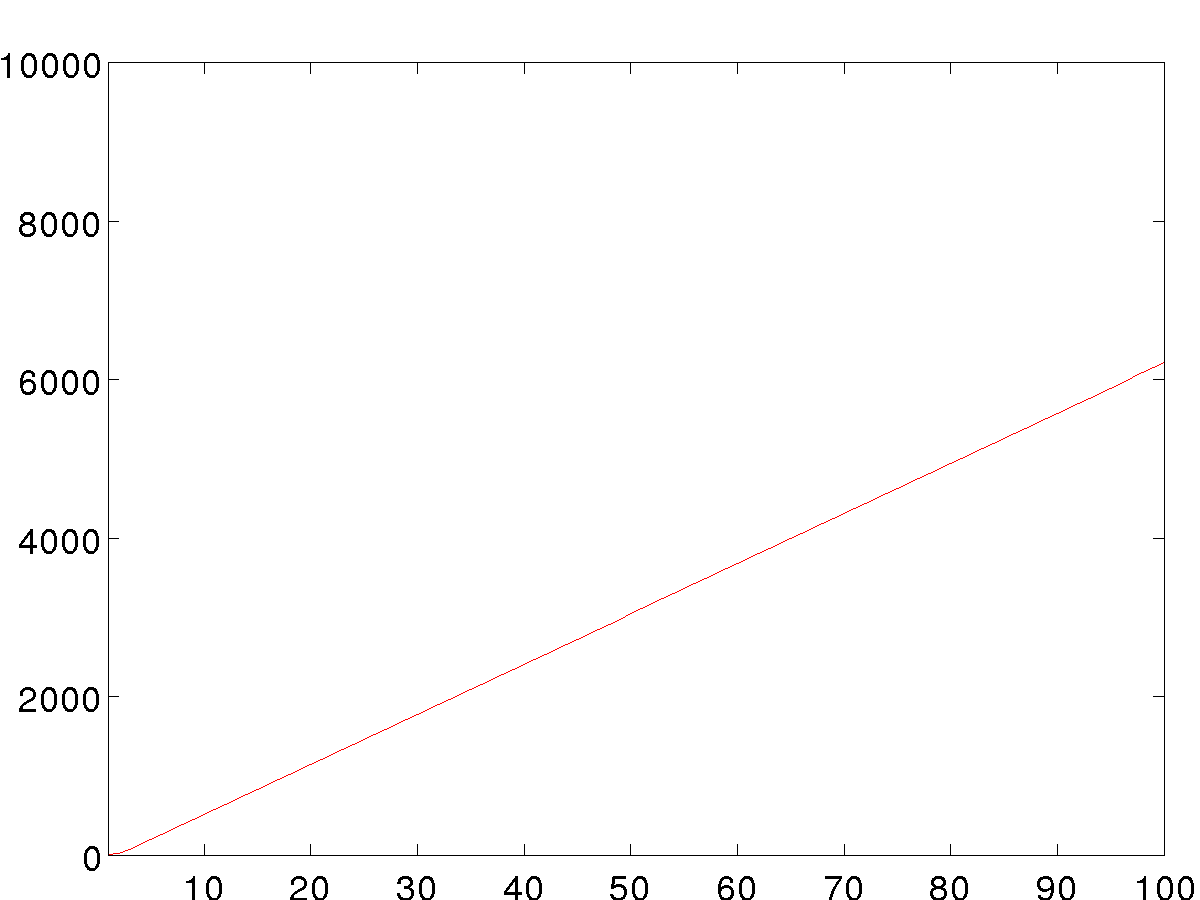}
	\includegraphics[width=\ancho]{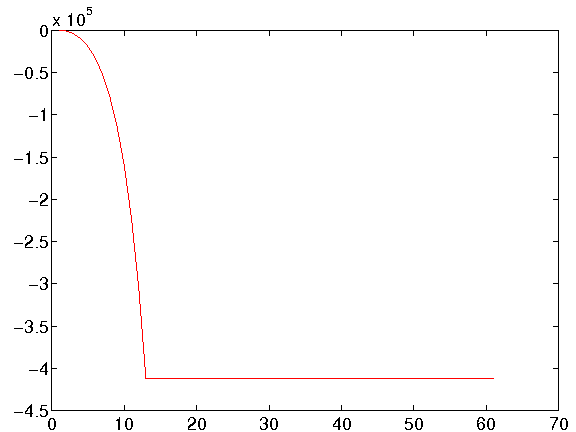}%
\includegraphics[width=\ancho]{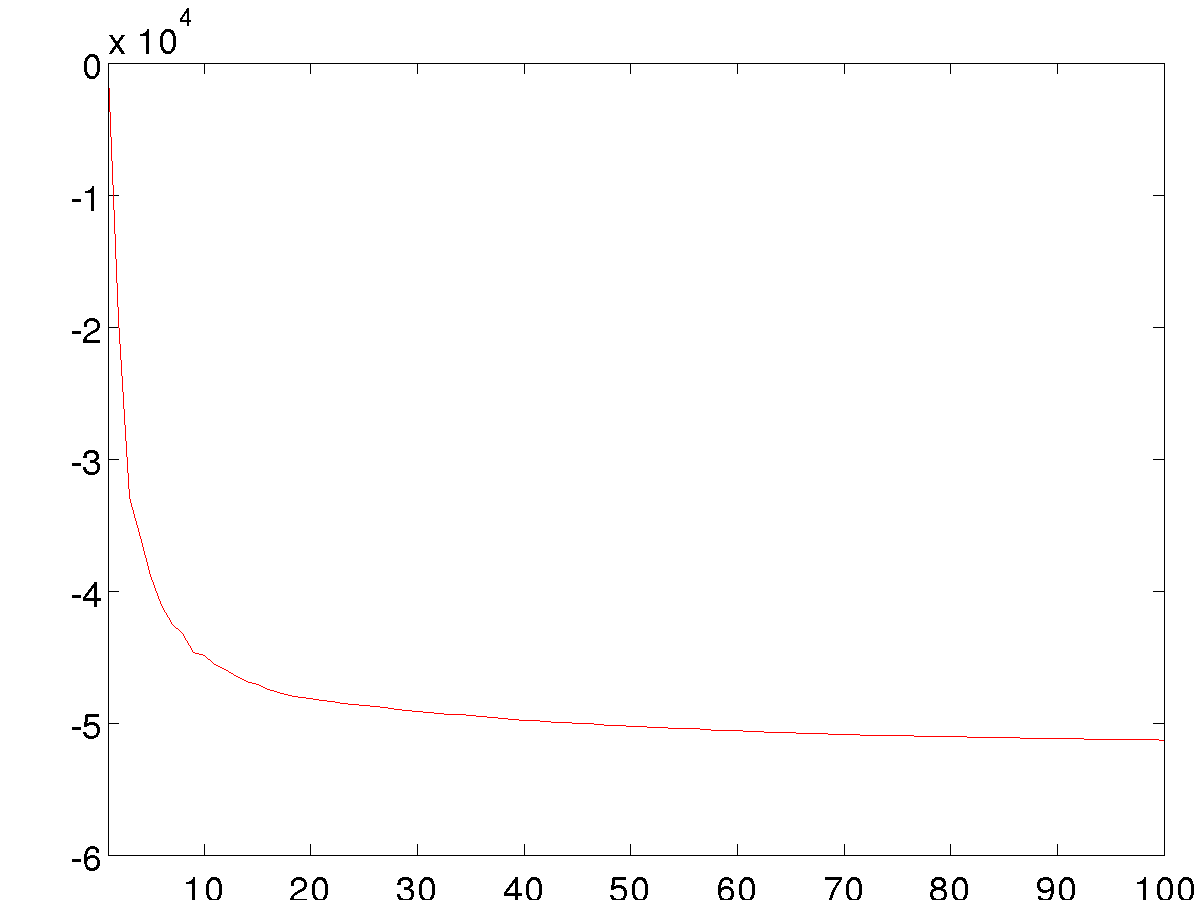}
	\includegraphics[width=\ancho]{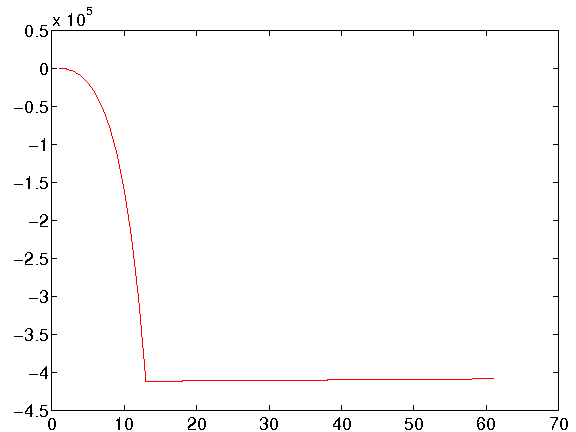}%
\includegraphics[width=\ancho]{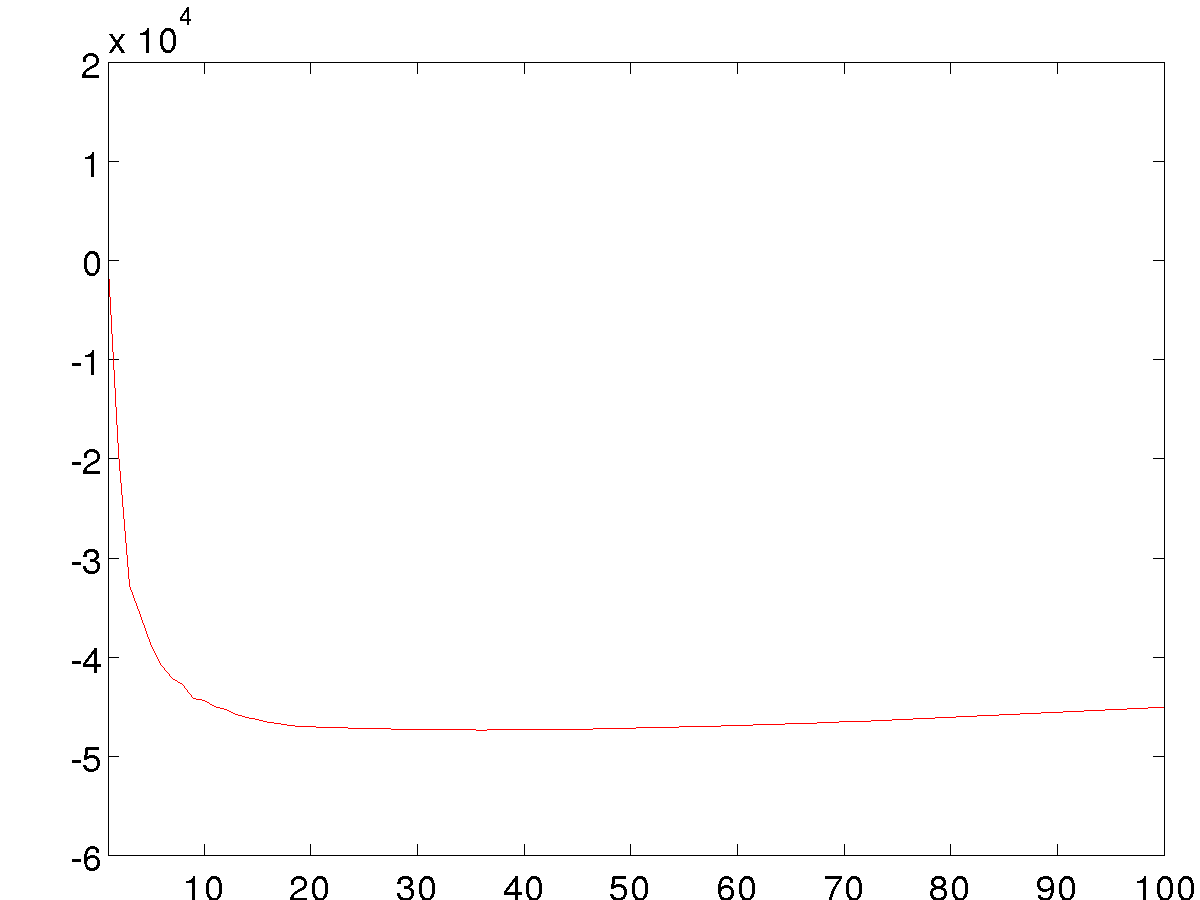}%
	\caption{Analysis of the results on the \emph{Blobs} image of figure \ref{fig:res_synt:blobs} (left) and the  \emph{Church} image of figure \ref{im:cielab_ranking:01:rank} (right). In each row we show the values of the different terms in eq. (\ref{eq:approach:nfa_multipart}). First row: logarithm of the number of tests. Second row: logarithm of the probability. Third row: LNFA. Note that the lower the LNFA value is, the more meaninfgul the partition.}
	\label{fig:res_synt:plots}
\end{figure}%
%
%
%
\subsection{Real Images}\label{sec:multipartition:real}
In the case of the Church image, the algorithm has selected the partition with 36 regions as the optimal one. However, as the difference in probabilities is low we are not certain of which one is the best partition. In this case we can also output a ranking of segmentations, and provide the user with the top $M$ segmentations and let him/her choose the one most suitable for his/her application. In Figure \ref{im:cielab_ranking:01:rank} we show the results of such approach. The numerical results corresponding to this experiment are shown in Table \ref{table:results:multipartition:church}, where we can confirm the fact that most of the top selected partitions have similar meaningfulness.
\renewcommand{\anchocinco}{2.5cm}
\renewcommand{\resDir}{im}
\renewcommand{\resSubDir}{cielab_ranking}
\renewcommand{\figDir}{im/cielab_ranking}
\figMeanRegionsBestRawB{01}{Ranking of partitions ordered by NFA for the \emph{Church} image. From left to right: Original image and the five best partitions selected (using data term only).}
\begin{table}[ht]
\centering
\begin{tabular}{|c|c|}
 \hline
 	order (k) & LNFA (best $\mathcal{P}$)\\
 \hline
 	36 & -47374.1   \\
 	35 & -47373.5    \\
 	37 & -47371.7  \\
 	39 & -47369.9  \\
 	38 & -47369.1  \\
 \hline
\end{tabular}
\caption{Best partitions selected by the multipartition (MP) algorithm for the \emph{Church} image.}\label{table:results:multipartition:church}
\end{table}

\subsection{Recursive computation}\label{sec:meaningful_partitions:recursive_computation}
The results in Table \ref{table:approach:multipartition} can be computed in a recursive way as follows. First we compute the NFA for each region in $\h$, using eq. (\ref{eq:approach:nfa_multipart}). Consider the subtree $\{6,2,0\}$ in the hierarchy of Figure \ref{fig:approach:algorithm:algo:tree}: it can only span 2 partitions, a $\p_1$ composed by node $6$ and a $\p_2$ composed by nodes $\{2,0\}$, whose probabilities are shown in Table \ref{table:approach:multipartition:detail}. The same reasoning can be applied to the $\{5,3,1\}$ subtree. Thus, when analyzing the whole tree, we can see that the two possible $\p_3$ can be obtained by combining $\p_1$ and $\p_2$ from both subtrees. Then as we are interested in the most meaningful $\p_3$, we will pick the one with lowest probability (which is $\{6,3,1\}$). In that way we can compute the probability of any $\kp$ from the probabilities of $i$-$\p$ and $j$-$\p$, with $i+j=k$, obtained from the subtrees associated to its siblings.
\begin{table}[ht]
\centering
\begin{tabular}{|c|c|c|c|c|}
\hline
 \multicolumn{4}{|c|}{node 5 (subtree \{5,3,1\})} \\
\hline
	order (k) & $\log N(n,k)$ & M-S error & $\log  P(best \mathcal{P})$ \\
\hline
	1 & 0 & 23.86 &-2394.6  \\
	2 & 18.47 & 1.20 &-5539.0 \\
\hline
 \multicolumn{4}{|c|}{node 6 (subtree \{6,2,0\})} \\
\hline
	order (k) & $\log N(n,k)$ & M-S error & $\log  P(best \mathcal{P})$ \\
\hline
	1 & 0 & 42.99 &-751.8 \\
	2 & 18.47 & 0.60 &-5640.3 \\
\hline
\end{tabular}
\caption{Intermediate results of the MP algorithm for the image in Fig. \ref{fig:approach:algorithm:algo}.
 For each internal node (5 and 6) we compute all possible partitions spanned by the corresponding subtree.}\label{table:approach:multipartition:detail}
\end{table}
The expensive part of this algorithm is the computation of the M-S data adjustment term (error) from eq. (\ref{eq:meaningful_regions:region_term:data_term}), which needs to be evaluated for every pixel on a given region. In addition, each pixel can be assigned to many different regions in $\h$, thus the computation is repeated many times for each pixel. One way to avoid this overhead, is to take advantage of the fact that the error (but not the probability) is additive. In this way, we can compute the error of the $\p_3$ composed by regions $\{5,3,1\}$ from the errors of partitions $\{5\}$ and $\{3,1\}$. In this way, each error is computed only once per region.
\subsection{Optimality and complexity}\label{sec:meaningful_partitions:optimality}
Our procedure for computing the probabilities is based on a dynamic programming scheme, and thus computes the globally optimum partition by taking local decisions at each level. It is important to determine if we can achieve the global optimum with this procedure, which is discussed in the following Theorem.\\

\begin{theorem}[Optimality]\label{theo:meaningful_partitions:optimality}

Given the number of regions $N$ of the partition of the image, the recursive computation of probabilities gives the optimum partition.\\

Note that the background model ensures that the probabilities of false alarms are always strictly greater than
zero. Indeed, the  only way to obtain a zero probability is to compute $P(E_R<-\infty)$, but that could not happen because the error $E_R$ is always finite.\\

\begin{IEEEproof}

Suppose A is a node of the hierarchy and B,C are its sons.

(binary case). Let $P_A$ be the partition of A which is optimal for n sets
(the partition is formed by n sets and is optimal between all of the partitions of n sets).

Let $P_A \cap B$ be the partition of B  made of the sets (say $n_1$) of $P_A$ that are in B, and 
$P_A \cap C$ be the partition of C  made of the sets (say $n_2$) of $P_A$ that are in C. Note that
$n= n_1 + n_2$.

Let us prove that $P_A \cap B$ is an optimal partition of B with $n_1$ sets.
Otherwise, it exists a partition of B with $n_1$ sets $P_B$ such that $NFA(P_B) < NFA(P_A\cap B)$.
By replacing $P_A \cap B$ by $P_B$, we have that $NFA(P_B)NFA(P_A\cap C) < NFA(P_A \cap B )NFA(P_A\cap C) =NFA(P_A)$ and this is a contradiction.

Similarly, $P_A \cap C$ is an optimal partition of C with $n_2$ sets.
Thus, given the number of regions $N$ of the partition of the image, the algorithm gives the optimal
partition in the hierarchy with $N$ sets.
\end{IEEEproof}
\end{theorem}
\vskip 11pt
As we are sure that we could reach the optimum partition, we can use this very efficient dynamic programming scheme to compute our probabilities.\\

From the practical point of view, we can also ensure that the probabilities are strictly positive.
 The problem could arise when they very small, because we can reach the precision of the data type and have representation errors leading to an exact zero value.
 However, as we internally work directly computing the logarithms of the probabilities, we circumvent this representation problem and never get an exact zero value.
 \footnote{We can also limit the probabilities using a small threshold $\epsilon > 0$, $P=\max (P,\epsilon)$, or add $\epsilon$ to the probabilities and then normalize them to sum one.}
\\

Regarding the complexity of the algorithm, again it has two parts: constructing the hierarchy and selecting the optimal partition. Assume a $n$-pixel image and a pruned tree of $N$ leaves.
From the analysis of the GREEDY algorithm in Section \ref{sec:meaningful_regions:complexity}, we have a cost of $\Theta (n\log{n})$ to construct the tree.
However, in this case, the selection part is slower, because for every node we need to compute the probabilities of every possible partition spanned by it which has a great number of combinations to explore, because we keep the best partition for every possible level. On the other hand, in the GREEDY case we only need to find a minimum NFA couple at each step.
The analysis of the complexity of the NFA computation is not trivial and it is shown in appendix \ref{sec:appendix:complexity}, as that computation shows the complexity of this part is $\Theta (N^3)$.
If we make no initial pruning the complexity will be $\Theta (n^3)$, which is unusable in practice for medium sized images. If we make a reasonable pruning of $N\approx n^{\gamma}$ with $\gamma = 0.5$, as mentioned in section \ref{sec:meaningful_regions:complexity}, the complexity with be $\Theta (N^{1.5})$. Finally, if we want our algorithm to run fast, we need to prune the tree to $N\approx n^{1/3}$. This could be a reasonable amount of prunning in practice (around 100 regions for a 1024x1024 image), and it is better as $n$ grows. 
If that condition holds, the worst case complexity of this selection could be reduced to $\Theta (n)$.
From this we can conclude that the complexity of the algorithm can be reduced to $\Theta (n\log{n})$ if proper pruning is applied.
\subsection{Boundary term}\label{sec:meaningful_partitions:boundary_term}
In the multi-partition case, our data term is based on a piece-wise constant model, and has no regularization term. For this reason, when faced with images which do not comply with the model, it presents some problems. Mainly, it tends to oversegment almost flat regions with a slow varying intensity gradient. This phenomenon is shown in Figure \ref{im:set_cielab_final:01:detalle:res}.
To overcome this problem, we could add a boundary term just in the same way we did with our previous algorithm, in section \ref{sec:meaningful_regions:boundary_term}. That is, instead of computing the probability $P(R1)=P(E_{R1}<\hat{E}_{R1})$, we could add another term measuring the meaningfulness of the boundary so that $P(R1)=P(E_{R1}<\hat{E}_{R1}).P(L_{R1}<\hat{L}_{R1})$, where $\hat{L}_{R1}$ is the length of the boundary of region $R1$ computed with Eq. \ref{eq:meaningful_regions:boundary_term:length}.
However this introduces the need of considering the relative importance of each term, which in turn leads us to introduce some kind of weighting parameter.
In addition, due to the way probabilities are constructed, we were not able find a definition of this term holding the decreasing property needed for a proper scale-climbing algorithm. 
For this approach to work, the length term should be always decreasing with scale, and our term is decreasing in almost every case, so we can not guarantee this property.

For these reasons, we propose the use of the boundary term as a post-processing, to refine the partition selected by the data term.
Thus after the best partition is selected, as shown in section \ref{sec:meaningful_partitions:algorithm}, we test every edge separating each pair of neighboring regions and if the boundary between them is not meaningful, we merge them into a single region.

So far, we have shown examples of the MP algorithm using the data term only and, as we have mentioned before, in images like \emph{Church} we see that our algorithm tends to oversegment large uniform regions, which are clearly far from the piecewise constant model. In Figure \ref{im:set_cielab_final:01:detalle} we show how the boundary term solves this problem and gives a more accurate segmentation.
%
\renewcommand{\anchodos}{3.5cm}
\renewcommand{\resDir}{im}
\renewcommand{\resSubDir}{set_cielab_final}
\renewcommand{\figDir}{im/set_cielab_final}
\begin{figure}[ht]
  \centering
  \subfloat{ \label{\resDir:\resSubDir:01:detalle:orig}%
  \includegraphics[width=\anchodos]{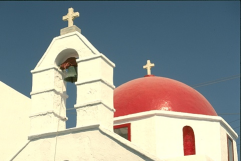}%
  }%
  \subfloat{ \label{\resDir:\resSubDir:01:detalle:init}%
  \includegraphics[width=\anchodos]{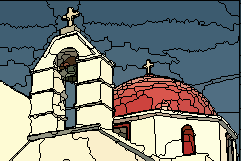}%
  }\\
  \subfloat{ \label{\resDir:\resSubDir:01:detalle:res}%
  \includegraphics[width=\anchodos]{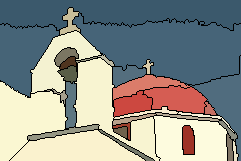}%
  }%
  \subfloat{ \label{\resDir:\resSubDir:01:detalle:bordes}%
  \includegraphics[width=\anchodos]{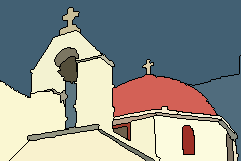}%
  }%
  \caption{Results of the multipartition algorithm on the \emph{Church} image. \protect\subref{\resDir:\resSubDir:01:detalle:orig} Original Image $N=38801$. \protect\subref{\resDir:\resSubDir:01:detalle:init} Initial regions after pruning with $\lambda=10$, $M=628$  \protect\subref{\resDir:\resSubDir:01:detalle:res} Segmentation results using the data term only $N_d=36$. \protect\subref{\resDir:\resSubDir:01:detalle:bordes} Segmentation results after adding the boundary term $N_c=29$.}
  \label{\resDir:\resSubDir:01:detalle}
\end{figure} 
\subsection{Extension to multi-valued images}\label{sec:meaningful_partitions:extension_to_color}
So far, we have presented our algorithms for gray-level images. In this section we will extend them to color images, and to the more general case of vector-valued images.
In this case, an image will be a vector valued function $I:\om \rightarrow R^d$.
We start by rewriting the M-S data term in \req{eq:meaningful_regions:region_term:data_term} as
\begin{equation}\label{eq:meaningful_regions:color:data_term}
E_R=\sum_{x \in R}d(I(x),\mu_R),
\end{equation}
where $\mu_R \in R^d$ and $d(x)$ is any monotone function of a norm in $R^d$.
In addition, this formulation leaves the door open to see this term as the goodness of fit of the image value $I(x)$ to the region $R$, which leads us to the more general formulation %
\begin{equation}\label{eq:meaningful_regions:color:data_term:general}
E_R=\sum_{x \in R}p(I(x)|R),
\end{equation}
where $p(.)$ is the probability that pixel $x$ belongs to region $R$. This formulation gives us a general and flexible way to improve our results by choosing different region models and metrics.
In the particular case of color valued images ($d=3$), to complete our model we need to choose a color space and a suitable metric in that space. Two important requirements for this space are to have a clear separation between color and intensity channels and to be coherent with human perception. For this reason, we choose the CIELab color space, in which equal color differences approximate equal differences perceived by a human observer, thus making possible to compare colors using an Euclidean norm.
For this reason, to compute the error term we choose $d(x)$ as the squared euclidean norm in $R^3$.
\subsection{Color-valued gradient}\label{sec:meaningful_partitions:color_gradient}
The final step on the extension to the multi-valued case, is to extend the boundary detection algorithm of Cao et al. \cite{segm:cao:03:extracting_meaningful} to work with vector images. The notion of gradient for multi-channel images is not well defined; however, there are some works in the literature that try to define similar notions.
In this work we follow the ideas of Di Zenzo \cite{reg:aniso:zenzo:86:vector_geometry}, which proposes to compute
the structure tensor and obtain its eigenvalues. In this setting, the largest eigenvector represents the direction of maximum change, and its corresponding eigenvalue the magnitude of that change. As we are interested in finding contrasted boundaries, we will compute the probability of the minimum value of the magnitude of the largest eigenvalue along the curve as we did with the gradient in the scalar case.
\subsection{Parameter choice}\label{sec:meaningful_partitions:parameter}
The algorithm, as presented in previous sections, does not have any free parameters. For that reason, it does not allow us to select the desired scale. Without any parameters, it works at a fixed scale which usually gives oversegmented images. This is explained by the collaboration of two phenomena. First, we have a lower bound on the number of tests that is a very loose estimate. Second, as we are using the approximation of independence between pixels to use the CLT, in many cases we obtain lower bounds of the joint probability by factorizing it into independent terms. These two approximations have the effect of making finer partitions more meaningful than coarser ones.

Summarizing, the main problem of the algorithm is to find the correct scale of an image. From a practical point of view, it would be interesting to have a way to select the optimal scale given a set of examples. We recall from section \ref{sec:meaningful_partitions:number_of_mp}, that our lower bound for the number of partitions was:
$N(n,k) \geq \Theta(n^{6(k-2)})$. We have simulated this bound for the $k=2$ case and varying $n$ and we have found that the difference between the actual quantity and the bound increases with $n$. As the number of tests is related to the complexity of the partition, underestimating this quantity will lead the algorithm to oversegment partitions.\\
With this in mind, we propose to add a simple correction of the form: $N(n,k) \geq O(n^{\alpha (k-2)})$, where parameter $\alpha$ is introduced as an exponent to $n$ to reduce the increasing gap between both quantities. This gives us the flexibility to adjust the $\alpha$ parameter to match the correct scale, this parameter is tuned according to the dataset used and the application. 
The value $\alpha=6$ is the theoretical estimate of the lower bound of the number of tests and it will be called the \emph{default} value of the parameter.

Tuning parameters is a difficult and time consuming task. For the sake of simplicity for the end user, we provide here an automatic way to determine $\alpha$ for a given dataset. We achieve this by optimizing a simple cost function given by the difference of the number of regions detected by the algorithm and by human subjects. Thus, given a parameter value $\alpha$, which gives the segmentation $Q_i$ for image $i$ and a  ground truth segmentation $P_i$, the error function is:
$$E(\alpha)=\sum_i^N(\#P_i-\#Q_i)^2$$
Then we perform a simple empirical error descent on this cost function, and find the \emph{optimal} value for $\alpha$.
In section \ref{sec:results:parameter_learning} we will discuss how to find suitable values for this parameter for the database used in the experiments, how the algorithm behaves with different values of $\alpha$ and how it compares with the automatically tuned value.
In the following, the variant of the algorithm using the default value will be called MP-D, and the variant using the modified  $\alpha$  will be called MP-M. When $\alpha$ is selected automatically, we denote the algorithm as MP-S (S stands for 'supervised').
\section{Results}\label{sec:results}
\renewcommand{\figDir}{im/quantitative/set_cielab_7_final_verif/out}
\renewcommand{\figDirB}{im/quantitative/trained_7_verif/out}
%
\renewcommand{\resDir}{im/quantitative/bsds300/images}
\renewcommand{\anchocinco}{3.5cm}
\begin{figure*}[ht]
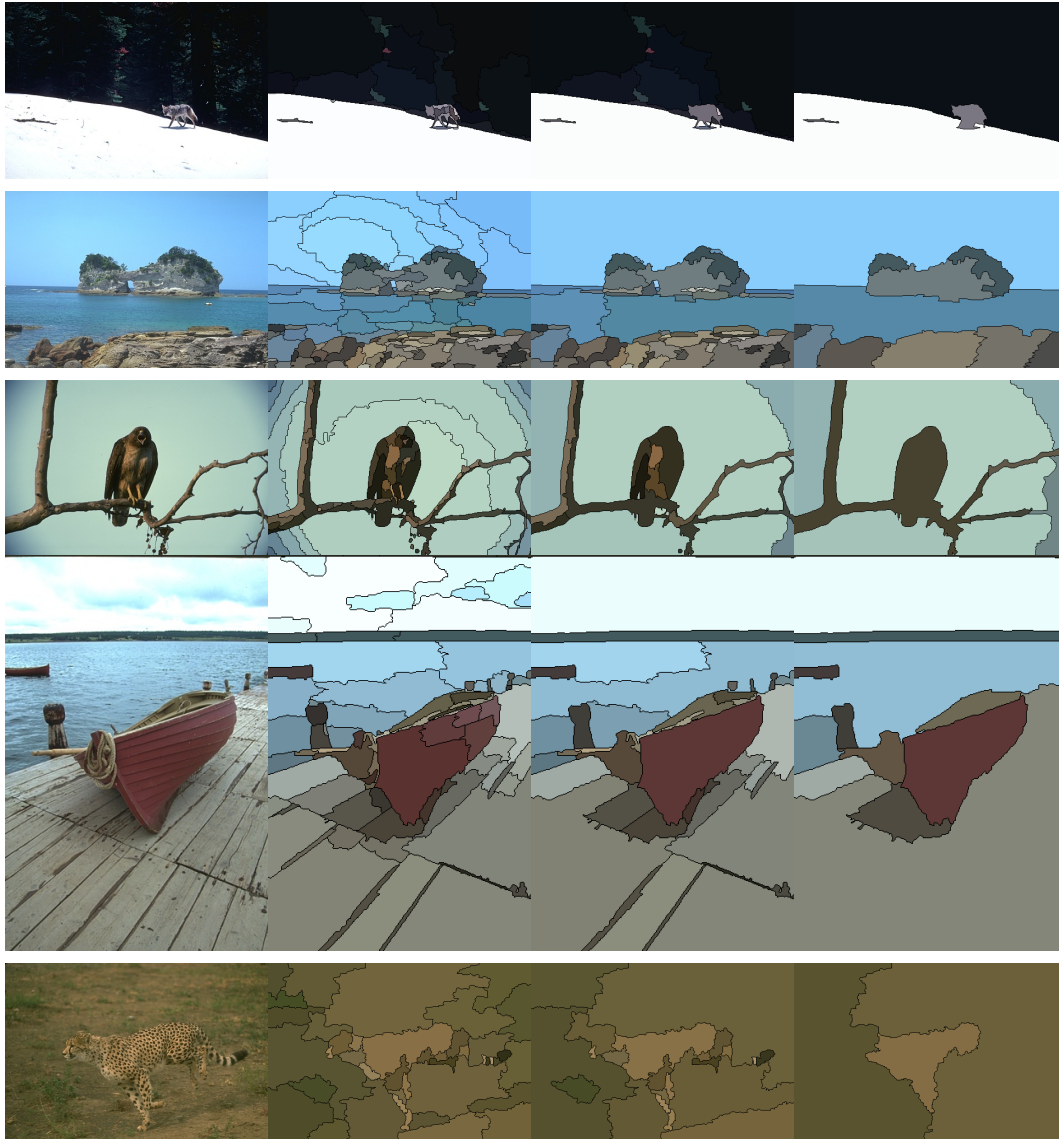

  \centering
  \figrowB{167062}
 \vskip 5pt
\figrowB{249061}
 \vskip 5pt
  \figrowB{42049}
  \figrowB{138078}
  \vskip 5pt
  \figrowB{134008}
  \caption{Results of the algorithm over real images extracted from the BSDS300 (images numbered  167062, 249061, 42049, 138078, 134008 respectively).
   From left to right: original image, initial partition, segmentation using default parameter ($\alpha=6$), segmentation using a manually tuned parameter ($\alpha=50$).
   In the last three images of each row, segmented regions are filled with their mean color and the boundaries are overimposed in black. The corresponding number of detected regions for each image is shown in Table \ref{table:results:various_bsds300:nb_regions}}
  \label{fig:results:various}
\end{figure*}
\begin{table}[t]
\centering
\begin{tabular}{|c|c|c|c|c|}
\hline
	image & \multicolumn{3}{c}{partition sizes}\\

\hline 
				& initial 	&  MP-D 	& MP-M  	\\
\hline
	wolf (167062)		& 39		& 19		& 4	 	\\
	sea (249061)		& 135		& 62		& 11		\\
	bird (42049)		& 102		& 28		& 10		\\
	boat (138078)		& 91		& 43		& 14		\\
	cheetah	 (134008)	& 41		& 22		& 3		\\
\hline
\end{tabular}
\caption{Number of detected regions corresponding to images shown in Fig. \ref{fig:results:various}.}
\label{table:results:various_bsds300:nb_regions}
\end{table}
\subsection{Real Images}\label{sec:results:real}
In the first place we present here qualitative results of the MP algorithm. In Figure \ref{fig:results:various} we show the results of the algorithm on a few selected images of the Berkeley Segmentation Database (BSDS300 subset) \cite{martin:2001:database_of_human_segmented}, with different values of the parameter. As this figure shows, the results are quite good even with the great variability presented by the images. However, we also observe that using the default parameter value, our algorithm tends to give oversegmented results in most of the cases. In addition, these illusory boundaries are not completely removed by the boundary term.
However, when adjusting the scale parameter the results are greatly improved and the results are visually pleasant.
Finally, as in this example we are not using texture descriptors, in natural images presenting a \emph{camouflage} effect (last row of Fig. \ref{fig:results:various}), the algorithm performs poorly regardless of the choice of the parameter.\\

\begin{figure*}[ht]
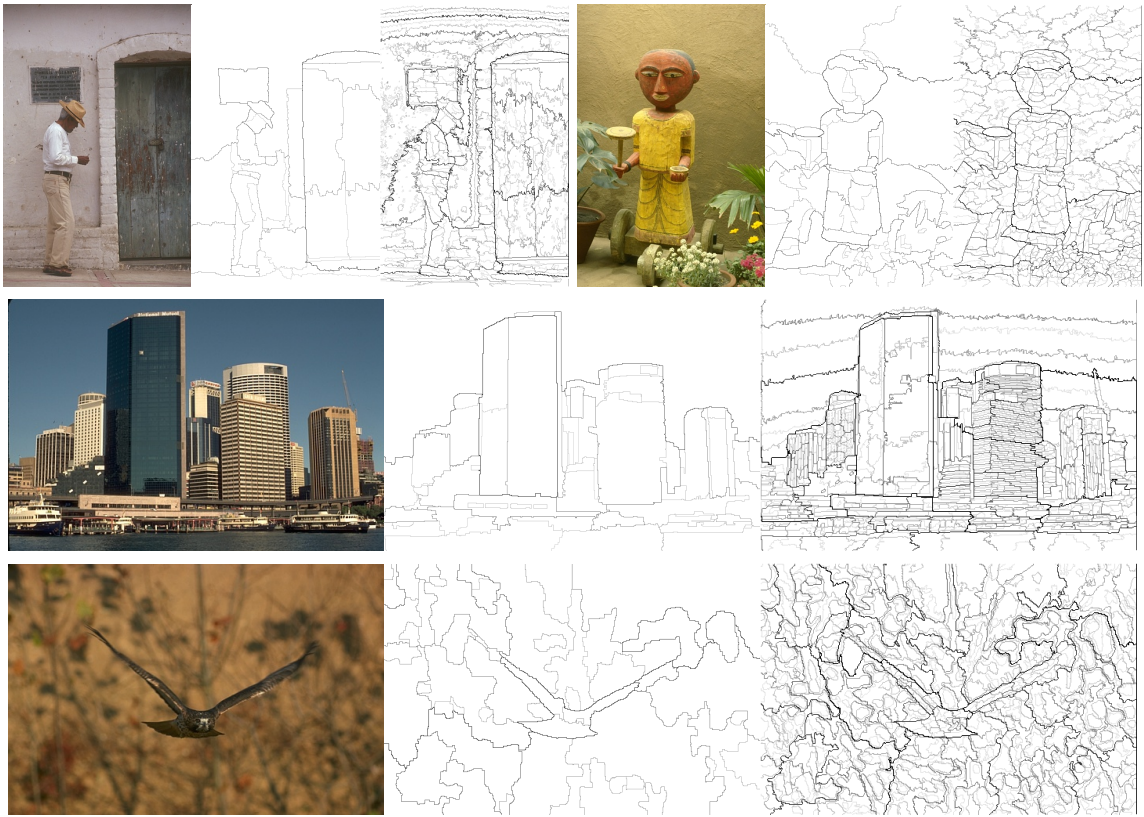

  \centering
\figRowMultiscaleV{64061}%
\figRowMultiscaleV{71076}%
\vskip 5pt
\figRowMultiscaleH{69007}
\vskip 5pt
\figRowMultiscaleH{70011}
\vskip 5pt
  \caption{Results of the algorithm over selected images of the BSDS500 (test set). For each example we show the original image on the left and a saliency map of the MP algorithm on the middle.
  The value of each contour corresponds to the scale of appearance  $\alpha^-$ of each region. On the right images we show a pseudo saliency map for the Guigues \cite{segm:graphs:guigues:06:scale_sets_image_analysis} algorithm.
  In this case, each contour is labeled with the number of times the contour appears in the hierarchy.}
  \label{fig:results:multiscale:saliency}
\end{figure*}

\begin{figure*}[ht]
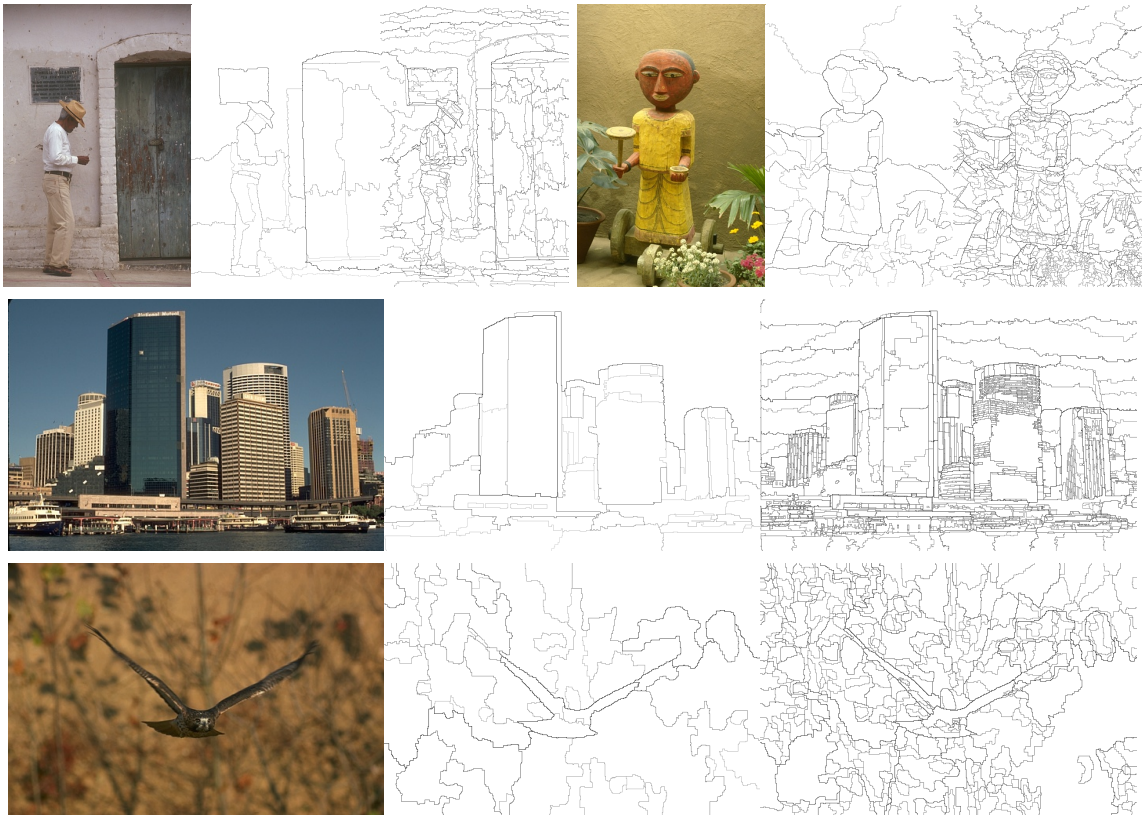

  \centering
\figRowMultiscaleVV{64061}%
\figRowMultiscaleVV{71076}%
\vskip 5pt
\figRowMultiscaleHH{69007}
\vskip 5pt
\figRowMultiscaleHH{70011}
\vskip 5pt
  \caption{Results of the algorithm over selected images of the BSDS500 (test set). For each example we show the following images. Left: original image. Middle: the saliency map of the MP algorithm. Right: saliency map of the initial hierarchy.
  For the MP algorithm the value of each contour corresponds to the scale of appearance  $\alpha^-$ of each region.
  For the initial hierarchy the value of each contour corresponds to the scale of appearance  $\lambda^-$ of each region (where $\lambda$ is the scale parameter in the M-S model).  
}
\label{fig:results:multiscale:saliency:initial}
\end{figure*}

In Figure \ref{fig:results:multiscale:saliency} we show results over real images taken from the test set of the BSDS500 database.
For each image, we show the corresponding saliency map which serves to asses the quality of the hierarchical segmentation obtained.
The saliency map is constructed by labeling each contour with its scale of disappearance $\alpha^-$, which is the maximum scale at which a region is present in the output partition.
These results show two important properties. First, the obtained stack of partitions are causal, i.e. as the scale parameter increases the segmentation gets coarser.
This can be verified by observing the values of $\alpha^-$ assigned to each contour.
If a contour has a high $\alpha^-$ value it means that it will survive for a long time in the merging process. On the other hand, small values mean that it will be removed at early stages.
As the figure shows, the highest values of $\alpha^-$ are assigned to the most coarser segmentation.
And second, the stack of partition is geometrically accurate, i.e. the borders present in each scale are a strict subset of the previous (finer) scale. If they were not included they will be seen as new contours in the saliency map.
These results empirically show that the obtained hierarchy holds the strong causality property.
\renewcommand{\figDir}{im/quantitative/set_cielab_7_final_verif/out}
\renewcommand{\figDirB}{im/quantitative/trained_7_verif/out}
\renewcommand{\figDir}{im/quantitative/bsds500/images}
\renewcommand{\anchocinco}{2.5cm}
\renewcommand{\anchocuatro}{5cm}
In Figure \ref{fig:results:multiscale:saliency:initial} we show a comparison of the hierarchy of partitions generated by the MP algorithm against the initial hiearchy over the same set of real images.
The results obtained by the MP algorithm are similar to those from the initial partition, with the difference that less salient boundaries are detected in the slowly varying regions.
This is mainly due to the introduction of the boundary term weighted by the gradient. An important property of this model, is that even with the smallest possible parameter value, the finest possible partition in the initial hierarchy can not be obtained.
This is due to the regularizing effect of the CLT approximation of the probabilities, which favors partitions with bigger regions.
\renewcommand{\figDirB}{im/quantitative/bsds500/images/initial}
\renewcommand{\figDir}{im/quantitative/bsds500/images}
\renewcommand{\anchocinco}{2.5cm}
\renewcommand{\anchocuatro}{5cm}
\subsection{Noise levels \& threshold analysis}\label{sec:results:noise}
In Figure \ref{fig:results:noise_simulation} (top) we show the behavior of the algorithm under different noise powers, on a 1000x1000 gray level version of Fig. \ref{fig:approach:algorithm:algo}. The regions have the following means: 50,100,150,200. Eight  test images were created, with a zero mean gaussian noise of power $\sigma$ added. The lower (logarithmic) bound for the number of tests has the same value of $165.8$ for all cases. As we are modeling regions by their means, when we increase noise power the adjustment to that model becomes worse, and thus the meaningfulness of the selected partition is lower. However, the algorithm is able to select the correct partition (4 regions) in each case, even under heavy noise conditions.

 In most \emph{a contrario} approaches there is a notion of detection: if the meaningfulness is lower than a threshold, they detect objects and if it is greater they give no detections. On the contrary, our algorithm always selects a partition, and it does not have a notion of no-detection. This happens because we are not thresholding the NFAs of each partition, but selecting the one with lower NFA. As Table \ref{table:approach:multipartition} shows, all the partitions are meaningful which means that our threshold on the NFA is not well adjusted. This is a common problem in \emph{a contrario} approaches and was studied in detail in \cite{grompone:2009:on_computational_gestalt_detection_thresholds}. Another problem with this approach has to deal with the size of the images: if we have two images with the same content and different sizes, the NFA values will change due to the use of the CLT and with bigger images all structures will be more meaningful.
 In Figure \ref{fig:results:noise_simulation} (middle and right) we show an experiment with an image composed of one region with a mean of 128. In addition, Gaussian noise with different powers was added, and two different image sizes where considered. As the Figure shows, the values for a bigger image are more extreme: meaningful partitions become more meaningful and non-meaningful ones become even less meaninfgul. However, our approach does not attempt to say if an individual partition is meaningful or not, but tries to get the partition that better explains the data in a relative way to the other partitions, so in the end, we are immune to these problems.
\renewcommand{\anchotres}{5.3cm}
\begin{figure*}[t]
  \centering
    \includegraphics[width=\anchotres]{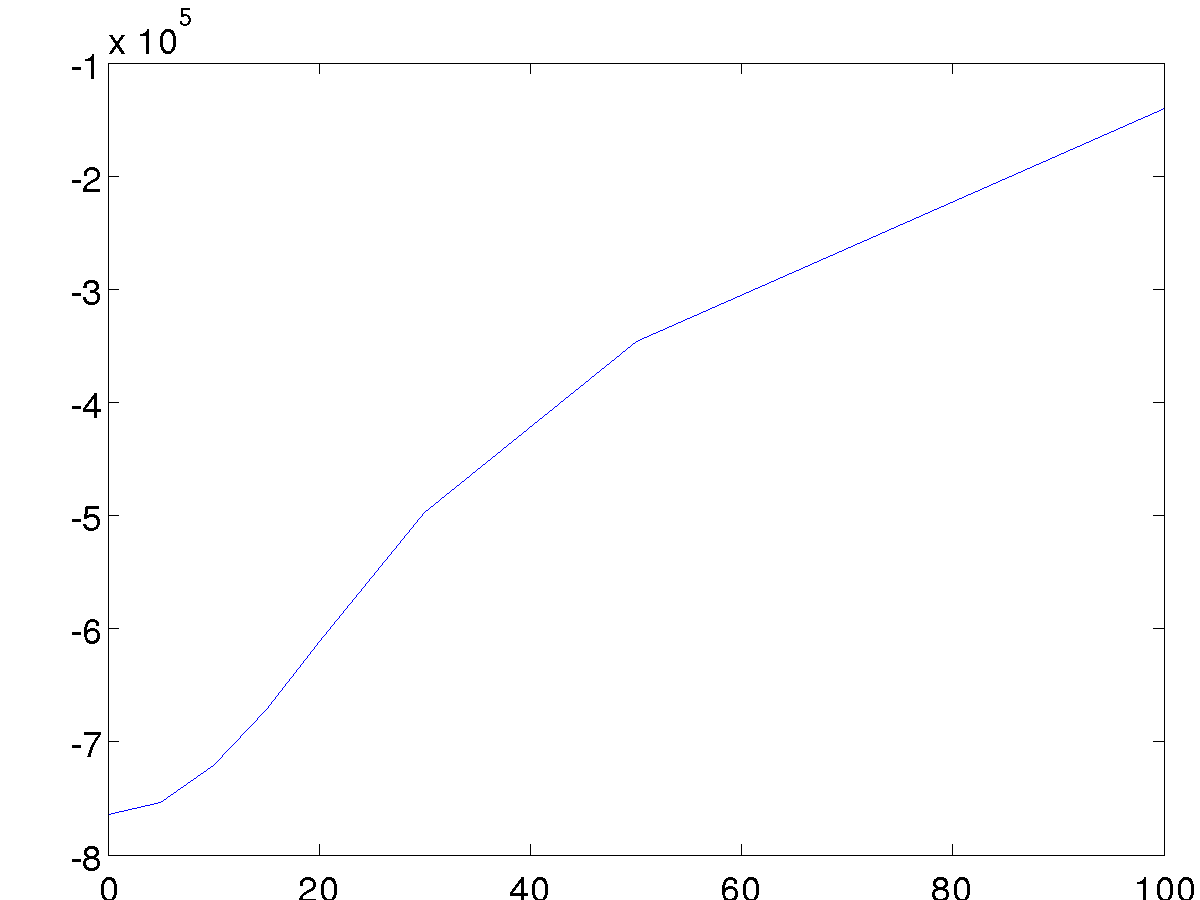}%
  \includegraphics[width=\anchotres]{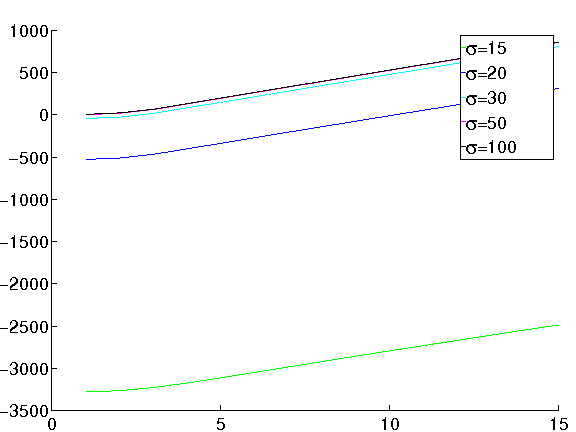}%
  \includegraphics[width=\anchotres]{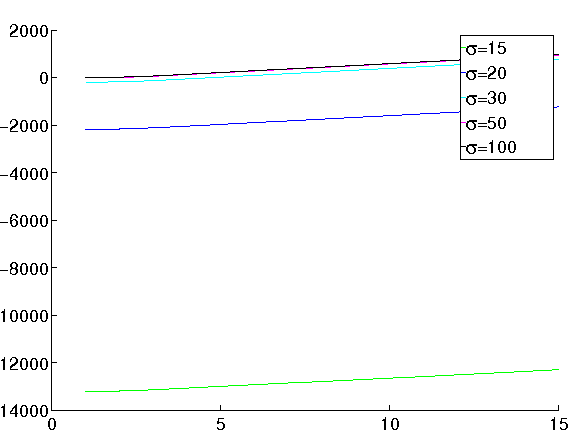}
  \caption{Left: LNFA of the best partition (4 regions) selected against noise power. Middle and Right: Effects of the size of the images and noise power on the NFA.
  In each image the LNFA is plotted against the number of regions of the partition, for two image sizes (250x250 and 500x500 in middle and right images respectively). $\sigma$ is the power of the added noise. The color codes green, blue, cyan, magenta and black, correspond to the increasing $\sigma$ values of 15,20,30,50,100 respectively. }
  \label{fig:results:noise_simulation}
\end{figure*}%
\subsection{Performance metrics}\label{sec:results:metrics}
In order to compare our results against human segmentations, we need suitable metrics to compare two different partitions of the same image. 
There are many metrics proposed in the literature, but most of them are heuristic and lack theoretical background and validation.
One of the most interesting works in this area is by Martin et al. \cite{martin:2003:empirical_approach_to_grouping}, where they built the BSDS and proposed some metrics consistent with human perception. In a posterior work Cardoso et al. \cite{cardoso:2005:toward_a_generic_evaluation} introduced a new set of metrics which aim to improve Martin's proposal, while maintaining the same basic idea.
One way to evaluate the performance of a segmentation algorithm is to model the (qualitative and quantitative) properties of human made partitions and then verify if a given segmentation has these properties. On the contrary, rather than model human segmentation, these works compare two given segmentations of the same image in a way which is consistent with our perception.

In the following we will briefly review Martin's work \cite{martin:2003:empirical_approach_to_grouping}. One key concept introduced is the notion of mutual refinement. Given two partitions P and Q, P is a refinement of Q if for every region $R_i \in P, \exists S_i \in Q$ such that $R_i \subset S_i$. On the other hand, P and Q are mutual refinements if $R_i \cap S_i$ is either equal to $R_i$ or $S_i$ or it is the empty set. This means that locally, either P refines Q or viceversa.

One of the conjectures in Martin's work is that human segmentations of the same scene are mutual refinements of each other, and that all possible human segmentations of a scene could be organized in a perceptual tree.
Thus, each human made partition is just a selection of regions among this tree. It is important to note that this tree induces a notion of \emph{scale}, not in the scale-space sense, but regarding the granularity of the partition.
Put in this way, each human segmentation differs from each other in the scale selected. However, this scale notion is somehow loose, because it is not global but local.
The conclusion from these observations is that differences due to mutual refinements are perceptually less important than those produced by misplaced regions.

Cardoso et al. (\cite{cardoso:2005:toward_a_generic_evaluation}) propose three different metrics to quantitatively evaluate results. We only give here an intuitive idea about how they work, for a detailed definition please refer to \cite{cardoso:2005:toward_a_generic_evaluation}.
Given two partitions $P$ and $Q$ to be compared, the following quantities are defined:
\begin{itemize}
\item \verb+SPD(P,Q)+: this quantity roughly reflects the percentage of incorrectly classified pixels. It does not tolerate mutual refinements. 
\item \verb+APD(P,Q)+: same as SPD but it does not count errors on pixels where P is a refinement of Q.
\item \verb+MPD(P,Q)+: same as SPD but it does not count errors on pixels where P and Q are mutual refinements.
\end{itemize}
These metrics are bounded between 0 and 1, and they count errors; thus small values mean good performance and viceversa. It is also important to note that SPD and MPD are symmetric w.r.t. their arguments, while APD is not. In the following we will denote the human partition as $P$ and the algorithm result as $Q$; and we will call this set of metrics Partition Distances (PD).

Although these metrics are not additive, $SPD$ should roughly reflect the total amount of error, which in turn is related to the sum of $APD(P,Q)$  and $APD(Q,P)$.
This means that two partitions with the same $SPD$ have the same performance, and the differences between the $APDs$ show us the balance between oversegmentation and undersegmentation.
However, this interpretation alone could be misleading, because $SPD$ includes errors of different nature. For instance, if a region in the human partition is divided in two regions in the computer segmentation, for the $SPD$ this error counts the same as if the region was completely misplaced.
Even between human observers (see \cite{martin:2003:empirical_approach_to_grouping}) partitions are not consistent in terms of an exact matching, but rather in terms of mutual refinements.
For this reason, global performance could be measured with the $SPD$ but the value of $MPD$ should be verified too.
While it is tempting to summarize both measures into only one, we believe that this would be too simplistic and we would lose the richness of the information of these metrics. For this reason our measure of overall performance will be a qualitative interpretation of both $SPD$ and $MPD$.
\subsection{Quantitative evaluation}\label{sec:results:quantitative}
Most of the images used in this paper for the validation process come from the BSDS, which is a subset of 1200 images from the Corel Database, with human made segmentations available. Each input image was manually segmented by at least 5 different subjects. One of the publicly available datasets, the BSDS300 is composed of 300 images of 481x321 pixels, divided into train and test subsets of 200 and 100 images respectively. Ground truth segmentation is provided for all images. 
The instructions given to the subjects were not to identify objects but uniform regions on the images, which makes this database well suited for low-level image segmentation.
In order to test the independence of our results from the BSDS300, we also conducted experiments with the Lotus Hill Institute (LHI) database \cite{Yao:2007:Introduction_to_a_large_scale}, which is a publicly available database with a ground truth labeling of the objects present in the image, mainly designed for object recognition tasks.
Recently, a new version of the BSDS was released under the name of BSDS500, which adds 200 new test images, along with a new set of benchmarks. In the final part of this section, we also present results using this database.

We present here just a brief excerpt of the extensive validation conducted, full results and detailed analysis of each algorithm can be found at \url{http://iie.fing.edu.uy/rs/wiki/ImageSegmentationAlgorithms}. We have also made available the source code and an online demo to run the algorithm via web. We developed a new benchmark based on PD metrics, and the code to run it is also available online.
\begin{table}[t]
\centering
\begin{tabular}{|c||c|c|c|c|c|}
\hline
 alg. & \tiny{$SPD$} & \hspace{-7pt} \tiny{$APD(P,Q)$} \hspace{-7pt} & \hspace{-7pt} \tiny{$APD(Q,P)$} \hspace{-7pt} &\hspace{-7pt} \tiny{$MPD$}  \hspace{-7pt}& n\\ 
\hline
	\hspace{-6pt} initial($\lambda \!\!=\!\! 50$) \hspace{-6pt}	&	& 	  	&	   &	   & 222.5 \hspace{-7pt}\\
	\hspace{-6pt} MP-D($\alpha \!\!=\!\! 6$) \hspace{-6pt}	& 0.56	& 	0.54  &	 0.09  &	 0.07  & 63.8\\
	\hspace{-6pt} MP-M($\alpha \!\!=\!\! 50$)\hspace{-6pt}	& 0.40	&	0.32	 &	0.17   &	0.10 & 16.0 \\
\hline
\end{tabular}
\caption{Quantitative evaluation of the proposed algorithms over the BSDS300 using PD metrics. The number $n$ is the average number of detected regions for each algorithm.}
\label{table:results:quantitative:cardoso:best}
\end{table}
In Table \ref{table:results:quantitative:cardoso:best} we show the average value of the performance metrics over all the images in the BSDS for the MP algorithm with two different parameter values. The small MPD value shows that most of the partitions given by the algorithm are coherent with the human segmentation, meaning that we get only few overlapping regions that are not refinements, and thus our result is a mutual refinement of the human segmentation. Moreover, as we have seen in the qualitative evaluation, the MP-D algorithm tends to over-segment. This conclusion is reinforced by the small value of $APD(Q,P)$, which indicates that there are very few under-segmented regions. On the other hand, the big value of  $APD(P,Q)$ also indicates that the amount of over-segmented regions is high.
In addition, the similarity between $SPD$ and $APD(P,Q)$ and the small value of $MPD$ tell us that most of the errors come from choosing the incorrect scale (finer) rather than having misplaced regions.

When we increase the parameter value $\alpha$, we see that the number of false positives ($APD(P,Q)$) is reduced and the number of false negatives is increased ($APD(Q,P)$). In addition, the number of detected regions is lower, which suggests that $\alpha$ is acting as a scale parameter.

\begin{table}[t]
\centering
\begin{tabular}{|c||c|c|c|c|}
\hline
alg. & \tiny{$SPD$} & \tiny{$APD(P,Q)$} & \tiny{$APD(Q,P)$} & \tiny{$MPD$}\\ 
	\hline
\multicolumn{5}{|c|}{Results over the complete database} \\	
\hline
	MP-D ($\alpha=6$) & 0.56 & 	 0.54  &	 0.09  &	 0.07 \\
	MP-G ($\alpha=6$) & 0.49 & 0.43	   &	0.16   &	0.12  \\
	GREEDY & 0.57 & 0.54 & 0.12 & 0.10 \\
\hline
 \multicolumn{5}{|c|}{Results over the 100 test images } \\
\hline
	MP-D ($\alpha=6$ )& 0.56 & 	 0.54  &	 0.10  &	 0.08 \\
	MP-S ($\alpha=21$) & 0.46 & 	 0.41  &	 0.13  &	 0.09 \\
	MP-M ($\alpha=50$) & 0.41  	& 0.31 & 	 0.19  &	 0.11 \\
	MP-M ($\alpha=150$) & 0.38  	& 0.18 & 	 0.28  &	 0.09 \\
\hline
	GREEDY & 0.58 & 0.54 & 0.13 & 0.10 \\
	MP-G ($\alpha=6$) & 0.50 & 0.45	   &	0.17   &	0.13  \\
	MP-G-S ($\alpha=21$) & 0.42 & 0.31	 &	0.23   & 0.14	  \\
\hline
\end{tabular}
\caption{Quantitative evaluation of the proposed algorithms over the BSDS300 using using PD metrics.}\label{table:results:quantitative:cardoso:test}
\end{table}
In Table \ref{table:results:quantitative:cardoso:test}, we show a comparison of performance among all the presented algorithms. We can see that the MP algorithm outperforms the GREEDY algorithm, improving almost every performance metric in around 2\%. However, as the MP algorithm uses color information, in order to make the comparison fair we have made a gray-level variant of it called MP-G.

The first point to be noticed is that the gray level version of the algorithm presents a more balanced result in terms of the SPD and MPD, but the overall performance is slightly worse than the MP color case, because it presents higher values for both SPD and MPD. However, the difference is not as big as we could have expected. One of the main problems of the MP algorithm is oversegmentation, and since MP-G does not use color information, all regions are more similar and we expect the algorithm to make more mergings and give coarser partitions. In this sense, discarding color information diminishes oversegmentation and thus the results are better than expected. 

There are two remarkable points in the results of the comparison between the MP-G algorithm and the GREEDY one: first, the MP-G algorithm shows overall better performance, showing a considerable improvement in SPD and a lower decrease in MPD; second, and probably more important, these results were computed by tuning the $\alpha$ parameter of the GREEDY algorithm and without tuning any parameter in the MP-G case.
These two observations together show the importance of validating complete partitions rather than single merging decisions.

Up to this point, we have shown only average values over the BSDS300 database, but it is also interesting to show how these quantities are distributed over the database. In Figure \ref{im:set_cielab_final:quantitative:multipartition} we show such distributions for each measure. This figure confirms the conclusions drawn from the average values shown before. The SPD is concentrated towards high values, while the MPD lies around small values. The shift to the left of the values of $APD(P,Q)$ shows that the algorithm with this paramer value has very few undersegmented images. On the contrary, the higher values of $APD(Q,P)$ shows that we have many oversegmented images. 

In addition, in the first row of Figure \ref{im:set_cielab_final:quantitative:regions} we show the distribution of the number of regions detected by human subjects and by the MP algorithm, which clearly confirms that our algorithm with the default parameter value is oversegmenting.%
%
%
\renewcommand{\figDir}{im/quantitative/set_cielab_7_final/plot}
\renewcommand{\anchocuatro}{3.8cm}
\begin{figure}[t]
  \centering
  \subfloat{ \label{im:set_cielab_final:quantitative:spd}%
  \includegraphics[width=\anchocuatro]{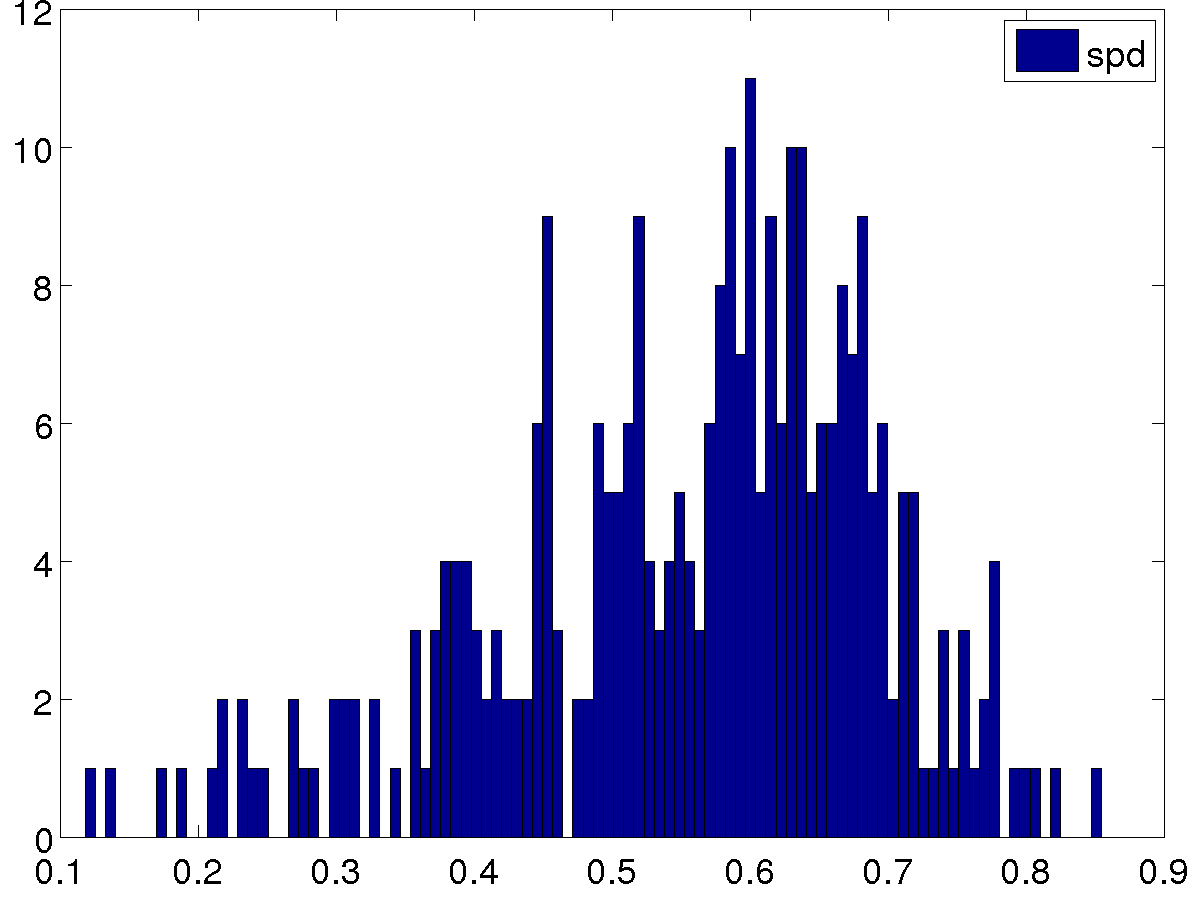}%
  }%
  \subfloat{ \label{im:set_cielab_final:quantitative:apd1}%
  \includegraphics[width=\anchocuatro]{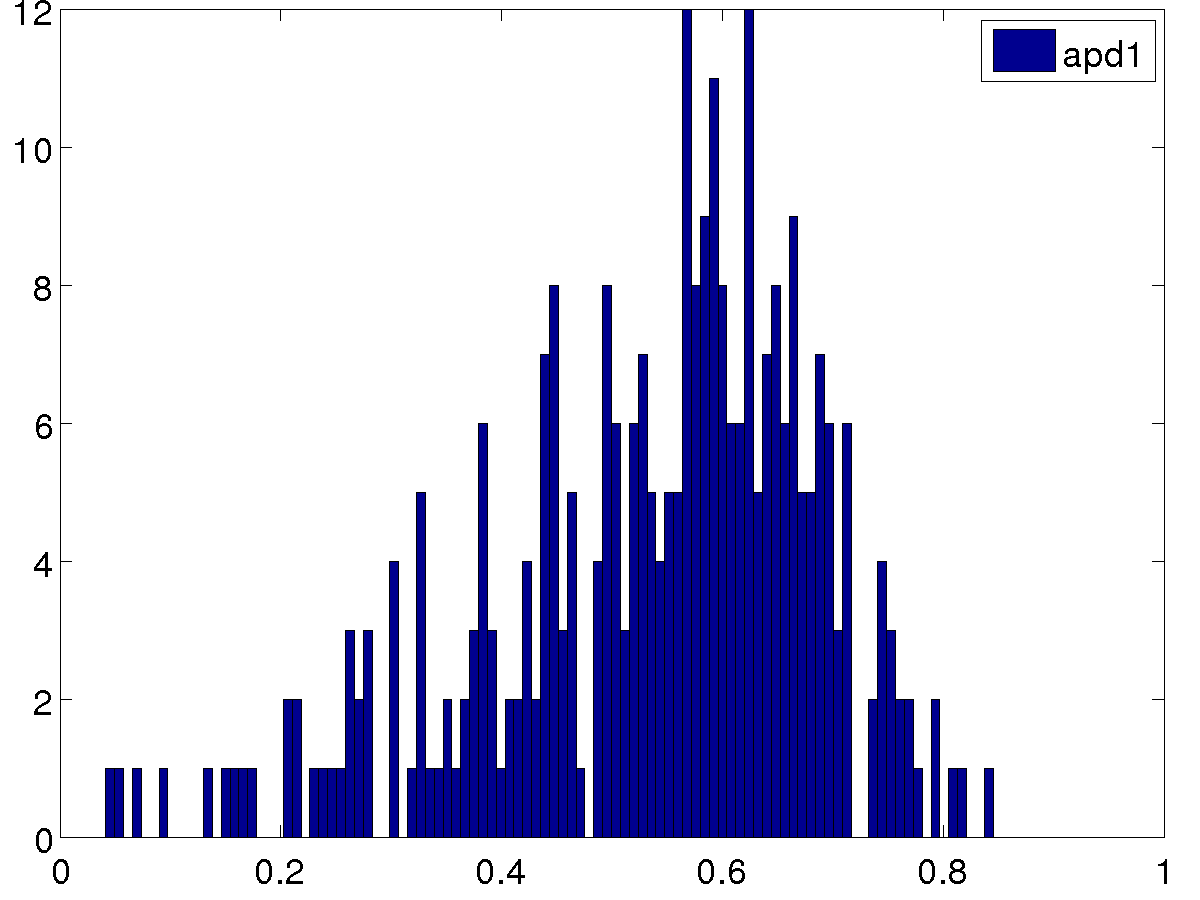}%
  }\\
  \subfloat{ \label{im:set_cielab_final:quantitative:apd2}%
  \includegraphics[width=\anchocuatro]{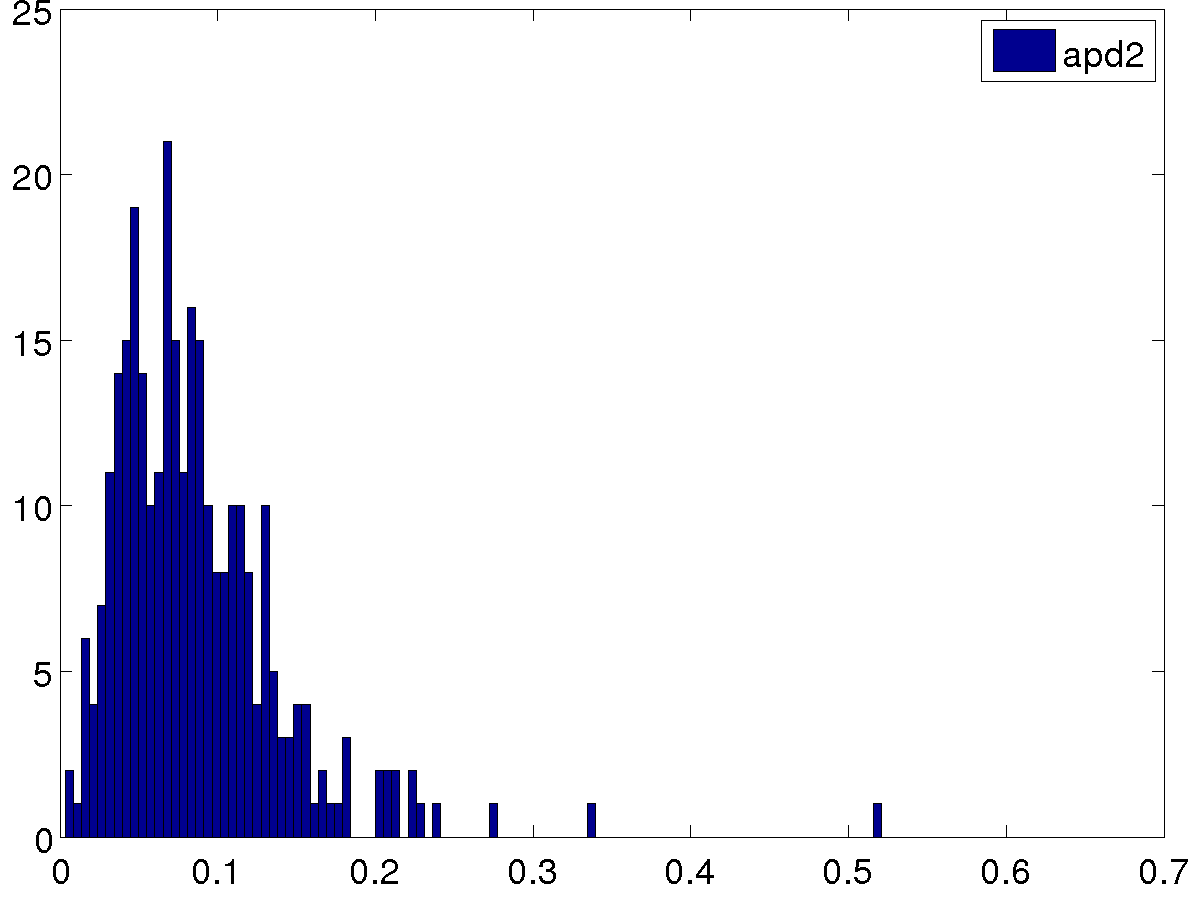}%
  }%
  \subfloat{ \label{im:set_cielab_final:quantitative:mpd}%
  \includegraphics[width=\anchocuatro]{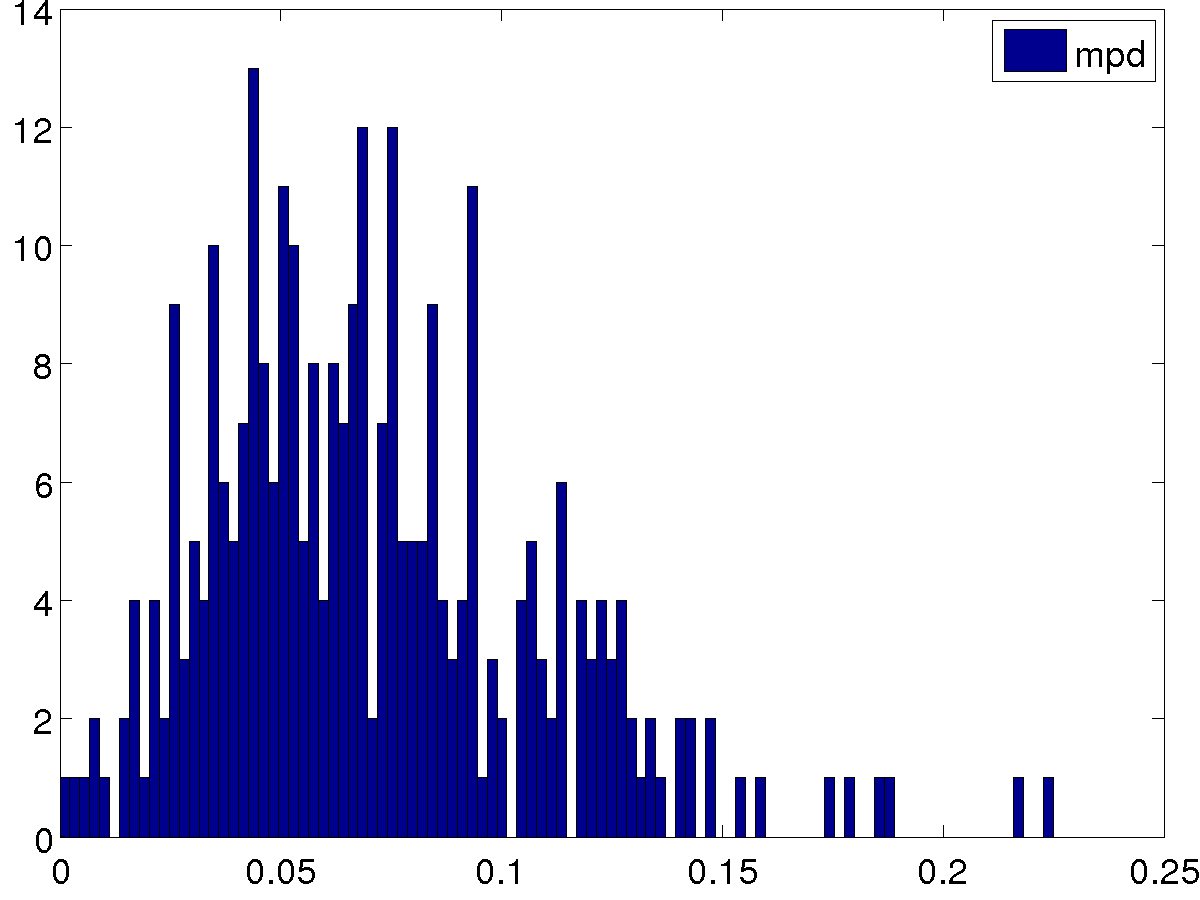}%
  }%
  \caption{Distribution of performance measures over the BSDS300 database. From left to right, and from top to bottom: $SPD$, $APD(P,Q)$, $APD(Q,P)$, $MPD$.}
  \label{im:set_cielab_final:quantitative:multipartition}
\end{figure}
\subsection{Parameter learning}\label{sec:results:parameter_learning}
What is left to complete the algorithm is to choose a suitable value of the scale parameter $\alpha$. As we stated in the introduction, there is no such thing as the \emph{correct scale} for the general case. This notion is application and dataset dependent. Even given a dataset like the BSDS, not every image has the same scale, thus we should define an optimal set scale, which is the best scale you can choose for the complete dataset; and an optimal image scale, which is different for every individual image. In this work,  $\alpha$ represents the notion of optimal set scale.

In the first place, we proceeded to tune the $\alpha$ parameter manually, using the training subset of the BSDS to find a suitable value for $\alpha$. The goals established were the following: obtaining more balanced results in terms of the APDs, diminishing SPD and keeping the increase of MPD controlled. As a result, we found that $\alpha=50$ is a good compromise for this set of metrics and for this database. The results of this choice of the parameter are shown in Table \ref{table:results:quantitative:cardoso:test}, were we can see that the results are better than those of the previous algorithms.

Regarding the automatic parameter tuning, for the particular case of the BSDS, we ran the parameter estimation algorithm presented in section \ref{sec:meaningful_partitions:parameter}  with the 200 images from the training set, and obtained a value of $\alpha \approx 21$.  We also ran the same procedure with the gray level version of the algorithm, and we called these variants multipartition-supervised (MP-S) and multipartition-gray-supervised (MP-G-S). Their results over the test set of 100 images are shown in Table \ref{table:results:quantitative:cardoso:test}. 
As this table shows, the training procedure greatly reduces the number of false positives (FP) and slightly increases false negatives (FN), so the overall performance is clearly improved (10\% reduction in SPD) and the FPs and FNs are more balanced.
Moreover, in Figure \ref{im:set_cielab_final:quantitative:regions:n2sup} we can see how the distribution of the number of detected regions is shifted to the left, meaning that in average we are detecting fewer regions than before.

The training procedure presented before provides a considerable improvement of the results over the default value of the parameter, but its performance is lower than the manually selected parameter. Looking at both APDs we can see that the results are still slightly unbalanced. Please note that the optimization goal was to optimize the number of regions, but not any particular metric.
Regarding this supervised procedure, two more points need to be remarked. First, the kind of parameter that needs to be learned is very simple, so we do not really need a big number of training images, and the performance of that training is not affected when using only a few images, as opposed to \cite{im_proc:segmentation:martin:04:learning_natural_image_boundaries} which requires a very complex training process. Second, our optimization goal is not a particular performance metric as in most reference works, but to choose the correct scale.
This means that our optimization procedure does not guarantee a decrease in the APD, SPD or any other performance metric, just in the number of detected regions. This optimization goal makes the results independent of the metrics used. 
In addition, in Figure \ref{im:set_cielab_final:quantitative:regions} we show the distribution of the number of regions detected by human subjects and by the proposed algorithms. From this figure we can conclude that each version of the algorithm improves the previous ones in terms of detected regions and that the MP-M achieves the distribution most similar to the human subjects.

Although the effort of quantifying the results of the algorithms is praiseworthy and constitutes a good step towards the objective evaluation of scientific work, we should bear in mind that every set of metrics has its drawbacks and they are still far from our subjective perception.
Over-optimizing the algorithms with respect to a particular metric is a dangerous path to follow, and could lead to results which are far from human perception.
In the particular case of this algorithm, the best numeric results are obtained with $\alpha=150$ (see Table \ref{table:results:quantitative:cardoso:test}), however from a subjective point of view we think that those results are not as good as the ones obtained with $\alpha=50$. In our experience this is because we tend to penalize more the sub-segmentation rather than over-segmentation, and thus the results using $\alpha=50$ are more pleasant to the human eye.

Finally, when comparing with human made segmentation, caution should be exercised, because there are fundamental differences among the process of segmenting an image by a human observer and a machine. The human segmentation uses a higher level of abstraction than our low level tools, and thus the latter have a natural limit on their performance, which explains the differences we see with the human subjects.
\renewcommand{\figDir}{im/quantitative/set_cielab_7_final/plot}
\renewcommand{\figDirB}{im/quantitative/trained_2/plot}
\renewcommand{\figDirC}{im/quantitative/mp-m/plot}
\renewcommand{\anchodos}{3.8cm}
\begin{figure}[ht]
  \centering
  \subfloat{ \label{im:set_cielab_final:quantitative:regions:n1}%
  \includegraphics[width=\anchodos]{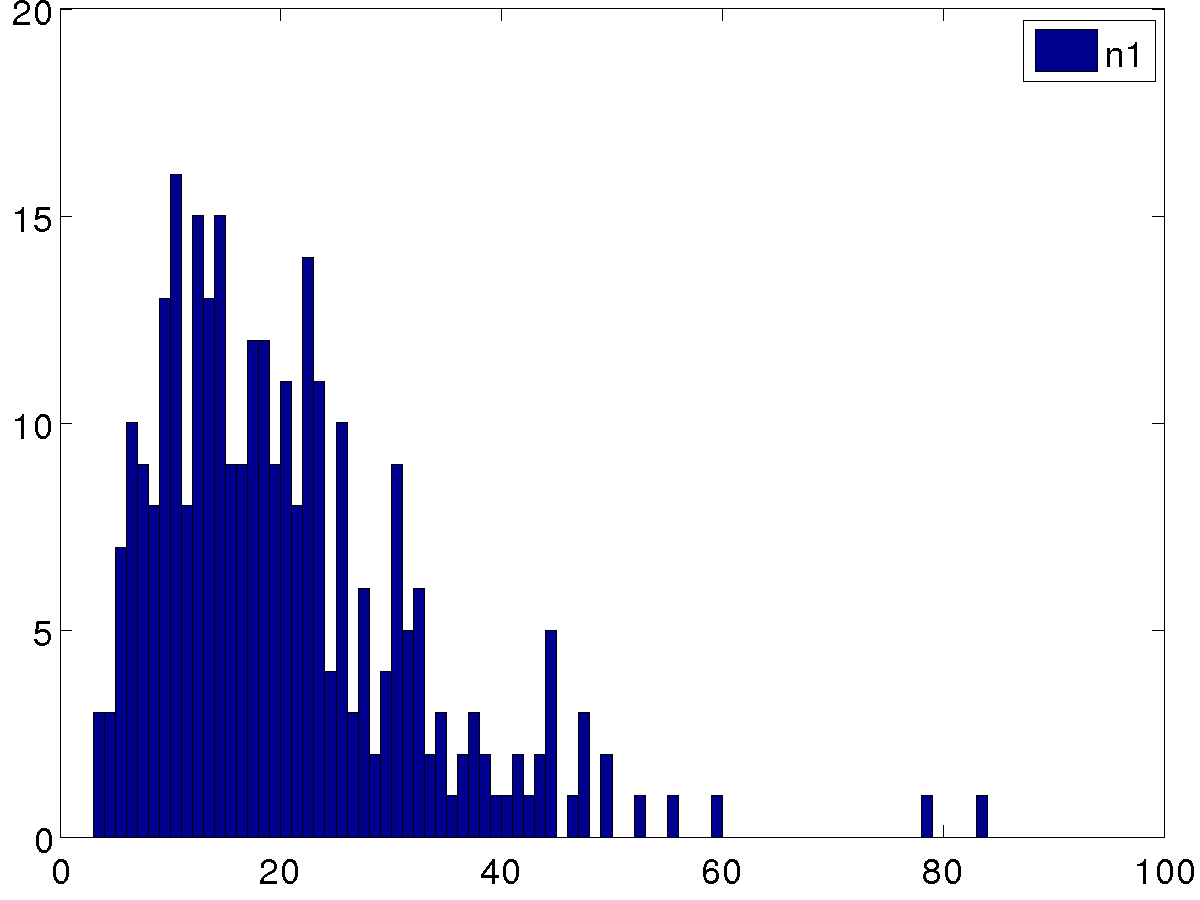}%
  }%
  \subfloat{ \label{im:set_cielab_final:quantitative:regions:n2}%
  \includegraphics[width=\anchodos]{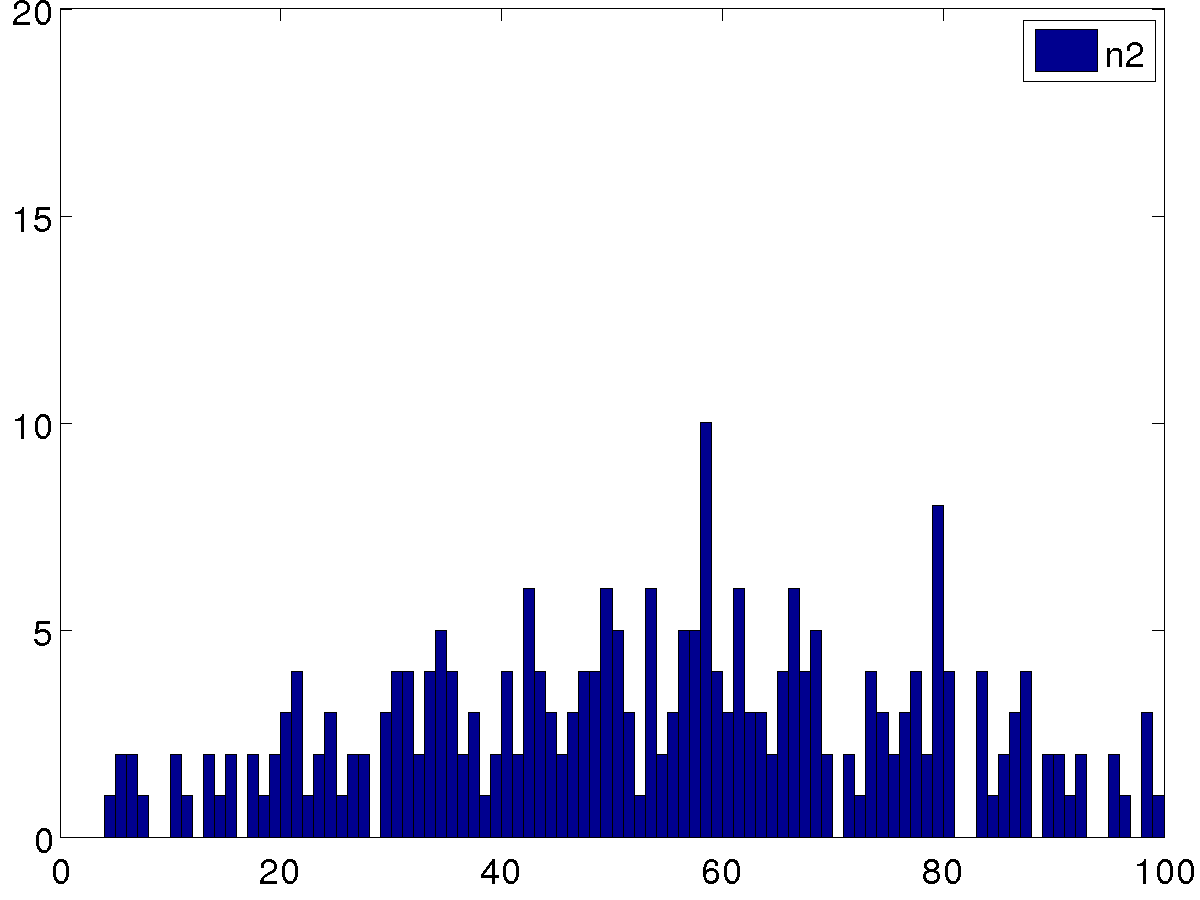}%
  }\\
  \subfloat{ \label{im:set_cielab_final:quantitative:regions:n2sup}%
  \includegraphics[width=\anchodos]{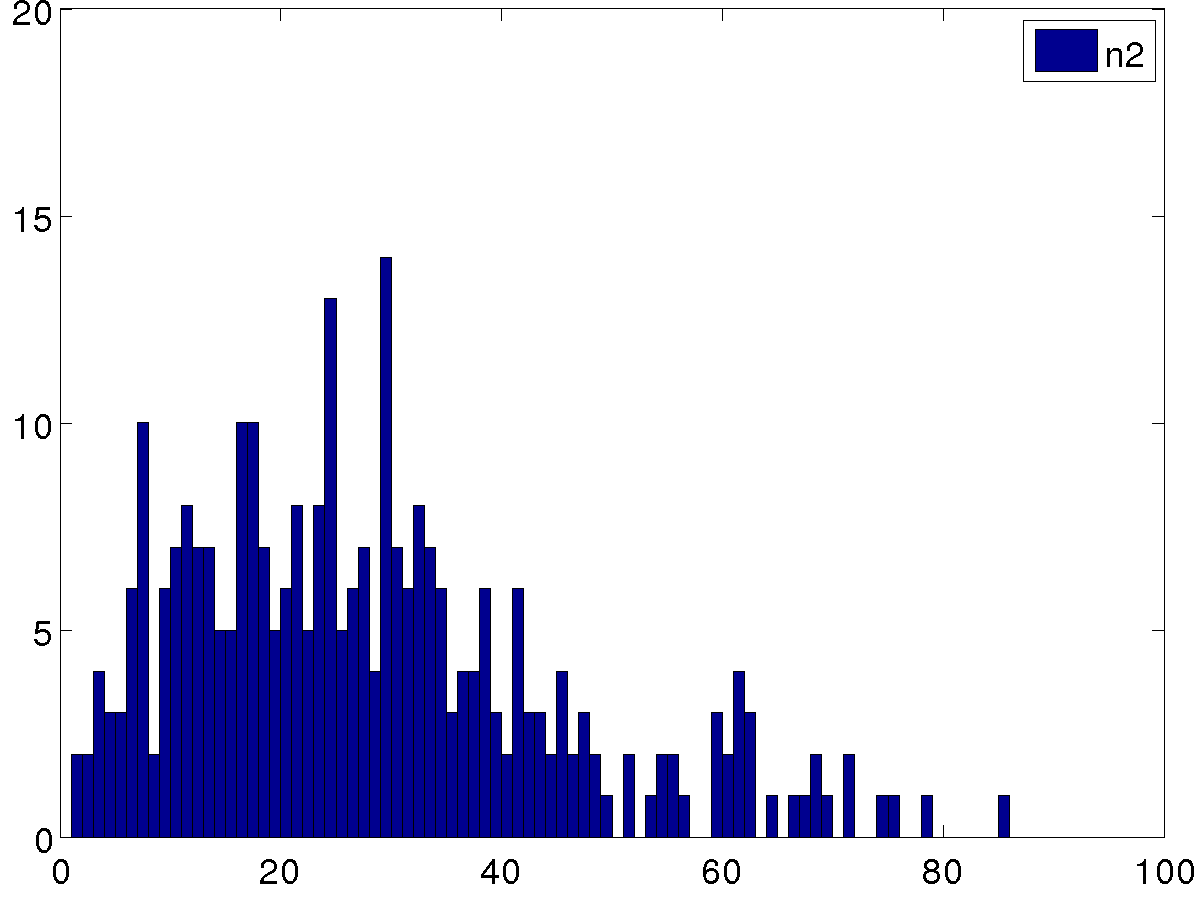}%
  }%
  \subfloat{ \label{im:set_cielab_final:quantitative:regions:n2_mp_m}%
  \includegraphics[width=\anchodos]{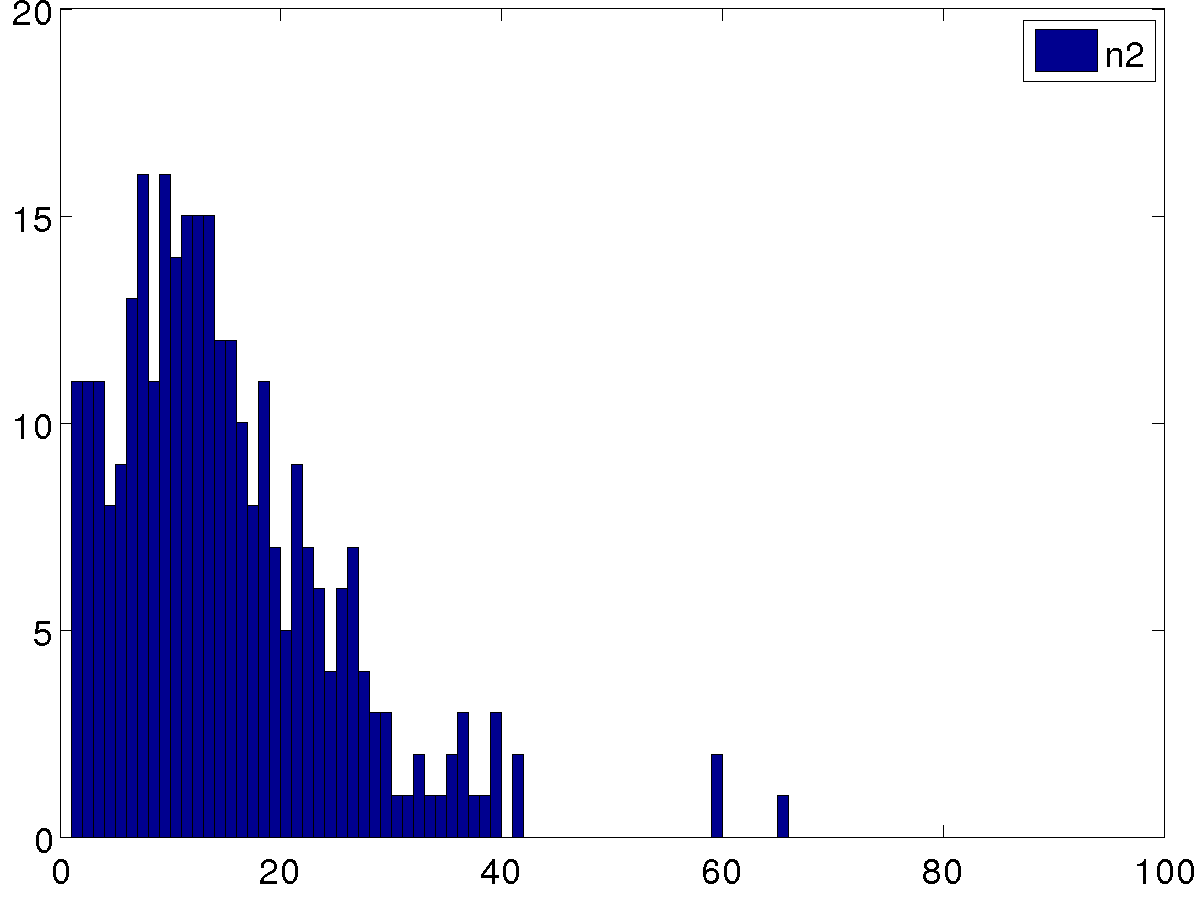}%
  }%
  \caption{Distribution of the number of detected regions over the database. From left to right, and top to bottom: human detected regions, result of the unsupervised (MP-D) algorithm, result of the supervised (MP-S) algorithm, result of the manually tuned (MP-M) algorithm.}
  \label{im:set_cielab_final:quantitative:regions}
\end{figure}
\subsection{Other datasets}
In order to extend our results we conducted the same validation procedure in the LHI database. One important remark about this database is the fact that it was designed for object recognition, and the human segmentations provided are of a higher level of abstraction, i.e. objects are marked as persons, cars, etc.
In Table \ref{table:results:quantitative:lhi:cardoso} we show the results of the proposed algorithms on 75 images from the following LHI subsets: 'Badminton', 'Highway', 'Ibis', 'Polo', 'Snowboarding', 'Street'. The obtained results are quite good in quality, although the scale is generally finer that the ground truth segmentations.
In the case of the MP-S algorithm, we didn't train the algorithm again but used the parameter obtained on the BSDS300. The first observation about these results is that all the conclusions extracted from the BSDS are still valid here, namely, the oversegmentation shown by the MP algorithm, and the further improvements provided by the MP-S and MP-M algorithms, which present more balanced results. The second observation is that the performance is lower than in the BSD. This is coherent since the 
human segmentations provided by the LHI database are of a higher level of abstraction, which in general corresponds to coarser partitions than those from the BSDS.
\begin{table}[t]
\centering
\begin{tabular}{|c||c|c|c|c|}
\hline
	alg. & \tiny{$SPD$} & \tiny{$APD(P,Q)$} & \tiny{$APD(Q,P)$} & \tiny{$MPD$} \\
\hline
	\verb+MP-D+	& 0.62 & 	0.61  &	 0.08  &	 0.07 \\
	\verb+MP-S+	& 0.55 & 	 0.52  &	 0.10  &	 0.08 \\
	\verb+MP-M+ 	&0.50	& 0.46	& 0.18	& 0.15 \\
\hline
\end{tabular}
\caption{Quantitative evaluation  of the proposed algorithms over a selected subset of the LHI database, using performance metrics from Cardoso et al. }
\label{table:results:quantitative:lhi:cardoso}
\end{table}%
\subsection{Comparison}\label{sec:results:comparison}
As we stated before, there are very few published algorithms presenting quantitative results. The first work on this area which presented an extensive evaluation was \cite{im_proc:segmentation:martin:04:learning_natural_image_boundaries}, followed by \cite{im_proc:segmentation:arbelaez:2006:boundary_extraction} and \cite{im_proc:segmentation:yang:2007:unsupervised_segmentation_of_natural}. These three works use the BSDS database and the Precision-Recall framework, which is good because it allows the community to compare results from a quantitative perspective. However this comparison is performed from a boundary detector oriented perspective, and they are not well suited to benchmark partitioning algorithms. The objections raised to this approach are discussed in detail in \cite{cardoso:2005:toward_a_generic_evaluation}. In \cite{im_proc:segmentation:yang:2007:unsupervised_segmentation_of_natural}  they also use the Probabilistic Rand Index (PRI) and the Variation of Information (VOI) to measure the quality 
of the results. In addition, \cite{calderero:2008:general_region_merging} shows quantitative results on a custom database (not publicly available at the moment of writing) using the metrics from Cardoso et al.

In addition, in \cite{im_proc:segmentation:arbelaez:09:from_contours_to_regions} and \cite{im_proc:segmentation:yima:2009:natural_image_segmentation_with_adaptive} the authors also made the great work of running the benchmarks for some  state of the art algorithms: 
\begin{itemize}
\item FH: "Efficient Graph Based Segmentation", Felzenszwalb and Huttenlocher (\cite{im_proc:segmentation:huttenlocher:2004:efficient_graph_based}).
\item CM: "Mean shift", Comaniciu and Meyer (\cite{im_proc:segmentation:comaniciu:02:mean_shift}).
\item NC: "Normalized Cuts", Shi and Malik (\cite{graph_cuts:malik:03:normalized_cuts}).
\item MS: "Multiscale algorithm for image segmentation", Koepfler et al. (\cite{Koepfler:1994:a_multiscale_algorithm_for_image_segmentation}).
\item WS: "Geodesic saliency of watershed contours and hierarchical segmentation", Najman et al. (\cite{morphology:watershed:najman:1996:geodesic_saliency}).
\item MFM: "Learning natural image boundaries", Martin et al. (\cite{im_proc:segmentation:martin:04:learning_natural_image_boundaries} and Canny \cite{segm:edge_detection:canny:86:computational_edge} ).
\end{itemize}
Those authors have made their code or binaries publicly available, but no quantitative results are given. Note that they had to tune the parameters of the algorithms manually, to perform a fair comparison.We also want to comment on the fact that many researchers are comparing with NC, FH and CM, since they have made their codes and/or results available although the results are not always the state of the art.

Even counting the aforementioned results, the number of algorithms with quantitative evaluation amounts to a couple of dozens counting this work, which is clearly insufficient for a broad area like low level image segmentation.

Finally, partition based performance measures are rarely used, so the results are hard to compare in this sense. We hope that this work could contribute to spread these kind of evaluation metrics and that we see in the future more works using them.

For the aforementioned reasons, in this section we show the comparison using Cardoso's metrics only for two algorithms: Texture and Boundary Encoding-based Segmentation (TBES) \cite{im_proc:segmentation:yima:2009:natural_image_segmentation_with_adaptive} and Ultrametric Countour Maps (UCM) \cite{im_proc:segmentation:arbelaez:2006:boundary_extraction,im_proc:segmentation:arbelaez:09:from_contours_to_regions}. 
We have chosen these algorithms because they have state of the art results and they provide a mean to reproduce their results.
In order to benchmark these algorithms, we ran the code provided by the authors with the parameters provided by them. The results of such comparison are shown in Table \ref{table:results:comparison:cardoso}, and they show that our algorithm presents a comparable (and slightly better) performance with the reference works.\\

The work of Guigues is very similar in spirit to our algorithm and thus it is important to compare with his approach.
We did our best effort to compare in a quantitative way, but at the moment of writing we did not have access to a suitable executable/source code to run the benchmarks.
\footnote{The authors pointed us to the open source Seascape software \cite{soft:segmentation:teixido::01:seascape}, which has a graphical user interface wich runs Guigues' algorithm.
However it was not possible to run it in batch mode in order to compute the results for the whole database. We also tried to develop our own executable based on Seascape code but without luck.}.
For these reasons we show in Figure \ref{fig:results:multiscale:saliency} only a qualitative comparison for a few random images taken from the BSDS500, the only criteria used to select them was to choose one image from each different category (city, animals, people, objects).
The current implementation of Scale-Sets uses a piecewise constant M-S model, using the three RGB channels as features. As this figure shows, Guigues has the same problem in slowly varying intensity regions, where they appear fragmented due to the piecewise constant model.
This can be seen in the sky of the \emph{city} image, where some salient boundaries are detected. In addition, some salient boundaries are detected in the background \emph{bird} image which are more salient than the bird itself, due to the camouflage effect.

These results show that the quality of the hierarchy obtained by the MP algorithm is better than the one from Guigues, because the most salient boundaries better match the structures present in the images.

\subsection{Alternative performance metrics}
Although we believe that the metrics presented by Cardoso et al. are the best choice for the evaluation of low-level segmentation algorithms, we faced the fact that very few works used them to present results. For that reason, and to ensure reproducibility, we present here performance results using the metrics proposed by Martin et al. in \cite{martin:2003:empirical_approach_to_grouping}. In that work, they introduce two sets of metrics: region based and boundary based.
From these metrics, we want to highlight the F-measure, which is the most widely used at the moment. The definition of the F-measure \cite{martin:2003:empirical_approach_to_grouping} relies in turn on the definition of two quantities: \emph{precision} and \emph{recall}. Precision (P) is the probability that a machine-generated boundary pixel is a true boundary pixel. Recall (R) is the probability that a true boundary pixel is detected. The F-measure is a way to summarize both quantities in an unique performance metric, and it is computed as the harmonic mean of P and R. The F-measure is bounded between 0 and 1 and values closer to 1 mean better performance. Note that precision and recall are boundary based metrics and thus they are not well suited (nor the F-measure) to evaluate region based algorithms.

The results using the aforementioned metrics are shown in Table \ref{table:results:quantitative:berkeley_boundaries}.
In spite of some differences, these results confirm the conclusions obtained from our first evaluation. First, that the MP algorithm clearly outperforms the GREEDY one. And second, that the MP-S algorithm improves MP performance and gives more balanced results in terms of false positives and negatives.

Regarding the supervised procedure, as we explained in section \ref{sec:results:parameter_learning}, it is not tuned to any particular metric, which could explain why our MP-S algorithm doesn't improve the F-measure obtained by the original MP algorithm.

\begin{table}[t]
\centering
\begin{tabular}{|c||c|c|c|c|}
\hline
 \multicolumn{5}{|c|}{Results over the 100 test images } \\
\hline
	alg. & \tiny{$SPD$} & \tiny{$APD(P,Q)$} & \tiny{$APD(Q,P)$} & \tiny{$MPD$} \\
\hline
	MP & 0.56 & 	 0.54  &	 0.10  &	 0.08 \\
	GREEDY & 0.58 & 0.54 & 0.13 & 0.10 \\	
	MP-S & 0.46 & 	 0.41  &	 0.13  &	 0.09 \\
	MP-M ($\alpha=50$) & 0.41  	& 0.31 & 	 0.19  &	 0.11 \\
	MP-M ($\alpha=150$) & 0.38  	& 0.18 & 	 0.28  &	 0.09 \\
	UCM & 0.39 & 0.32	   &	0.14   &	0.09  \\
	TBES & 0.41 & 0.34	 &	0.17   & 0.11	  \\
\hline
\end{tabular}
\caption{Quantitative comparison of algorithms over the BSDS using performance metrics from Cardoso et al.}\label{table:results:comparison:cardoso}
\end{table}%
\begin{table}[t]
\centering
\begin{tabular}{|c||c|c|c|}
\hline
 \multicolumn{4}{|c|}{Results over the 100 test images of the BSDS300} \\
\hline
	alg. & \tiny{F-meas} & \tiny{P} & \tiny{R}  \\
\hline
	\verb+human+ & 0.79 & 	 X  &	 X   \\
\hline
	\verb+MP+ & 0.64 & 	 0.736  &	 0.564   \\
	\verb+GREEDY+ & 0.544 & 0.772 & 0.420  \\
	\verb+MP-S+ & 0.624 & 	 0.606  &	 0.644  \\
	\verb+MP-M+ & 0.590 & 	 0.505  &	 0.710  \\
\hline
\end{tabular}
\caption{Quantitative evaluation  of the proposed algorithms over the BSDS300 using the F-measure.}\label{table:results:quantitative:berkeley_boundaries}
\end{table}
\begin{table}[t]
\centering
\begin{tabular}{|c|c|}
\hline
	alg. & \tiny{F-meas}  \\
\hline
	\verb+human+ & 0.79     \\
\hline
	\verb+MP+ & 0.639 	   \\
	\verb+GREEDY+ & 0.544  \\
	\verb+MP-S+ & 0.624  \\
	\verb+MP-M+ & 0.590  \\
\hline
	\verb+TBES+ \cite{im_proc:segmentation:yima:2009:natural_image_segmentation_with_adaptive} & 0.645  \\
\hline
\end{tabular}%
\begin{tabular}{||c|c||}
\hline
	alg. & \tiny{F-meas}  \\
\hline
\multicolumn{2}{||c||}{Results from \cite{im_proc:segmentation:arbelaez:2006:boundary_extraction}}\\
\hline
	\verb+UCM+ & 0.67     \\
	\verb+MS+ & 0.63  	 \\
	\verb+WS+ & 0.62  	 \\
	\verb+MFM+ & 0.65  	 \\
	\verb+canny+ & 0.58    \\
	\hline
\end{tabular}%
\begin{tabular}{|c|c|}
\hline
	alg. & \tiny{F-meas}  \\
\hline
\multicolumn{2}{|c|}{Results from \cite{im_proc:segmentation:arbelaez:09:from_contours_to_regions}}\\
\hline    
\verb+c-UCM+ & 0.58  \\
\verb+g-UCM+ & 0.71 \\
\verb+CM+ & 0.63  	   \\
\verb+FH+ & 0.58  	   \\
\verb+NC+ & 0.62  	   \\
\hline
\end{tabular}%
\caption{Quantitative comparison  of the proposed algorithms with the state of the art. Results computed over the test BSDS300 subset using the F-measure (ODS). Results were taken from Figure 4 of \cite{im_proc:segmentation:arbelaez:2006:boundary_extraction} and Table 1 of \cite{im_proc:segmentation:arbelaez:09:from_contours_to_regions}.}\label{table:results:quantitative:berkeley_boundaries:comparison}
\end{table}
In order to extend our comparison, we compare our algorithms with the state of the art using the F-measure. In Table \ref{table:results:quantitative:berkeley_boundaries:comparison} we present an excerpt of results presented in
 \cite{im_proc:segmentation:arbelaez:09:from_contours_to_regions} and \cite{im_proc:segmentation:yima:2009:natural_image_segmentation_with_adaptive} with the addition of our results. Note that c-UCM and g-UCM refer to the canny-owt-UCM and the gPb-owt-UCM algorithms presented in \cite{im_proc:segmentation:arbelaez:09:from_contours_to_regions}. We report here only the F-measure for the Optimal Dataset Scale (ODS), because some of the precision-recall values from the reference works are not available.

As these results show, our algorithm has a similar performance than the other state of the art algorithms, measured with the F-measure. Some algorithms have a significantly better performance than the MP algorithm.
\subsection{Multiscale evaluation}\label{sec:results:multiscale}
\begin{figure}[ht]
  \centering
  \includegraphics[width=6.5cm]{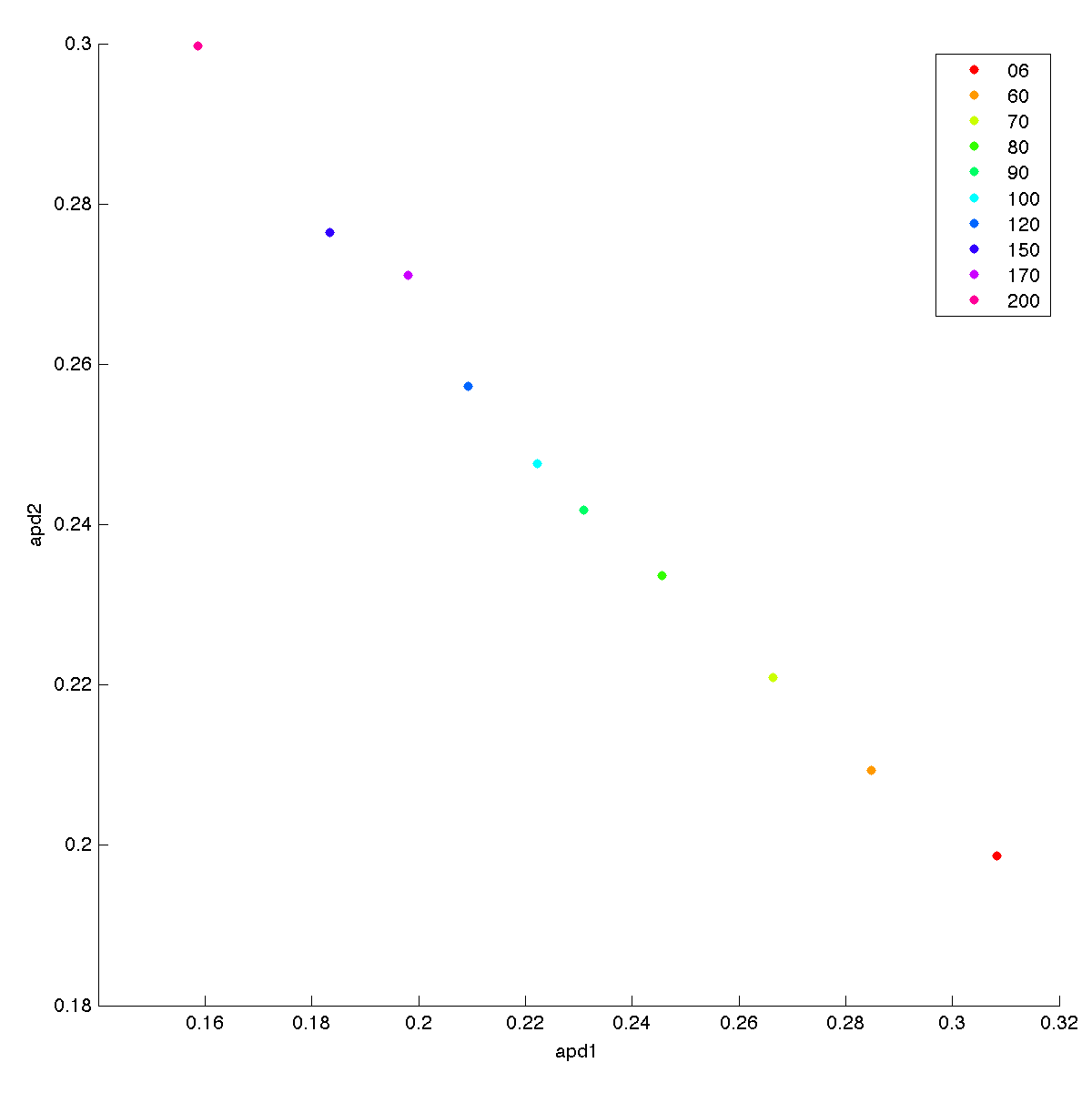}
   \includegraphics[width=6.5cm]{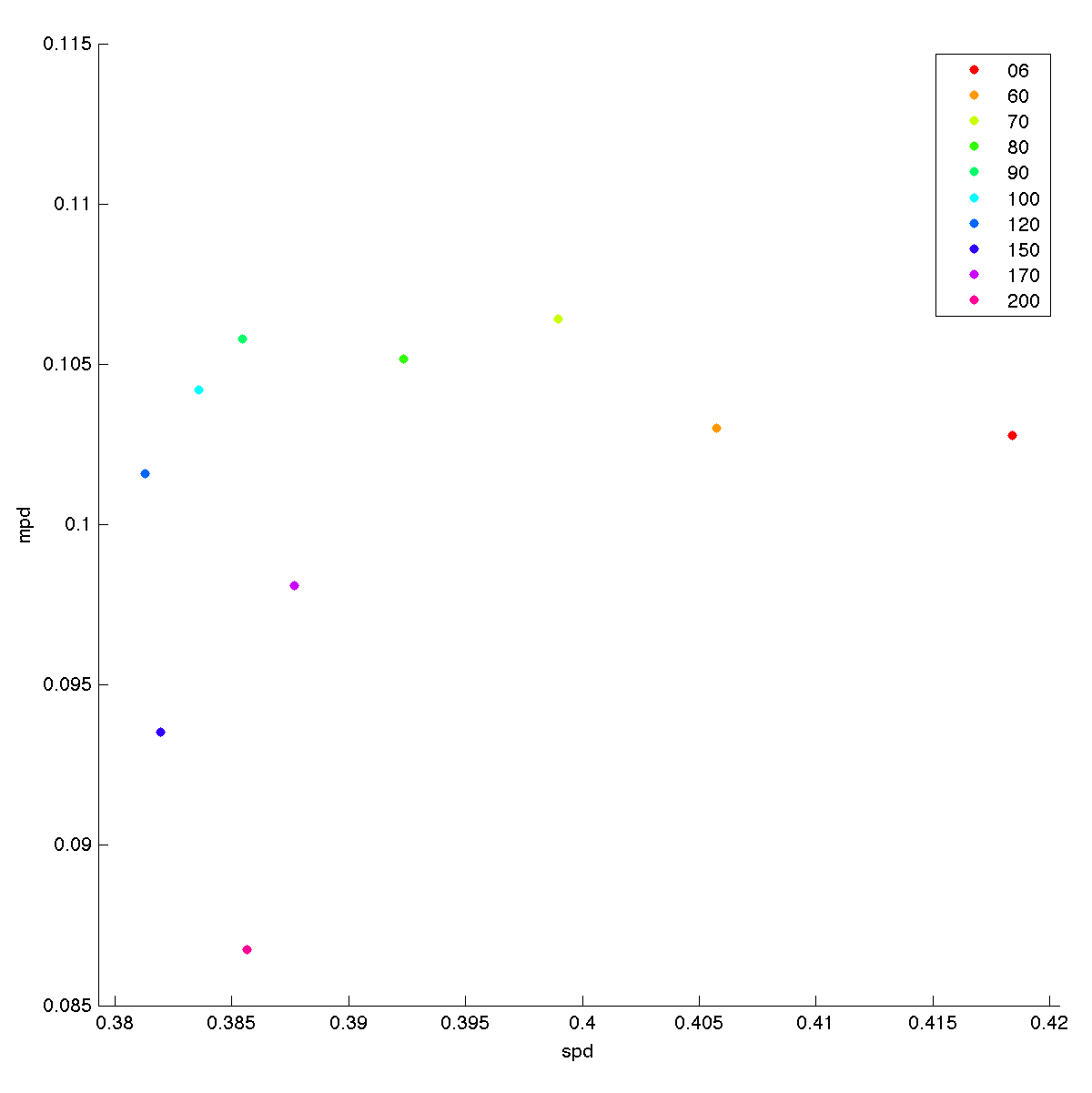}
  \caption{Results of the region based multiscale evaluation proposed in this work, using the BSDS300 database. Top: APD(P,Q) vs. APD(Q,P). Bottom: SPD vs. MPD. }
  \label{im:benchmark:cardoso:multi_scale}
\end{figure}

 So far we have shown single scale results using the PD metrics for a selected number of scales. In this section we show a multiscale evaluation based on these metrics, in the same spirit as the BSDS multiscale evaluation.
In Figure \ref{im:benchmark:cardoso:multi_scale} we show the results of varying the $\alpha$ parameter in the range $[6,200]$ for the BSDS300 database.
 As this figure shows, with the smaller parameter choice we have an over segmentation (high $APD(P,Q)$, low $APD(Q,P)$), as the parameter changes the balance of false positives vs false negatives towards an undersegmentation. This supports our claim that $\alpha$ behaves as a scale parameter.\\
\subsection{Results over the BSDS500 database}\label{sec:results:bsds_new}
In this section, we extend the evaluation to a third dataset, the 200 test images from the BSDS500 database. 
These images are new (not included in the BSDS300) and no training or tuning of the algorithms was carried out in this database. We use the exact configuration tested in the BSDS300.
This evaluation was computed with the new version of the Berkeley Benchmark, which now includes the computation of some region based quantities (PRI, VOI, Covering).

\begin{figure}[ht]
  \centering
  \includegraphics[width=8cm]{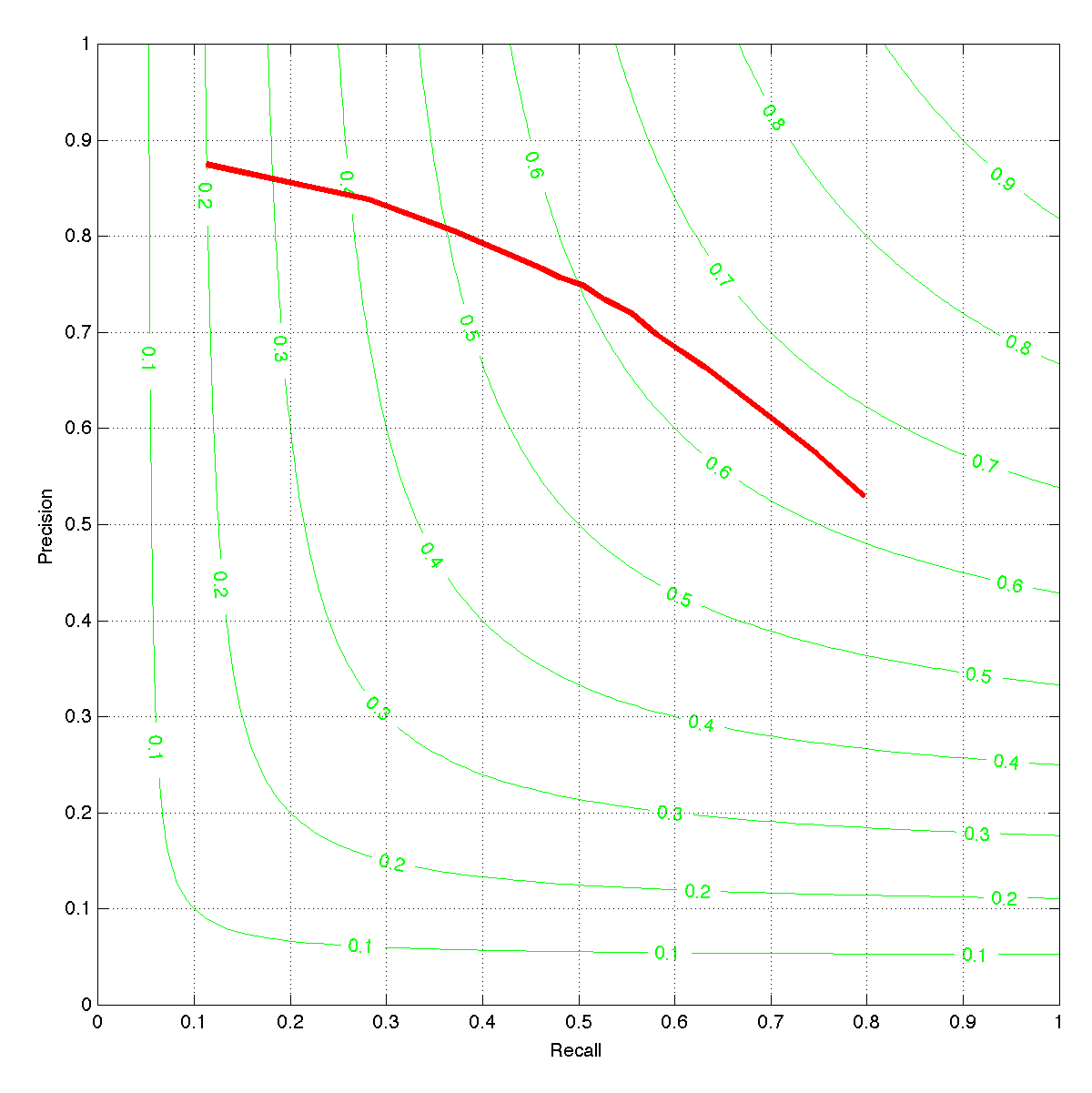}
  \caption{Results of the boundary based multiscale evaluation from Berkeley \cite{arbelaez:pami:2011:contour_detection_and_hierarchical_image_segmentation}.
  The red curve depicts the behaviour of the F-measure of the MP algorithm as the scale parameter $\alpha$ varies in the range $[0.01,5000]$. The curves in green are isocontours of the F-measure.}
  \label{im:benchmark:berkeley_new:multi_scale}
\end{figure}

The boundary based results are shown in Table \ref{table:results:quantitative:bsds500:berkeley_boundaries_detail}. These results confirm what we previously commented about the F-measure, i.e. it is unfair with region based approaches.
It is also interesting to note that the chosen ODS corresponds to $\alpha=10.8$ which is very different from the chosen scale to optimize the PD metrics $\alpha=150$.
In Figure \ref{im:benchmark:berkeley_new:multi_scale} we show the results of the boundary based multiscale benchmark of Berkeley.
These results confirm that $\alpha$ acts as a scale parameter: when $\alpha$ is small, the algorithm shows high Precision and low Recall values, which corresponds to a segmentation with a big number of false negatives and a low value of false positives; when 
$\alpha$ increases the balance leans towards oversegmentation, i.e. Precision falls and Recall rises.
Finally, to complete the boundary based evaluation, in Table \ref{table:results:quantitative:bsds500:berkeley:boundaries_comparison} we compare with the reference works using the F-measure on the BSDS500.
As the results show, we obtain state of the art results, only surpased by Arbelaez et al. \cite{arbelaez:pami:2011:contour_detection_and_hierarchical_image_segmentation}.
Note that the difference in performance between Arbelaez's algorithm and ours is bigger when quantified with the F-measure, which is explained by the fact that Arbelaez's algorithm is trained with the F-measure as the optimization goal.
As our algorithm uses an initial pruning, the $R=0$ value is never attained, and the same happens with $P=0$, thus the P-R curve is not \emph{complete}. 
This means that the measured $AP$ value is not accurate.\\

In Table \ref{table:results:quantitative:bsds500:berkeley_region} we show the results of the Berkeley region based metrics. In the ODS case, we show the value of $\alpha$ that achieved that result.
In Table \ref{table:results:quantitative:bsds500:berkeley_region_comparison} we compare our results with the available results from \cite{arbelaez:pami:2011:contour_detection_and_hierarchical_image_segmentation} using the Berkeley region based metrics. 
As these results show, our algorithm obtains good results, only surpassed by the work of Arbelaez et al. which is significantly better. For the rest of the compared algorithms we obtain better values for almost every metric.
\begin{table}[t]
\centering
\begin{tabular}{|c||c|c|c||c|c|c||c|}
\hline
 \multicolumn{8}{|c|}{Results over the 200 test images of the BSDS500} \\
\hline
alg & \multicolumn{3}{|c||}{\tiny{ODS} \tiny{($\alpha\; =\; 10.8$)}} &  \multicolumn{3}{|c||}{\tiny{OIS}} &  \\
\hline
alg & P & R & \tiny{F-meas} &  P & R & \tiny{F-meas} & AP \\
\hline
MP & 0.70 & 0.61 & 0.65 & 0.72 & 0.66 & 0.69 & 0.51 \\
\hline
\end{tabular}
\caption{Quantitative evaluation of the proposed algorithms over the BSDS500 using Berkeley boundary based measures. }
\label{table:results:quantitative:bsds500:berkeley_boundaries_detail}
\end{table}
\begin{table}[t]
\centering
\begin{tabular}{|c||c|c|c||c|c|c||}
\hline
 \multicolumn{7}{|c|}{Results over the 200 test images of the BSDS500} \\
\hline
  & \multicolumn{3}{|c||}{\tiny{ODS} } &  \multicolumn{3}{|c||}{\tiny{OIS} } \\
 \hline
	      & GT	& PRI	& VOI	& GT	& PRI	& VOI  \\
 \hline
 metric value & 0.52	& 0.79	& 1.87	& 0.58	& 0.83	& 1.65 \\
 $\alpha$ values & 80 & 20 & 300 & &   &  \\
\hline
\end{tabular}
\caption{Quantitative evaluation of the MP algorithm over the BSDS500 using Berkeley region based measures.}\label{table:results:quantitative:bsds500:berkeley_region}
\end{table}
\begin{table}[t]
\centering
\begin{tabular}{|c||c|c|c|}
\hline
 \multicolumn{4}{|c|}{Results over the 200 test images of the BSDS500} \\
\hline
alg & \multicolumn{2}{|c|}{\tiny{F-meas}} & \tiny{AP} \\
\hline
	 & \tiny{ODS} & \tiny{OIS} &   \\
\hline
	human & 0.79 & 	 0.79  &	 X   \\
\hline
	gPb-owt-ucm & 0.73 & 	 0.76  &	 0.73  \\
	MP & 0.65 & 	 0.69  &	 0.51   \\
	CM & 0.64 & 	 0.68  &	 0.56   \\
	NC & 0.64 & 	 0.68  &	 0.56   \\
\hline
\end{tabular}
\caption{Quantitative evaluation of the proposed algorithms over the BSDS500 using boundary based measures.
The results of the related works were extracted from Table 1 of \cite{arbelaez:pami:2011:contour_detection_and_hierarchical_image_segmentation}.}
\label{table:results:quantitative:bsds500:berkeley:boundaries_comparison}
\end{table}
\begin{table}[t]
\centering
\scriptsize{
\begin{tabular}{|c||c|c|c||c|c||c|c|}
\hline
 \multicolumn{8}{|c|}{Results over the 200 test images of the BSDS500} \\
\hline
 \tiny{ alg} & \multicolumn{3}{|c||}{\tiny{Covering} } &  \multicolumn{2}{|c||}{\tiny{PRI} }  & \multicolumn{2}{|c|}{\tiny{VI} }\\
 \hline
		& \tiny{ODS}	& \tiny{OIS}	& \tiny{Best}	& \tiny{ODS}	& \tiny{OIS}	& \tiny{ODS}	& \tiny{OIS}	\\
 \hline
 \tiny{human}		& 0.72	& 0.72	& 	& 0.88	& 0.88	& 1.17	& 1.17 \\
 \hline
 MP			& 0.52	& 0.58	& 0.68	& 0.79	& 0.83	& 1.87	& 1.65 \\
 gPb-owt-ucm		& \bf{0.59}	& \bf{0.65}	& \bf{0.74}	& \bf{0.83}	& \bf{0.86}	& \bf{1.69}	& \bf{1.48} \\
 CM		& 0.54	& 0.58	& 0.66	& 0.79	& 0.81	& 1.85	& 1.64 \\
 FH			& 0.52	& 0.57	& 0.69	& 0.80	& 0.82	& 2.21	& 1.87 \\
 Canny-owt-ucm\!	& 0.49	& 0.55	& 0.66	& 0.79	& 0.83	& 2.19	& 1.89 \\
 NCuts			& 0.45	& 0.53	& 0.67	& 0.78	& 0.80	& 2.23	& 1.89 \\
 Quad-Tree		& 0.32	& 0.37	& 0.46	& 0.73	& 0.74	& 2.46	& 2.32 \\
\hline
\end{tabular}
}
\caption{Quantitative evaluation of the MP algorithm over the BSDS500 using Berkeley region based measures. The results of the related works were extracted from Table 2 of \cite{arbelaez:pami:2011:contour_detection_and_hierarchical_image_segmentation}.}
\label{table:results:quantitative:bsds500:berkeley_region_comparison}
\end{table}
\subsection{Fixed number of regions}\label{sec:results:fixed}
\begin{figure}[ht]
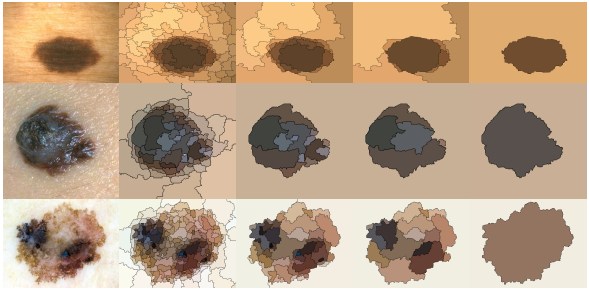

  \centering
  \figRowAlgos{ejemplo14}
  \figRowAlgos{ejemplo17}
  \figRowAlgos{ejemplo18}
  \caption{Results of three of the proposed algorithms over skin lesion images. From left to right: original image, initial partition, result of the MP algorithm, result of the MP-S algorithm, result of the algorithm imposing a 2-region constraint.}
  \label{im:fixed_regions}
\end{figure}

As we stated before, our approach gives us the flexibility of specifying the number of desired regions beforehand. This could be particularly useful in a number of applications such as skin lesion detection, were we want to detect one single mark in the skin. In Figure \ref{im:fixed_regions} we show the result of such approach for the detection of melanomas in dermatological images. As this figure shows, the original algorithm gives the correct number of regions in some cases and fails in some others. In these cases the fixed region variant helps solving the problem.
\renewcommand{\figDir}{im/fixed_regions/melanomas/melanomas_mp}
\renewcommand{\figDirB}{im/fixed_regions/melanomas/melanomas_train}
\renewcommand{\figDirC}{im/fixed_regions/melanomas/melanomas_orden}
\renewcommand{\anchocinco}{1.55cm}
\subsection{Summary of experimental results}\label{sec:results:summary}
We evaluated our algorithm in a very exhaustive way, using many different sets of metrics and various databases. From all the results we can conclude that our algorithm presents good results, which are comparable with the state of the art.
However the results of Arbelaez et al. are better than ours. This is explained by various factors, but mainly by the fact that we focus on the model of partition selection developed and we did not tune the ingredients. 
Choosing a better initial segmentation, a different set of features (we use only color information), a better region model (we use a piecewise constant one) and a better boundary detector, we could improve our results considerably. 
In Arbelaez's work, all these factors were carefully chosen. In addition, these choices and the values of the parameters were made by training over the train subsets of the BSDS, whereas we don't use any training process.
\section{Implementation}\label{sec:implementation}
Our algorithms are implemented in C and C++, with automatic code optimization turned on. The execution times of the algorithms depend on image content; they have a strong dependency on the size of the initial partition, and a weak dependency on the size of the image. In Figure \ref{im:results:implementation:time} we show the distribution of the execution times over the BSDS300 database. The average times and their standard deviations are show in Table \ref{table:results:execution_times}. These times were computed on an Intel(R) Core(TM)2 Duo CPU E7300  @ 2.66GHz, with 3Gb RAM, which was a standard desktop computer in 2008.
The first conclusion is that the family of developed algorithms has reasonable execution times, approximately 5 seconds in average. However, with a careful implementation these times could be considerably lowered.
Second, the different variants of the MP algorithms (obtained by varying the scale parameter) do not change the execution time, which is coherent because the main computational effort is invested in the construction of the tree and the computation of the NFA. This is the opposite behaviour as in many merging algorithms where the execution time depends on the amount of merging performed, which in turn depends on the value of the scale parameter.
Third, the execution time of all the algorithms depends only on the size ($N_i$) of the initial partition, as predicted by our formulation in sections \ref{sec:meaningful_regions:complexity} and \ref{sec:meaningful_partitions:optimality}.
\begin{table}[ht]
\centering
\begin{tabular}{|c|c|c|c|}
\hline
	alg	& $N_i $(regions)	& $\mu$(secs.)  & $\sigma$(secs.) \\
\hline
	GREEDY	& 605	& 14.14	& 3.74		\\
	MP-D 	& 222	& 5.50	& 1.71		\\
	MP-S 	& 222	& 5.37	& 1.74	\\
	MP-M 	& 222	& 5.27	& 1.66	\\
\hline
\end{tabular}
\caption{Execution times for the different versions of the algorithms on the complete BSDS300 database. In each case $N_i$ is the average number of regions of the initial partition used. $\mu$ and $\sigma$ are the corresponding mean and standard deviation of the execution times (in seconds).}\label{table:results:execution_times}
\end{table}
\renewcommand{\anchotres}{3.5cm}
\newcommand{\figDirA}{im/quantitative/set_cielab_7_final/plot}
\renewcommand{\figDirB}{im/quantitative/ms_greedy/plot}
\newcommand{\figDirD}{im/quantitative/mp-m/plot}
\renewcommand{\figDirC}{im/quantitative/trained_2/plot}
\begin{figure}[ht]
  \centering
  \subfloat{ \label{im:results:implementation:time:greedy}%
  \includegraphics[width=\anchotres]{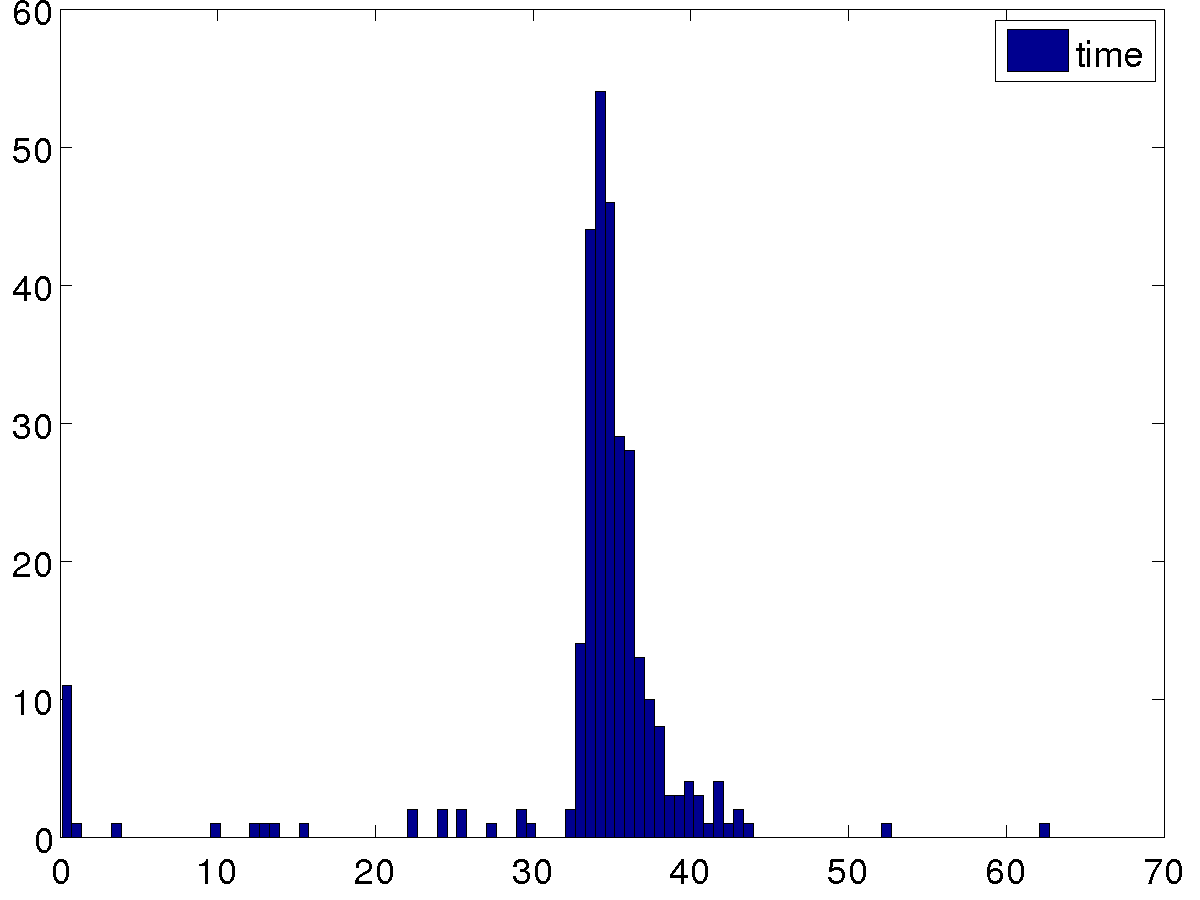}%
  }%
  \subfloat{ \label{im:results:implementation:time:mp}%
  \includegraphics[width=\anchotres]{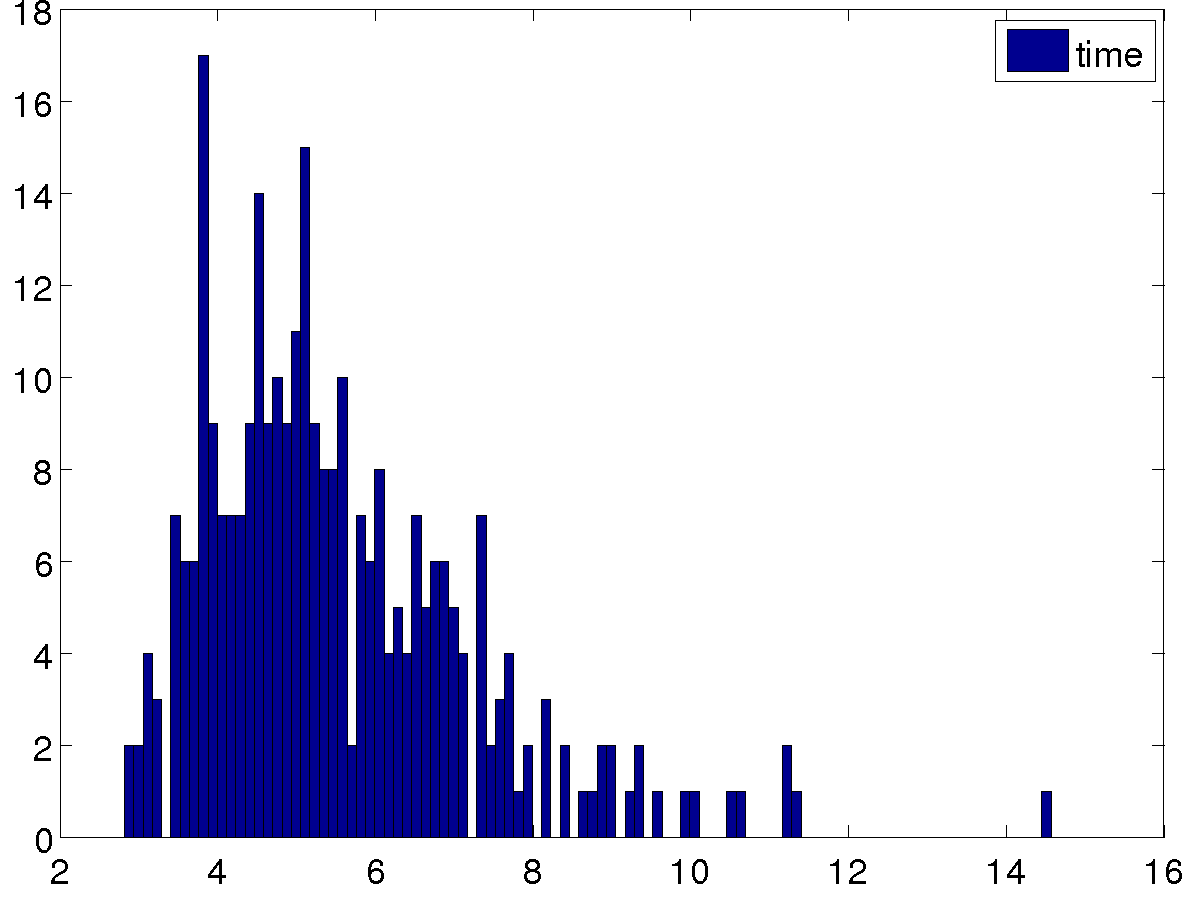}%
  }\\	
  \subfloat{ \label{im:results:implementation:time:mp_sup}%
  \includegraphics[width=\anchotres]{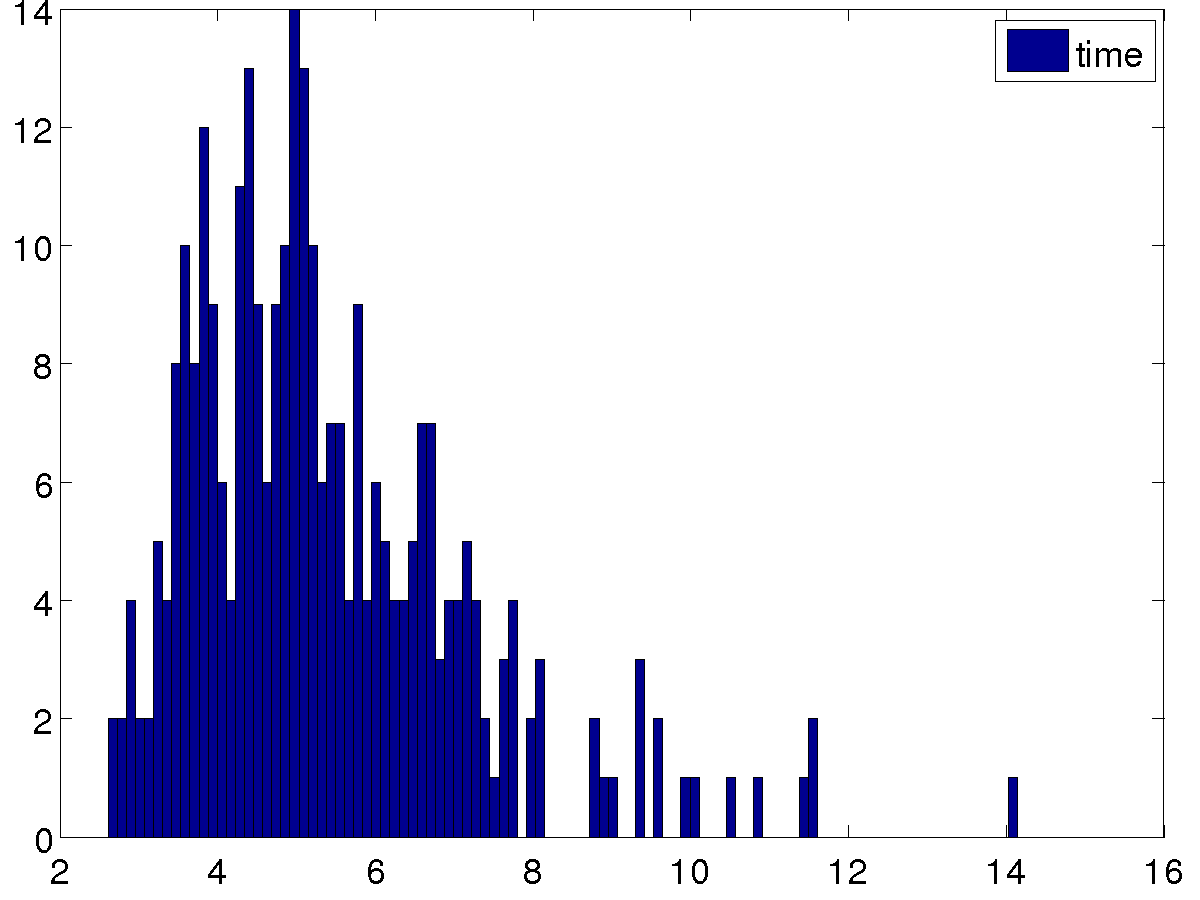}%
  }%
  \subfloat{ \label{im:results:implementation:time:mp_m}%
  \includegraphics[width=\anchotres]{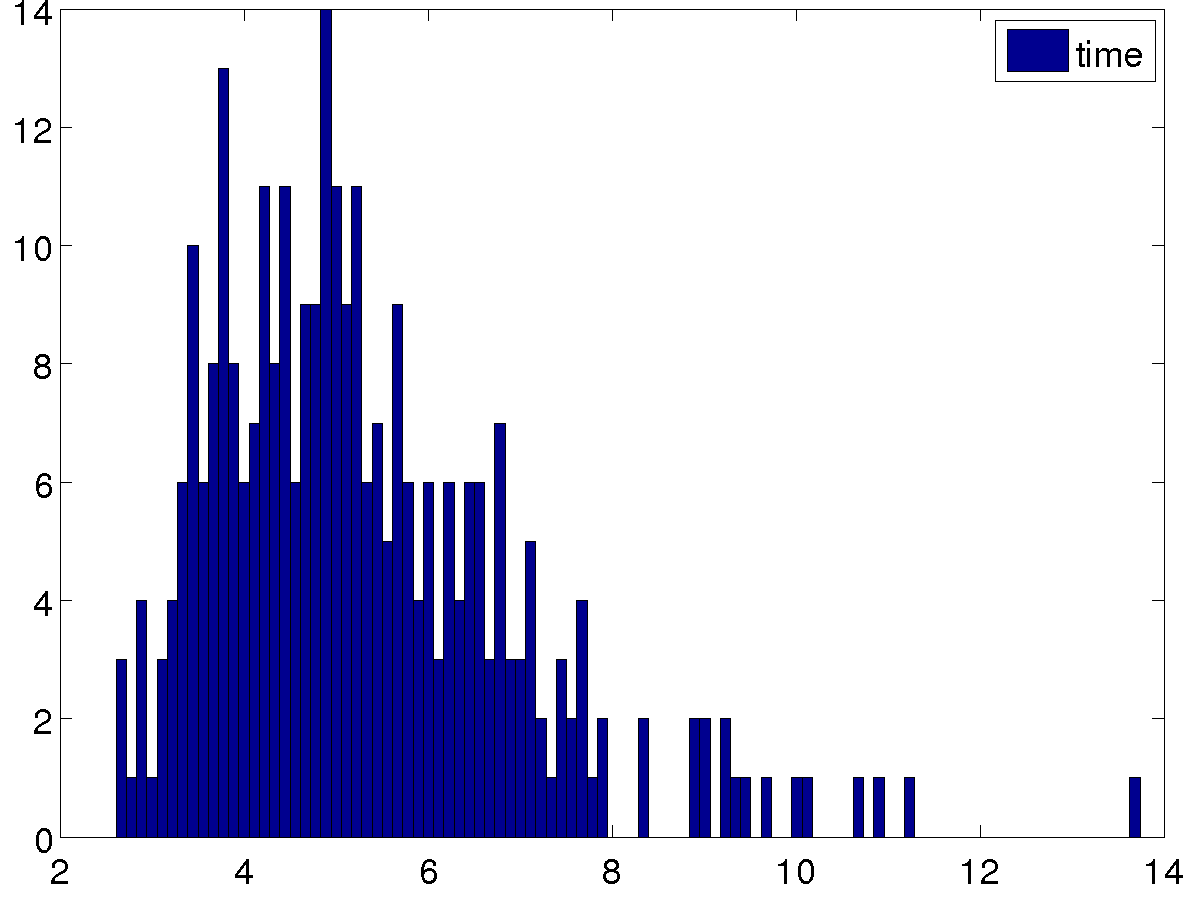}%
  }%
  \caption{Distribution of the execution times (in seconds) over the BSDS300 database for the different algorithms presented. From left to right:  GREEDY, MP, MP-S, MP-M.}
  \label{im:results:implementation:time}
\end{figure}
\section{Conclusions}\label{sec:conclusions}
In this work we presented a method for image segmentation based on a hierarchy of partitions of an image.
We introduced an \emph{a contrario} statistical model based on the combination of region and boundary based descriptors, which allows us to validate the partitions present in the hierarchy.
In this way we reduce the number of free parameters and alleviate the problem of local minima shown by greedy algorithms.
The results obtained by our algorithm are comparable to those presented in the literature. We have conducted an extensive quantitative evaluation using many different performance metrics and two publicly available datasets.

The main drawback of our approach is the underestimation of the NFA, which leads our algorithm to oversegment images. We have addressed this problem using an example-based training process to adjust the optimal scale for a given database. However, part of the problem is still there, and we need to address this on a future work. 
Another point that needs further work is the way we include boundary information in the process, because we are currently using a simple version of the boundary detector of Desolneaux et al. Thus, we could improve this by modifying the boundary detector to introduce all the improvements presented in \cite{segm:cao:03:extracting_meaningful}, which would help to remove the apparent boundaries found by our algorithm.

The main focus of this work was to present a generic partition validation approach, which has proven to be useful for the segmentation tasks. However, our results are conditioned by the quality of the algorithm used for the initial segmentation (M-S in this case) and the features used to describe regions (color information). In addition, our boundary detection term is somehow weak, as mentioned before. In this sense, there is still room for improvement in many ways. In the first place, we could use the gPb contour detector presented in \cite{segmentation:Arbelaez:2010:contour_detection_and_hierarchical_segmentation} to improve the initial segmentation, and we could also add texture information to the region descriptors.

We also want to test our framework with 3D and general multi-valued images (to deal with textured images). While our model is in fact independent of the dimension of the image domain and of the pixel dimension, and we have some preliminary results, we still need extensive experimentation in these cases.

It would also be interesting to add a summarizing quantity to the PD metrics analogous to the F-measure, which could allow us to compute a single score for a segmentation in our benchmark.
Finally, the computational cost of the partition selection mechanism needs to be improved, which will make the algorithm useful in more cases.
\section{Acknowledgements}
We thank J. Cardoso for kindly sharing and explaining the code of his algorithm to us. We also thank P. Arbelaez for sharing their benchmarks and results with us.
We thank L. Guigues for pointing us to the Seascape software where his algorithm runs.
We would like to thank Starlabs' staff, in particular to Aureli Soria and Anton Albajes-Eizaguirre for helping witht the Seascape code.
J. Cardelino and V. Caselles acknowledge partial support by MICINN project,
reference MTM2009-08171 and by GRC, reference 2009 SGR 773, funded by the Generalitat de Catalunya.
V. Caselles also acknowledges partial support by ''ICREA Acad\`emia'' prize for excellence in research funded by the Generalitat de Catalunya, and by the ERC Advanced Grant INPAINTING (Grant agreement no.:  319899).
M. Bertalm\'{i}o acknowledges support by European Research Council, Starting Grant ref. 306337, and Spanish grants AACC, ref. TIN2011-15954-E, and Plan Nacional, ref. TIN2012-38112.
J. Cardelino also acknowledges partial support by ALFA-CVFA project and Tecnocom scolarship.
\appendices
\section{Complexity analysis}\label{sec:appendix:complexity}
\begin{figure*}[ht]
  \centering
  \includegraphics[width=12cm]{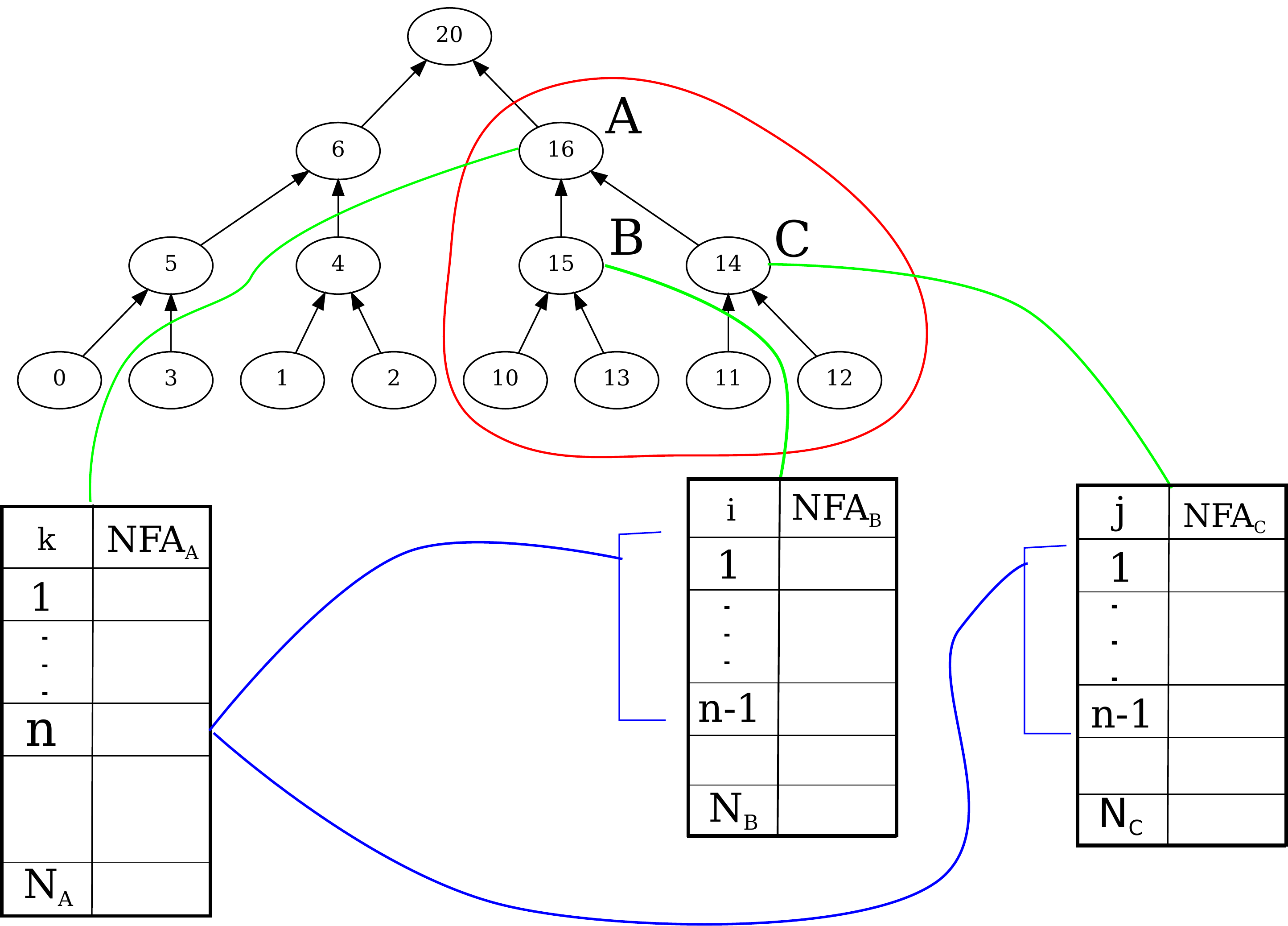}
  \caption{Example computation of the NFA.}
  \label{review:nfa_table}
\end{figure*}
In this appendix we will analyze the complexity of the dynamic programming scheme used to update the NFA tables. We will analyze what we believe are the most extreme cases with respect to complexity, which are dependent on the structure of our binary tree.
Those cases are: a completely degenerated or a completely balanced tree. This is not a formal proof but merely a conjecture. The hypothesis behind this reasoning is that the balanced tree is the worst case with respect to computational cost and the degenerated tree is the best case. We will discuss this assumption at the end of this section.\\

Given a node A, of a tree $T$, let us define $T_A$ as the subtree which has $A$ as root.
Suppose that we want to fill the NFA table of a node A, which will be denoted $table_A$, which has children B and C (See Figure \ref{review:nfa_table}). Suppose that we have already computed both tables corresponding to the subtrees $T_B$ and $T_C$.
Given a node A, the number of possible partitions that spans is bounded by $[1,N_A]$, where $N_A$ is the number of leaves $T_A$. We will compute this number later.

\subsection{Computation of the NFA table for a node}
We need to fill a table of NFAs of $N_A$ entries. To compute the NFA at level $n$, we will explore all the NFA values at levels $i,j$ such that $i+j=n$, taking $i$ from $table_B$ and $j$ from $table_C$.
In the following we will compute the number of possible ways to do that, denoted as $f_A(n)$.\\
Remember that $N_A=N_B+N_C$. Without losing generality, we can assume that $N_B<N_C$ and we will split the range of $N_A$ in three ranges:
\begin{itemize}
 \item $n \leq min(N_A,N_B)$. In this case the number of combinations is $f_A(n)=n-1$.
 \item $n > min(N_A,N_B)$ and $n \leq max(N_A,N_B)$. In this case, after we use all the values of $N_B$ we don't have more options, so $f_A(n)=N_B$.
 \item $n > max(N_A,N_B)$. This case is the same as the previous. $f_A(n)=N_B$.
\end{itemize}

Thus, the total number of computations needed to fill the table is 
\begin{eqnarray}\label{eq:complexity_analysis:cases}
F_A=\sum_{n=1}^{N_A-1}f_A(n)= \sum_{n=1}^{N_B-1} (n-1) + \sum_{n=N_B}^{N_A-1} N_B \nonumber \\
= \frac{(N_B-1)N_B}{2} + (N_A-1-N_B)N_B \\
= \frac{(N_B-1)N_B}{2} + (N_C-1)N_B \nonumber 
\end{eqnarray}

The numbers $N_A,N_B,N_C$  depend on the structure of the tree, but there are two extreme cases: the degenerate tree and the balanced tree.
In the degenerate case: $N_B=1$ and $N_C=N_A-1$, on the other hand, in the balanced case, $N_B=N_C=\frac{N_A}{2}$. Substituting this into Eq. \ref{eq:complexity_analysis:cases} we obtain the two boundary values for the number of cases:
\begin{equation}
F_A= 
\begin{cases} 
N_C-1=N_A & \text{ if } T_A  \text{ is degenerated}\\ 
\frac{3}{8} N_A^2-\frac{1}{4} N_A & \text{ if } T_A  \text{ is balanced}
\end{cases}
\end{equation}

\subsection{Computation for every node in the tree}
In order to carry on the computation, we need to add up the cost of computing the table for every node. That is
\begin{equation}
\sum_{A \in T}F_A
\end{equation}
This of course depends on the structure of the tree, so it can be counted in this form. However, if we take into account that the complexity depends on $N_A$ and on the structure of the tree, we can rewrite this equation in a more compact way if we arrange the nodes by the type of their subtrees.
\begin{equation}
\sum_{t}F_A(t).\mathcal{N}(t)
\end{equation}
where $t$ is a type of subtree and $\mathcal{N}(t)$ is the number of those subtrees. Of course this notion is vague, and cannot be counted, but in certain cases we could do it.
For instance, if the tree is balanced, every possible subtree is also balanced, same if the tree is degenerated.
The important thing to be noted is that the complexity depends on the number of type, the size of that type and the $F_A$ of that type.
Thinking in that way, the balanced tree is the worst case for 2 (out of 3) of those quantities.
For example, this tree has the minimum number of different sizes of subtrees, it has the biggest number of nodes per level and the highest $F_A$ per node.

One possible way to count this number is classifying nodes by their height $h$. We have a total of $\log(n)$ height values, and for each subtree of height $h$ there are $\mathcal{N}(h)=2^{-h}$ nodes, and each $A$ node has $N_A=2^h$ leaves.
Summarizing, in the balanced case, the number of operations is:
\begin{equation}
\sum_{h=1}^{\log(N/2)}F_{A_h}.\mathcal{N}(A_h)
\end{equation}
where $N$ is the total number of nodes of the tree. The key observation is that $F_A$ and $N_A$ in these cases only depends on $h$, because all the subtrees have the same structure.
\begin{equation}
\sum_{h=1}^{\log(N/2)} \left( \frac{3}{8} N_h^2-\frac{1}{4} N_h \right) \mathcal{N}(h)
\end{equation}
taking into account that \mbox{$N_h=2^h$} and \mbox{$\mathcal{N}(h)=N/2^{h+1}$} the total number of operations yields
\begin{equation}
\mathcal{C}(N)=\Theta(N^3)
\end{equation}
The main reason for the high complexity of this part, is due to the fact that we maintain a table of all possible partitions for each node which is far more complex than keeping only the best partition for each node.
\bibliographystyle{unsrt}
\bibliography{mp_siims_v7.bbl}

\end{document}